\documentclass[11pt, hidelinks]{article}
\usepackage{graphicx, bbm, setspace, longtable, lscape, verbatim, dcolumn, natbib, multirow, hhline, xr, amsmath, authblk, fullpage, amsthm, amssymb, ifthen, mathrsfs, float, xcolor, authblk, tikz, etoolbox, wrapfig, subcaption, bibentry, eurosym, amsthm, pdfpages, rotating, adjustbox, endnotes, mathtools, thmtools, thm-restate, hyperref, mdframed}
\declaretheorem[name=Proposition]{proposition}
\declaretheorem[name=Theorem]{theorem}
\declaretheorem[name=Lemma]{lemma}
\declaretheorem[name=Corollary]{corollary}

\declaretheorem[name=Assumption]{assumption}

\declaretheoremstyle[
headfont=\normalfont\bfseries, 
bodyfont = \normalfont,
qed=$\square$
]{simpleQED}
\declaretheorem[name=Example, style=simpleQED]{example}

\declaretheoremstyle[
headfont=\normalfont\bfseries, 
bodyfont = \normalfont,
]{simple}
\newmdtheoremenv{algo}{Algorithm}

\nobibliography*

\newcommand{\hideText}[1]{}

\newcommand{\indicator}[1]{\mathbbm{1}\{#1\}}

\newcommand{\iid}{$i.i.d.$}

\DeclareMathOperator*{\argmin}{arg\,min}
\DeclareMathOperator*{\argmax}{arg\,max}

\DeclareMathOperator*{\stt}{s.t.}
\DeclareMathOperator*{\colonBreak}{\,:\,}

\DeclareMathOperator*{\E}{E}

\DeclareMathOperator*{\cov}{Cov}

\DeclareMathOperator{\sign}{sign}

\definecolor{deepBlue}{rgb}{0.2, .2, .7}
\definecolor{LightGrey}{rgb}{0.9, .9, .9}

\newcolumntype{g}{D{.}{.}{3}}

\newcommand{\cites}[1]{\citeauthor{#1}'s \citeyearpar{#1}}

\newcommand{\citeauthors}[1]{\citeauthor{#1}'s}

\newcommand{\setKeywords}[1]{
	~ \li \noindent\emph{Keywords}: #1
}

\newcommand{\li}{\\~\\}

\ifdefined\largePage
\fi

\newcommand{\n}{\nonumber\\}
\newcommand{\no}{\nonumber}

\newcommand{\aq}{\text{and}\quad}
\newcommand{\wq}{\text{where}\quad}

\newcommand{\q}[1]{
	\newcount\quadCount
	\quadCount=#1
	\loop
	\quad 
	\advance \quadCount -1
	\ifnum \quadCount>0	
	\repeat
}

\newcommand{\courier}[1]{{\fontfamily{pcr}\selectfont#1}}

\newcommand{\verti}[1]{
	{\vert #1 \vert}
}

\newcommand{\vertiiSingle}{\vert\kern-0.25ex\vert}

\newcommand{\vertii}[1]{
	{\vert\kern-0.25ex\vert #1 \vert\kern-0.25ex\vert}
}

\newcommand{\vertiii}[1]{
	{\vert\kern-0.25ex\vert\kern-0.25ex\vert #1 
		\vert\kern-0.25ex\vert\kern-0.25ex\vert}
}

\newcommand{\vertib}[1]{
	{\left\vert #1 \right\vert}
}
\newcommand{\vertiib}[1]{
	{\left\vert\kern-0.25ex\left\vert #1 
		\right\vert\kern-0.25ex\right\vert}
}
\newcommand{\vertiiib}[1]{
	{\left\vert\kern-0.25ex\left\vert\kern-0.25ex\left\vert #1 
		\right\vert\kern-0.25ex\right\vert\kern-0.25ex\right\vert}
}

\begin{document}
\author{Robert L. Bray}
\affil{Kellogg School of Management, Northwestern University}
\title{
	Logarithmic Regret in Multisecretary and \\  Online Linear Programs with Continuous Valuations
}
\maketitle

\begin{abstract}
	\singlespacing \noindent
	I study how the shadow prices of a linear program that allocates an endowment of $n\beta \in \mathbbm{R}^{m}$ resources to $n$ customers behave as $n \rightarrow \infty$. I show the shadow prices (i) adhere to a concentration of measure, (ii) converge to a multivariate normal under central-limit-theorem scaling, and (iii) have a variance that decreases like $\Theta(1/n)$. I use these results to prove that the expected regret in \cites{Li2019b} online linear program is $\Theta(\log n)$, both when the customer variable distribution is known upfront and must be learned on the fly. I thus tighten \citeauthors{Li2019b} upper bound from $O(\log n \log \log n)$ to $O(\log n)$, and extend \cites{Lueker1995} $\Omega(\log n)$ lower bound to the multi-dimensional setting. I illustrate my new techniques with a simple analysis of \cites{Arlotto2019} multisecretary problem. 
	\setKeywords{online linear program; multisecretary problem; network revenue management; dual convergence; regret bounds; empirical process}
\end{abstract}

\section{Introduction}

\cite{Caley1875} introduced the secretary problem in the nineteenth century. The problem is to hire a man to serve as your secretary (a man because most secretaries were men back then). There are $ n $ applicants for the position whom you interview sequentially. But there's a hitch: once you interview a man, you must decide whether or not to hire him before interviewing the next man. So you face an optimal stopping problem, with the objective being to maximize the expected capability of the man you hire or, equivalently, to minimize the expectation of your \emph{regret}, the capability difference between the most competent man and the man you hire.

\cite{Arlotto2019} studied the \emph{multisecretary} problem, which is the same as above except with $ n \beta $ posts to fill, for some $ \beta \in [0, 1] $. In this version of the problem, your regret is the expected capability difference between the $ n \beta $ most capable men and the $ n \beta $ men you hire. \citeauthor{Arlotto2019} made a striking discovery: If secretary valuations are \iid\ random variables with finite support, then your expected regret is uniformly bounded across $ n \in \mathbbm{N} $ and $ \beta \in [0, 1] $. In other words, you're never expected to make more than some finite number of hiring mistakes, regardless of the number of positions you must fill. For this breakthrough result, \cite{Arlotto2019} won the 2021 Applied Probability Society Best Publication Award.

Citing an example by Robert Kleinberg, however, \citet[p. 234, 251]{Arlotto2019} showed that the regret could be large if one valuation had a small probability mass, which suggests that ``one cannot generally expect a bound that does not depend on the minimal mass." Nevertheless, they concluded their article by explaining ``At this point, it is not clear whether bounded regret is achievable also with continuous [valuation] distributions." 

In Section \ref{s:bound}, I show that bounded regret is not achievable with continuous valuation distributions. Specifically, I show that if secretary valuations are \iid\ uniform random variables then the expected regret lies between $(\beta / 8)(1-\beta / 8)(\log(n)/2 - \log(6))$ and $(\log(n+1) + 7)/8$ for all $n \ge 2^{20}\beta^{-8}$ and $\beta \in [0, 1/2]$. Moreover, I show that the most obvious heuristic satisfies the upper bound. 

Switching from finite to continuous valuations completely changes the mechanics of the model. With finite valuations, the probability of making a period-$ t $ hiring mistake decreases exponentially in $ t $, whereas the expected cost of such a mistake remains constant. Hence the total regret grows with $ n $ like $ \sum_{t=1}^{n} \exp(-t) = \Theta(1)$. With continuous valuations, however, the probability of making a period-$ t $ hiring mistake and the expected cost of such a mistake both decrease like $ 1/\sqrt{t} $. Hence the total cost grows with $ n $ like $ \sum_{t=1}^{n} (1/\sqrt{t})\cdot (1/\sqrt{t}) = \Theta(\log n)$.

In Section \ref{s:rmp}, I extend this logic and $ \Theta(\log n) $ regret rate to \cites{Li2019b} more general online linear programming" (OLP) problem. In this problem, you start with inventory vector $ n\beta \in \mathbbm{R}_{+}^{m} $, and you exchange inventory $ a_{t} \in \mathbbm{R}_{+}^{m} $ for utility $ u_{t} \ge 0$ if you fulfill the period-$ t $ customer's demand. Since none of your stocks can become negative, you must carefully husband each of your $ m $ resources. But doing so is difficult, as you have no foreknowledge of the nature of demand; instead, you must learn the demand distribution the old-fashioned way---by serving customers. 

The engine underlying my analysis of the online linear program is a set of shadow price convergence results I develop in Section \ref{s:convergence}. \citet[p, 2952]{Li2019b} lamented that ``there is still a lack of theoretical understanding of the properties of the dual optimal solutions," so I begin by characterizing their limiting behavior. I show that an online linear program's shadow prices (i) conform to a concentration of measure, (ii) converge to a multivariate normal under CLT-like scaling, and (iii) have a covariance matrix whose elements fall like $\Theta(1/t)$. I derive these results with a new approach that hems in the shadow prices with empirical processes.  

\section{Related Works}\label{s:lit}

\subsection{Primary Antecedents}
I thought I was the first to resolve \cites{Arlotto2019} question as to whether bounded regret is achievable with continuous valuation distributions. Unfortunately, after two rounds of review, I became aware of \cites{Lueker1995} article, which answered the question before \citeauthor{Arlotto2019} posed it. (My understanding is that Sid Banerjee found the reference.) \citeauthor{Lueker1995} study the ``zero-one knapsack problem," in which you successively decide whether to add a given object to your backpack. He proves the expected regret grows like $\Theta(\log n)$ under the optimal policy. 

And he establishes this bound with proof unlike any other I have found in the literature. Specifically, he constructs lower and upper envelopes across the entire surface of the offline and online value functions. The induction required to creating these bounds was painstaking because he had to weaken them just so as the inventory level departed the initial resource endowment. In contrast, I use \cites{Veraa} simpler \emph{compensated coupling} scheme, which sums up the myopic regret incurred over the random walk traversed by the inventory level under the online policy.\endnote{
	I got the idea for compensated coupling from \cite{Arcidiacono2011}, which used a similar technique to structurally estimate dynamic programs.
} 

As \citeauthors{Lueker1995} specification generalizes the multisecretary model, it also generalizes \cites{Arlotto2020} stochastic knapsack problem, which is the multisecretary's mirror image: one has continuous valuations and fixed capacity consumption, and the other has fixed valuations and continuous capacity consumption. \citet[p. 190]{Arlotto2020} develop an $ O(\log n) $ upper bound for the regret in their model, but they do not develop a corresponding lower bound, since ``it is well known that the optimal policy often lacks desirable structural properties, so proving [this lower bound] is unlikely to be easy." 

\cite{Jasin2014} extend the $O(\log n)$ upper regret bound to the multi-variate setting. Incorporating multiple dimensions enables inventories and shadow prices to veer off course in an \emph{uncoutable} number of ways. Indeed, whereas \citeauthor{Lueker1995} has only to prevent these variables from being too high or too low, \citeauthor{Jasin2014} has to prevent them from following any unit vector too far. However, \citeauthor{Jasin2014} only supports a finite number of consumption bundles, as he considers the network revenue management problem in which you price a set number of products (e.g., flight itineraries), each of which comprises a set number of resources (e.g., flight legs).

\cite{Li2019b} relax the finite-product assumption, allowing a given customer's resource consumption to be any number in a bounded region of $\mathbbm{R}^{m}$. And, more importantly, they incorporate \emph{online learning}, as they suppose you don't know the demand distribution upfront. Not knowing this distribution induces a complex correlation between inventory levels and shadow price estimates. \citeauthor{Li2019b} nevertheless managed to establish an $O(\log n \log \log n)$ upper regret bound.

Now, if you compare this $O(\log n \log \log n)$ bound with the prior $O(\log n)$ bounds, you can't help but wonder: Is revenue management with and without learning in the same class of difficulty? I show that they are. Specifically, I establish that the regret is $O(\log n)$ when the demand distribution is unknown and $\Omega(\log n)$ when it is known. 

Removing the $\log \log n$ fudge factor from \citeauthors{Li2019b} upper bound requires (i) more sharply characterizing the limiting behavior of the shadow prices and (ii) more tightly controlling the inventory process. Whereas \citeauthor{Li2019b} show that the magnitude of the period-$t$ shadow price covariance matrix is $O((\log \log t)/t)$, I show that it is $\Theta(1/t)$. And whereas they constrain inventories for all but the last $O(\log n \log \log n)$ periods, I constrain them for all but the last $O(1)$ periods. New methodological innovations underpin both improvements.

First, I sharpen the shadow price asymptotics by applying empirical process techniques to the subgradient of the dual linear program. Casting this subgradient as an empirical process enables me to create shadow price convergence results that hold uniformly in the inventory level. And this, in turn, allows me to overcome the hopeless entanglement between the current inventory level and the current shadow price estimate.

Second, I create new techniques to constrain the inventory level's random walk. For the upper bound with a known demand distribution, I control the process with a standard martingale concentration inequality. For the upper bound with an unknown demand distribution, I split the process into martingale and drift parts, and I then apply the martingale concentration inequality to the former and inductively bound the latter, showing that the inventory level being ``in control" up until period $t+1$ implies that the period-$(t+1)$ shadow price is ``in control," which in turn implies that the period-$t$ inventory level is ``in control." (This induction wouldn't have been possible without the empirical process' uniform bounds.) Finally, for the lower bound, I regulate the probability of the inventory levels spiraling out of control with the cost of splitting the offline linear program into two separate linear programs. For example, suppose you have 1,000 applicants for 100 secretarial positions and can interview all the men upfront; now, suppose I told you that you can only hire ten of the last 500 candidates. This additional constraint will substantially decrease the value of your hires, with high probability. And your regret conditional on having fewer than ten open positions with 500 remaining applicants is at least as large as the cost imposed by this offline constraint. Hence, the probability of having such a low inventory level must be sufficiently small, or the optimal policy would violate the $O(\log n)$ regret upper bound.

Before discussing the rest of the literature, let me recapitulate this work's primary contributions relative to what came before.
\begin{itemize}
	\item I extend \cites{Lueker1995} $\Theta(\log n)$ regret convergence rate to a multivariate setting, both with and without online learning. Neither generalization is trivial (see the discussion at the end of Section \ref{s:model}).
	\item I show that we can tightly regulate an online linear program's shadow prices by casting the subgradient of the dual value function as an empirical process.
	\item I use this newfound control over shadow prices to establish their asymptotic normality, concentration of measure, and $\Theta(1/t)$ variance. (N.B., the last result does not stem from the first result because convergence in distribution does not imply convergence of variance.)
	\item I create new methods for constraining the inventory process under the online policy. 
\end{itemize}

\subsection{Recent Developments}\label{s:Recent}

I will now discuss some noteworthy advancements that emerged since I first circulated my results. First, \cite{Balseiro2023a} have produced an insightful and comprehensive survey article that organizes models corresponding to ``dynamic pricing with capacity constraints, dynamic bidding with budgets, network revenue management, online matching, and order fulfillment" under a unified umbrella, ``dynamic resource-constrained reward collection ($\text{DRC}^{2}$) problems." The $\text{DRC}^{2}$ framework is similar to \cites{Vera} ``online resource allocation" framework, except it can accommodate an infinite number of customer types. Accordingly, whereas the online linear program cannot be positioned in \citeauthors{Vera} framework, it can be positioned in \cites{Balseiro2023a} framework (by making utilities bounded and the demand distribution known). \citeauthor{Balseiro2023a} explain that their class of problems is especially amenable to the ``\emph{certainty-equivalent principle:} replace quantities by their expected values and take the best actions given the current history." And, indeed, \cite{Li2019b} and I employ this approach to upper bound the regret of the online linear program.

An anonymous reviewer brought \citeauthor{Balseiro2023a}'s article to my attention, highlighting that we both independently devised how to exploit the certainty equivalent heuristic's martingale inventories:
\begin{quote}
	I want to note that Balseiro, Besbes, and Pizarro (2021) were written after the initial draft of [Bray's] paper. The certainty equivalent (CE) heuristic studied seems like a close cousin of [Bray's] ``martingale policy." Specifically, Balseiro, Besbes, and Pizarro (2021) show that under similar assumptions, the resource vector under the CE heuristic evolves like a martingale until it exits some region. ...  Both heuristics seem to achieve a Log(T) regret under seemingly similar conditions (the author may want to comment if there are key differences). I want to emphasize that I do not think Balseiro, Besbes, and Pizarro (2021) diminishes the contribution of this paper because a) it is clear that the two papers are developed independent at around the same time, and b) the main contribution of this paper is the lower bound. In fact, Balseiro et al.'s work for a similar class of problems further highlights that a general lower bound is significant to the literature.		
\end{quote}

Next, \cite{Wang2022} establish an $\Omega(\log n)$ gap between the expected online value and the fluid approximation value (as opposed to the expected offline value) in \cites{Jasin2014} network revenue management problem. However, they only establish this result for the one-dimensional version of the problem. For the multi-dimensional version, they show that the optimal policy yields only $O(1)$ more expected value than the policy \citeauthor{Jasin2014} used.

\cite{Besbes2022} point out that a multisecretary problem's $O(\log n)$ regret critically hinges on the probability density function being bounded away from zero near the acceptance-rejection threshold. Akshit Kumar explained it to me like this: If you have $n$ applicants for $n\beta$ open positions, then the marginal applicant would have a valuation of $F^{-1}(1-\beta + O_{p}(\sqrt{n}))$, where $F$ is the utility CDF. Now, if the utility PDF equals $f(u) = \verti{u - u^{*}}$ in a neighborhood of $u^{*} \equiv F^{-1}(1-\beta)$ then we would have $F(u) = 1 - \beta + \text{sign}(u - u^{*})(u - u^{*})^{2}/2$ and hence $F^{-1}(q) = u^{*} + \text{sign}(q - 1 + \beta) \sqrt{2 \verti{q - 1 + \beta}}$. And in this case, the expected myopic regret would exceed $1/n$ because rather than the usual $n^{-1/2}$ tolerance, we could now only discern the marginal man's utility to within a $n^{-1/4}$ tolerance: $F^{-1}(1-\beta + O_{p}(\sqrt{n})) = u^{*} + \text{sign}(O_{p}(\sqrt{n})) \sqrt{2 \verti{O_{p}(\sqrt{n})}} = u^{*} + O_{p}(n^{-1/4})$. To avoid these low-density regions, \citeauthor{Besbes2022} create a version of \cites{Balseiro2023a} certainty-equivalent principle that is ``conservative with respect to gaps." Their algorithm steers the inventory random walk away from regions with high expected myopic regret.

Finally, \cite{Jiang} independently developed an $O(1/n)$ bound for the shadow price variance (I posted the result in January 2022, and they posted it in October). They combine this dual convergence result with a technique that's similar in spirit to \citeauthors{Besbes2022} ``conservative with respect to gaps" to establish an $O(\log^{2} n)$ regret bound for the network revenue management problem without imposing a non-degenerate fluid limit. In contrast, previous models have assumed the fluid approximation's constraints bind with pressure or are slack, with additional leeway. Assuming extra wiggle room in the fluid model is unreasonable, however, as it implies that some buffer stocks scale \emph{linearly} with demand, which is a way over investment since safety stocks ought to scale with the square root of sales.

\section{Multisecretary Problem}\label{s:bound}

I will begin with the simple multi-secretary problem to demonstrate my regret-bounding approach. \cite{Lueker1995} has already established that this model's regret scales like $\Theta(\log n)$, but I will provide a far simpler proof, and my bounds won't have any hidden constants. 

\subsection{Setup}\label{s:boundSetup}

You have $n \in \mathbbm{N}$ applicants for $n\beta \in \mathbbm{N}$ positions, where $\beta \in [0, 1/2]$. (It suffices to consider $\beta \in [0, 1/2]$, because the the expected regret with $n\beta$ initial open slots equals that with $n(1-\beta)$ initial open slots.\endnote{\label{en:symmetry}
	To see that the expected regret with $n\beta$ initial open slots equals that with $n(1-\beta)$ initial open slots, note that we can re-express the problem of maximizing the capability of each of the $n\beta$ men you hire to maximizing one minus the capability of the $n (1-\beta)$ men you reject. But one minus a uniform is also a uniform, so this mirror image problem must yield mirror image regrets. 
}) You interview the candidates sequentially, starting with the $ n $th man and ending with the first man so that the applicant number corresponds with the size of the remaining candidate pool. Interviewing the $ t $th man reveals the utility you would get from hiring him, $ u_{t} $, a standard uniform random variable independent of the other candidates' utilities. After interviewing this candidate, you must hire him on the spot or reject him for good. You seek to maximize the expected total utility from your hires. Characterizing this utility will take a few steps.

First, let $v_{t}^{b}$ denote the utility you receive starting from period $t$ with $tb \in \mathbbm{N}$ open positions. The expectation of this variable satisfies the following Bellman equations:
\begin{align*}
	\E(v_{t}^{b}) \equiv & \E\Big(\max_{x_{t} \in \{0, 1\}} x_{t} u_{t} + \E(v_{t-1}^{\psi_{t}^{b}(x_{t})}) \quad \stt \quad x_{t} \le tb\Big), \\
	\E(v_{0}^{b}) \equiv & 0, \\
	\aq \psi_{t}^{b}(a) \equiv & 
	\begin{cases}
		(t b - a) / (t-1) & t > 1 , \\
		0 & t = 1 .
	\end{cases}
\end{align*}
I will explain the logic underlying these equations after line \eqref{eq:supermodular}. But first note that the $\psi_{t}^{b}$ function maps the fraction of men you can hire from period $t$ onwards, $b$, and your period-$t$ hiring decision, $x_{t}$, to the fraction of men you can hire from period-$(t-1)$ onwards. For example, if the period-$t$ superscript is $b$ and you hire the period-$t$ applicant---i.e., set $x_{t} = 1$---then the period-$(t-1)$ superscript is $\underbrace{tb-1}_{\text{positions left}} / \underbrace{(t - 1)}_{\text{applicants left}}$.

The Bellman equations above specify the following optimal action:
\begin{align}
	\pi_{t}^{b} \equiv \argmax_{x_{t} \in \{0, 1\}} x_{t} u_{t} + \E(v_{t-1}^{\psi_{t}^{b}(x_{t})}) \quad \stt \quad x_{t} \le tb .\label{eq:supermodular}
\end{align}
The $x_{t} \le tb$ constraint ensures that you don't extend a job offer if you don't have any positions available---i.e., that you set $x_{t} = 0$ if $tb = 0$. The expression above states that you hire the period-$t$ man (i.e., set $x_{t} = 1$) if you have a job opening (i.e., $1 \le tb$) and if the total expected utility conditional on hiring him (i.e., $u_{t} + \E(v_{t-1}^{\psi_{t}^{b}(1)})$) exceeds the total expected utility conditional on rejecting him (i.e., $\E(v_{t-1}^{\psi_{t}^{b}(0)})$). 

Your corresponding realized value is 
\begin{align*}
	v_{t}^{b} \equiv & \pi_{t}^{b} u_{t} + v_{t-1}^{\psi_{t}^{b}(\pi_{t}^{b})} \quad \text{and} \quad v_{0}^{b} \equiv 0 .
\end{align*}
Hence, you garner value $v_{n}^{\beta}$ from your $n$ applicants and $n\beta$ positions under the expected-utility-maximizing policy. But if you could have interviewed every applicant before extending any job offers, then you would have garnered value $V_{n}^{\beta}$, where 
\begin{align*}
	V_{t}^{b} & \equiv \sum_{s=1}^{tb} h_{t}^{s},
\end{align*}
and $ h_{t}^{s} $ is the $ s $th highest value in $ \{u_{t}, \cdots, u_{1}\} $. Since the utilities follow a uniform distribution, order statistic $ h_{t}^{s} $ follows a beta$ (t-s+1, s) $ distribution.

The difference between the aggregate utility received in the offline problem and that received in the online problem is your regret: 
\begin{align*}
	R_{n} \equiv V_{n}^{\beta} - v_{n}^{\beta}.
\end{align*}
The following two propositions bound the expectation of this random variable.
\begin{proposition}\label{p:prop1Multisecretary}
	The optimal policy of the multisecretary problem yields an expected regret that grows at no more than a $\log n$ rate: $\E(R_{n}) \le (\log(n+1) + 7)/8$, for all $n \in \mathbbm{N}$ and $\beta \in [0, 1/2]$.
\end{proposition}
\begin{proposition}\label{p:prop2Multisecretary}
	The optimal policy of the multisecretary problem yields an expected regret that grows at no less than a $\log n$ rate: $\E(R_{n}) \ge (\beta / 8)(1-\beta / 8)(\log(n)/2 - \log(6))$, for all $n \ge 2^{20}\beta^{-8}$ and $\beta \in [0, 1/2]$.
\end{proposition}

N.B., these theorems provide non-asymptotic results---i.e., do not rely on big-O notation. Proposition \ref{p:prop1Multisecretary}'s finite-sample bound is especially interesting, as it highlights the near worthlessness of the value of future information. For example, suppose you have a billion applicants for 500 million jobs. In this case, your online value would be around $\underbrace{(1/2 + 1)/2}_{\text{value of average hire}} \cdot \underbrace{500 \text{ million}}_{\text{number of hires}} =  375 \text{ million}$ and your offline value would exceed your online value by around $(\log(10^9+1) + 7)/8 = 3.47$. Hence, knowing the billion worker utilities upfront increases your workforce's value by around $3.47 / 375 \text{ million} = .00000093\%$.

\subsection{Upper Bound}\label{s:upperEasy}

I will now prove Proposition \ref{p:prop1Multisecretary} by showing that Algorithm \ref{alg:upperMultisecretary} honors its bound. The proof has two parts: the first decomposes the total regret into a sum of ``myopic regrets," in the fashion of \cite{Veraa}, and the second shows that the expectation of the period $t$ myopic regret is $O(1/t)$ under the myopic-regret-minimizing Algorithm \ref{alg:upperMultisecretary}, and hence that the expected total regret is $O(\sum_{t=1}^{n} 1/t) = O(\log n)$.

To derive the policy underling Algorithm \ref{alg:upperMultisecretary}, suppose that you hire the $t$th man with $tb$ available positions if and only if his valuation exceeds $\tau_{t}^{b}$, where $\{\tau_{t}^{b} \colonBreak t \in [n],\ tb \in \{0,\cdots, t\} \}$ is a collection of thresholds that have yet to be defined. These thresholds will satisfy $\tau_{t}^{0} =  1$ and $\tau_{t}^{1} = 0$ for all $t \in [n]$, to ensure that $b_{t} \in [0, 1]$ for all $t \in [n]$, where 
\begin{align*}
	b_{n} & \equiv \beta\\
	\aq b_{t-1} & \equiv  \psi_{t}^{b_{t}}(\indicator{u_{t} > \tau_{t}^{b_{t}}}).
\end{align*}
In other words, you start period $t$ with $tb_{t} \in \{0, \cdots, n\}$ open positions under the threshold policy. And you receive corresponding value $\bar{v}_{n}$, where 
\begin{align*}
	\bar{v}_{t} &\equiv \indicator{u_{t} > \tau_{t}^{b_{t}}}u_{t} + \bar{v}_{t-1},\\
	\aq \bar{v}_{0} & \equiv 0. \no
\end{align*}
Since $\E(\bar{v}_{n}) \le \E(v_{n}^{\beta})$, it follows that 
\begin{align*}
	\E(\bar{R}_{n}) \ge & \E(R_{n}),\\
	\wq \bar{R}_{t} \equiv & V_{t}^{b_{t}} - \bar{v}_{t} .\no
\end{align*}
Accordingly, it will suffice to upper bound $ \E(\bar{R}_{n}) $. To this end, first note that the offline value function satisfies the following recurrence relations:
\begin{align}
	V_{t}^{b} & = (u_{t} - h_{t-1}^{tb})^{+} + V_{t-1}^{\psi_{t}^{b}(0)} \label{eq:VhVkuhV} \\
	\aq V_{t-1}^{\psi_{t}^{b}(0)} & = h_{t-1}^{tb} + V_{t-1}^{\psi_{t}^{b}(1)} . \label{eq:VhVk}
\end{align}
Line \eqref{eq:VhVkuhV} states that if there are $tb$ open positions then the value of increasing the applicant pool from $t - 1$ to $t$ applicants equals the option value of replacing the $tb$th most capable man, out of the first $t - 1$ applicants, with the $t$th man. And line \eqref{eq:VhVk} states that if there are $t - 1$ remaining applicants then the value of increasing the number of job openings from $(t-1) \psi_{t}^{b}(1) = tb - 1$ to $(t-1) \psi_{t}^{b}(0) = tb$ positions equals the value of the $tb$th best man out of the $t - 1$ candidates.

\begin{algo}\label{alg:upperMultisecretary}
	~\\ \vspace{-25pt} 
	\begin{enumerate}
		\item \courier{input $n$, $\beta$, $\{u_{t}\}_{t = 1}^{n}$}, 
		\item \courier{initialize $b_{n} \coloneqq \beta$}
		\item \courier{for $t$ from $n$ to $1$ do}
		\begin{enumerate}
			\item \courier{set $x_{t} \coloneqq \indicator{u_{t} > 1 - b_{t}} $}
			\item \courier{set $b_{t-1} \coloneqq \psi_{t}^{b_{t}}(x_{t})$}
		\end{enumerate}
		\item \courier{end for}
		\item \courier{output $\{x_{t}\}_{t=1}^{n}$}
	\end{enumerate}
\end{algo}

Now suppose that $u_{t} \le \tau_{t}^{b_{t}}$, and hence that $b_{t-1} = \psi_{t}^{b_{t}}(0)$ and $\bar{v}_{t} = \bar{v}_{t-1}$. In this case, \eqref{eq:VhVkuhV} implies that
\begin{align*}
	\bar{R}_{t} & = V_{t}^{b_{t}} - \bar{v}_{t} \\
	& =  (u_{t} - h_{t-1}^{tb})^{+} + V_{t-1}^{\psi_{t}^{b}(0)} - \bar{v}_{t-1} \\
	& =  (u_{t} - h_{t-1}^{tb})^{+} + V_{t-1}^{b_{t-1}} - \bar{v}_{t-1} \\
	& =  (u_{t} - h_{t-1}^{tb})^{+} + \bar{R}_{t-1}.
\end{align*}
Next, suppose that $u_{t} > \tau_{t}^{b_{t}}$, and hence that $b_{t-1} = \psi_{t}^{b_{t}}(1)$ and $\bar{v}_{t} = u_{t} + \bar{v}_{t-1}$. In this case, \eqref{eq:VhVkuhV} and \eqref{eq:VhVk} imply that 
\begin{align*}
	\bar{R}_{t} & = V_{t}^{b_{t}} - \bar{v}_{t} \\
	& =  (u_{t} - h_{t-1}^{tb})^{+} + V_{t-1}^{\psi_{t}^{b}(0)} - \bar{v}_{t} \\
	& =  (u_{t} - h_{t-1}^{tb})^{+} + (h_{t-1}^{tb} + V_{t-1}^{\psi_{t}^{b}(1)}) - (u_{t} + \bar{v}_{t-1}) \\
	& = (u_{t} - h_{t-1}^{tb})^{-} + \bar{R}_{t-1}.
\end{align*}
Combining these two recurrence relations inductively yields
\begin{align}
	\bar{R}_{n} & = \bar{r}_{n} + \bar{R}_{n-1} = \sum_{t = 1}^{n} \bar{r}_{t}, \label{eq:Dcomp} \\
	\wq \bar{r}_{t} & \equiv \indicator{u_{t} \le \tau_{t}^{b_{t}}}(u_{t} - h_{t-1}^{tb_{t}})^{+} + \indicator{u_{t} > \tau_{t}^{b_{t}}}(u_{t} - h_{t-1}^{tb_{t}})^{-}\n
	& = (\indicator{u_{t} > h_{t-1}^{tb_{t}}} - \indicator{u_{t} > \tau_{t}^{b_{t}}})(u_{t} - h_{t-1}^{tb_{t}}).\no
\end{align}
In the expression above, $\bar{r}_{t}$ is your ``myopic regret," which is the cost of your period-$t$ hiring mistake. Total regret can always be decomposed into a sum of myopic regrets.

Now here's the key: we can integrate over $u_{t}$ and $h_{t-1}^{tb}$ when taking the expectation of $\bar{r}_{t}$ because these variables are independent of each other and $b_{t}$. To integrate over $u_{t}$, we use the fact that this uniform random variable satisfies $\E((\indicator{u_{t} > h} - \indicator{u_{t} > \tau})(u_{t} - h)) = h^{2}/2 - h\tau + \tau^{2}/2$, for constants $h$ and $\tau$. And to integrate over $h_{t-1}^{tb}$, we use the fact that this beta$ (t-tb, tb) $ random variable satisfies $\E(h_{t-1}^{tb}) = 1-b$ and $\E(h_{t-1}^{tb})^{2} = \frac{(1- b) + t (1 - b)^{2}}{t+1}$. These properties enable us to express the expected myopic regret in terms of $b_{t}$ and $\tau_{t}^{b_{t}}$:
\begin{align}
	\E(\bar{r}_{t}) & = \E\big(\E\big((\indicator{u_{t} > h_{t-1}^{tb_{t}}} - \indicator{u_{t} > \tau_{t}^{b_{t}}})(u_{t} - h_{t-1}^{tb_{t}}) \colonBreak  h_{t-1}^{tb_{t}} = h,\ b_{t} = b\big)\big) \n
	& = \E\big(\E\big(((h_{t-1}^{tb_{t}})^{2}/2 - h_{t-1}^{tb_{t}} \tau_{t}^{b_{t}} + (\tau_{t}^{b_{t}})^{2}/2)\colonBreak b_{t} = b\big)\big) \n
	& = \E\Big(\frac{(1-b_{t}) + t (1 - b_{t})^{2}}{2(t+1)} - \tau_{t}^{b_{t}} (1-b_{t}) + (\tau_{t}^{b_{t}})^{2}/2\Big). \label{eq:toUseInLowerBound}
\end{align}
I will now minimize the expectation above by setting $\tau_{t}^{b} = 1-b$ (as specified by Algorithm \ref{alg:upperMultisecretary}), in which case the expression above simplifies to 
\begin{align*}
	\E(\bar{r}_{t}) & = \frac{\E(b_{t}(1-b_{t}))}{2(t+1)}.
\end{align*}
And, with this, we find that the regret incurred under Algorithm \ref{alg:upperMultisecretary} satisfies our logarithmic bound:
\begin{align}
	\E(R_{n}) \le & \E(\bar{R}_{n}) \n
	= & \sum_{t = 1}^{n} \E(\bar{r}_{t}) \n
	= & \sum_{t = 1}^{n} \frac{\E(b_{t}(1-b_{t}))}{2(t+1)} \n
	\le & \sum_{t = 1}^{n} \sup_{b \in (0, 1)} \frac{b(1-b)}{2(t+1)} \n
	= & \sum_{t = 1}^{n} \frac{1}{8 (t +1)} \n
	\le & (\log(n+1) + 7)/8.\no
\end{align}

\subsection{Lower Bound}\label{s:lowerMultisecretary}

I will now prove Proposition \ref{p:prop2Multisecretary}. The proof has four steps. The first creates an optimal-policy version of the regret decomposition derived in the last section. The decomposition is the same as before, except $b_{t}$ now denotes the number of open positions under the optimal algorithm rather than under Algorithm \ref{alg:upperMultisecretary}. The second part of the proof shows that $\Omega(\log n)$ expected regret follows immediately from the regret decomposition, provided that there's an $\Omega(1)$ chance of $b_{t}$ being bounded away from either endpoint. Finally, the third part of the proof bounds the chance of $b_{t}$ being too close to one, and the fourth part bounds the chance of it being too close to zero.

To begin the proof, note that the objective in \eqref{eq:supermodular} is supermodular in $x_{t}$ and $u_{t}$. Hence, Topkis's theorem implies that there exists threshold collection 
\begin{align}
	\{\tau_{t}^{b} \colonBreak t \in [n],\ tb \in \{0,\cdots, t\} \}, \label{eq:optThreshold}	
\end{align}
such that the optimal policy hires the $t$th man with $tb$ available positions if and only if $u_{t} > \tau_{t}^{b}$. And as before, these thresholds satisfy $\tau_{t}^{0} =  1$ and $\tau_{t}^{1} = 0$, since the optimal policy always makes exactly $n$ job offers. 

Now, since the optimal policy has a threshold structure, lines \eqref{eq:Dcomp} and \eqref{eq:toUseInLowerBound} imply that 
\begin{align}
	\E(R_{n}) & = \sum_{t = 1}^{n} \E\Big(\frac{(1-b_{t}) + t (1 - b_{t})^{2}}{2(t+1)} - \tau_{t}^{b_{t}} (1-b_{t}) + (\tau_{t}^{b_{t}})^{2}/2\Big) \n
	& \ge \sum_{t = 1}^{n} \E\Big(\min_{\tau}\Big(\frac{(1-b_{t}) + t (1 - b_{t})^{2}}{2(t+1)} - \tau (1-b_{t}) + \tau^{2}/2\Big)\Big)\n
	& = \sum_{t = 1}^{n} \frac{\E(b_{t}(1-b_{t}))}{2(t+1)}. \label{eq:decompMSLB}
\end{align}
Keep in mind that that $b_{t}$ now characterizes the number of open positions under the optimal thresholds defined in \eqref{eq:optThreshold}:
\begin{align}
	b_{n} & \equiv \beta\n
	\aq b_{t-1} & \equiv  \psi_{t}^{b_{t}}(\indicator{u_{t} > \tau_{t}^{b_{t}}}). \label{eq:processB}
\end{align}

Lower bounding expression \eqref{eq:decompMSLB} will require upper bounding the probability that $b_{t}$ veers too closely to either endpoint. For this, I will show that $n \ge 2^{20} \beta^{-8}$ and $\sqrt{n}\le t \le n/2$ imply
\begin{align}
	\Pr(b_{t} < \beta/8) & \ge \Pr(b_{t} > 1 - \beta/8)  \label{eq:probprob}\\
	\aq \Pr(b_{t} < \beta/8) & \le 1/4.\label{eq:probprob3}
\end{align}
Combining these bounds with line \eqref{eq:decompMSLB} yields Proposition \ref{p:prop2Multisecretary}:
\begin{align*}
	\E(R_{n}) & \ge \sum_{t = 1}^{n} \frac{\Pr(\beta/8\le b_{t} \le 1 - \beta/8) \beta/8 (1 - \beta/8)}{2(t+1)}\\
	& = \sum_{t = 1}^{n} \frac{\big(1 - \Pr(b_{t} < \beta/8) - \Pr(b_{t} > 1 - \beta/8)\big)\beta/8 (1 - \beta/8)}{2(t+1)}\\
	& \ge \sum_{t = \lceil \sqrt{n} \rceil}^{\lfloor n/2\rfloor} \frac{(1 - 1/4 - 1/4)\beta/8 (1 - \beta/8)}{2(t+1)}\\
	& \ge \int_{t = 2\sqrt{n}}^{n/3} (\beta / 8)(1-\beta / 8)/(8t) dt \\
	& = (\beta / 8)(1-\beta / 8)(\log(n)/2 - \log(6)).
\end{align*}

Accordingly, it will suffice to establish lines \eqref{eq:probprob} and \eqref{eq:probprob3}. I will begin with the former, because it is more straightforward. Simply put, the $\{b_{t}\}_{t = n}^{1}$ is process is more likely to approach the left endpoint than the right endpoint because it starts at $\beta \le 1/2$ and is symmetric about 1/2. 

I will now formalize this intuition with a coupling argument. First, note that the problem symmetry discussed in Endnote \ref{en:symmetry} implies that the acceptance thresholds satisfy
\begin{align}
	\tau_{t}^{b} = 1 - \tau_{t}^{1 - b}. \label{eq:probSymmetry}
\end{align}
Basically, this holds because one minus a uniform is also a uniform. Second, consider the following benchmark process:
\begin{align*}
	\hat{b}_{n} & \equiv 1 - \beta\\
	\aq \hat{b}_{t-1} & \equiv  \psi_{t}^{\hat{b}_{t}}(\indicator{u_{t} > \tau_{t}^{\hat{b}_{t}}}). 
\end{align*}
The $\{b_{t}\}_{t = n}^{1}$ and $\{\hat{b}_{t}\}_{t = n}^{1}$ processes can't jump over one another, because the number of open positions can only decrease by one or remain constant in a given period. And the processes couple whenever they meet, with $b_{t} = \hat{b}_{t}$ implying $b_{t-1} = \hat{b}_{t-1}$. Accordingly, $\hat{b}_{t} < \beta/8$ implies $b_{t} < \beta/8$, and hence $\Pr(\hat{b}_{t} < \beta/8) \le \Pr(b_{t} < \beta/8)$. Third, since one minus a uniform is also a uniform, the process $\{\hat{b}_{t}\}_{t = n}^{1}$ has the same distribution as the process $\{\tilde{b}_{t}\}_{t = n}^{1}$, where
\begin{align*}
	\tilde{b}_{n} & \equiv 1 - \beta\\
	\aq \tilde{b}_{t-1} & \equiv  \psi_{t}^{\tilde{b}_{t}}(\indicator{1 - u_{t} > \tau_{t}^{\tilde{b}_{t}}}).
\end{align*}
And with  \eqref{eq:probSymmetry} these equations can be rearranged like this:
\begin{align*}
	1 - \tilde{b}_{n} & = \beta \\
	\aq 1 - \tilde{b}_{t-1} & = 1 - \psi_{t}^{\tilde{b}_{t}}(\indicator{1 - u_{t} > \tau_{t}^{\tilde{b}_{t}}})\\
	& = 1 - \psi_{t}^{\tilde{b}_{t}}(\indicator{u_{t} < \tau_{t}^{1 - \tilde{b}_{t}}})\\
	& = \frac{t-1 - t\tilde{b}_{t} + \indicator{u_{t} < \tau_{t}^{1 - \tilde{b}_{t}}}}{t-1} \\
	& = \frac{t (1 - \tilde{b}_{t}) - \indicator{u_{t} \ge \tau_{t}^{1 - \tilde{b}_{t}}}}{t-1}\\
	& = \psi_{t}^{1 - \tilde{b}_{t}}(\indicator{u_{t} \ge \tau_{t}^{1 - \tilde{b}_{t}}}) .
\end{align*}
Compare this system to \eqref{eq:processB}, and you will see that $1 - \tilde{b}_{t} = b_{t}$, almost surely. Accordingly, $\Pr(b_{t} > 1 - \beta/8) = \Pr(\tilde{b}_{t} < \beta/8) = \Pr(\hat{b}_{t} < \beta/8) \le \Pr(b_{t} < \beta/8)$, which establishes \eqref{eq:probprob}.

Finally, I will establish \eqref{eq:probprob3}. The argument has three steps. First, I establish that the regret conditional on $b_{t} < \beta/8$ is at least as high as the value you'd get by replacing the the worst $\lfloor t \beta/8\rfloor$ men hired before period $t$ with the best $\lfloor t \beta/8\rfloor$ men rejected after period $t$, which is at least as high as $\lfloor t \beta/8\rfloor$ times the difference between the value of the $(tb_{t} + \lfloor t \beta/8\rfloor)th$ best man to arrive after $t$ and the $(n\beta - tb_{t} - \lfloor t \beta/8\rfloor + 1)$th best man to arrive before $t$. Second, I use the binomial Chernoff to establish that there's at least a $1 - 1/12 - 1/12 = 5/6$ chance that the $(tb_{t} + \lfloor t \beta/8\rfloor)th$ best man to arrive after $t$ is at least $\beta/2$ units better than the $(n\beta - tb_{t} - \lfloor t \beta/8\rfloor + 1)$th best man to arrive before $t$. Third, I use these results to show that the optimal policy would violate the $(\log(n+1) + 7)/8$ upper regret bound if the event $b_{t} < \beta/8$ were not sufficiently rare.

Now to begin the proof of line \eqref{eq:probprob3}, note that conditional on having $tb_{t}$ open positions at the start of period $t$, the best the online policy can do is hire the best $tb_{t}$ men out of the last $t$ applicants and hire the best $n\beta - tb_{t}$ men out of the first $n - t$ applicants. Thus, the online value must satisfy
\begin{align*}
	v_{n}^{\beta} & \le \sum_{s=1}^{tb_{t}} h_{t}^{s} + \sum_{s=1}^{n\beta - tb_{t}} \underleftarrow{h}_{t}^{s},
\end{align*}
where look-back order statistic $\underleftarrow{h}_{t}^{s}$ is the $s$th largest value in $ \{u_{n}, \cdots, u_{t+1}\} $ (i.e., it equals $h_{n - t}^{s}$, but with the order of the applicants reversed). Further, if $b_{t} < \beta/8$ then the offline policy could hire the best $tb_{t} + \lfloor t \beta/8\rfloor$ men out of the last $t$ applicants and the best $n\beta - tb_{t} - \lfloor t \beta/8\rfloor$ men out of the first $n - t$ applicants. Hence, the offline value must satisfy the following when $b_{t} < \beta/8$:
\begin{align*}
	V_{n}^{\beta} \ge \sum_{s=1}^{tb_{t} + \lfloor t \beta/8\rfloor} h_{t}^{s} + \sum_{s=1}^{n\beta - tb_{t} - \lfloor t \beta/8\rfloor} \underleftarrow{h}_{t}^{s}.
\end{align*}
Differencing the last two inequalities yields the following, for $b_{t} < \beta/8$:
\begin{align*}
	R_{n} & \ge \sum_{s=tb_{t}+1}^{tb_{t} + \lfloor t \beta/8\rfloor} h_{t}^{s} - \sum_{s=n\beta - tb_{t} - \lfloor t \beta/8\rfloor + 1}^{n\beta - tb_{t}} \underleftarrow{h}_{t}^{s} \n
	& \ge \lfloor t \beta/8\rfloor h_{t}^{tb_{t} + \lfloor t \beta/8\rfloor} - \lfloor t \beta/8\rfloor \underleftarrow{h}_{t}^{n\beta - tb_{t} - \lfloor t \beta/8\rfloor + 1} \n
	& \ge \lfloor t \beta/8\rfloor  (h_{t}^{\lfloor t \beta/4\rfloor} -  \underleftarrow{h}_{t}^{ n \beta-\lfloor t \beta/4 \rfloor}) \n
	& \ge \lfloor t \beta/8\rfloor  \indicator{h_{t}^{\lfloor t \beta/4\rfloor} \ge 1 - 3\beta /8} \indicator{\underleftarrow{h}_{t}^{n\beta-\lfloor t \beta/4 \rfloor} \le 1 - 7\beta/8}(7\beta/8 - 3\beta /8)\n
	& \ge  \lfloor t \beta^{2}/16\rfloor \indicator{h_{t}^{\lfloor t \beta/4\rfloor} \ge 1 - 3\beta /8} \indicator{\underleftarrow{h}_{t}^{n\beta-\lfloor t \beta/4 \rfloor } \le 1 - 7\beta/8}.
\end{align*}
The first line above states that your regret is at least as large as the benefit you'd get by replacing the worst $\lfloor t \beta/8\rfloor$ men hired before period $t$ with the best $\lfloor t \beta/8\rfloor$ men rejected after period $t$. The second line maintains that the value of this difference is at least as large as $\lfloor t \beta/8\rfloor$ (i.e., the number of men being exchanged) times the difference between $h_{t}^{tb_{t} + \lfloor t \beta/8\rfloor}$  (i.e., the value of the worst man added) and $\underleftarrow{h}_{t}^{n\beta - tb_{t} - \lfloor t \beta/8\rfloor + 1}$ (i.e., the value of the best man removed). The remaining three lines use the fact that $h_{t}^{s}$ decreases in its superscript to connect the bound with the following binomial Chernoff results: If $t \ge 48 \log(12) / \beta$, $n \ge 336 \log(12) / \beta$, and $\sqrt{n}\le t \le n/2$ then
\begin{align*}
	\Pr(h_{t}^{\lfloor t \beta/4\rfloor} \ge 1 - 3\beta /8) \ge 11/12  \\
	\aq \Pr(\underleftarrow{h}_{t}^{n\beta-\lfloor t \beta/4 \rfloor} \le 1 - 7\beta/8) \ge 11/12,
\end{align*}
Accordingly, Proposition \ref{p:prop1Multisecretary} and Bonferroni's inequality imply the following, for the specified range of $n$ and $t$:
\begin{align*}
	(\log&(n+1) + 7)/8 \\
	& \ge 
	\E(R_{n}) \\
	& \ge \lfloor t \beta^{2}/16\rfloor \Pr\big(b_{t} < \beta/8 \ \cap\ h_{t}^{\lfloor t \beta/4\rfloor} \ge 1 - 3\beta /8 \ \cap\  \underleftarrow{h}_{t}^{\lfloor (n-t)\beta\rfloor} \le 1 - 7\beta/8\big)\\
	& \ge \lfloor \sqrt{n} \beta^{2}/16\rfloor \big(\Pr(b_{t} < \beta/8) + 11/12 + 11/12 - 2\big).
\end{align*}
Finally, this inequality implies \eqref{eq:probprob3} when $n \ge 2^{20} \beta^{-8}$ and $\sqrt{n}\le t \le n/2$. 

\section{Online Linear Programming Problem} \label{s:rmp}

\subsection{Model} \label{s:model}

I will now extend the techniques developed in the last section to \cites{Li2019b} online linear program.\endnote{\label{en:threeMinor}
	I make three minor changes to the online linear programming model: I impose additional non-negativity constraints, $u_{1}, a_{1} \ge 0$, I do not include constraints that are slack in the limit, and I use a cleaner version of the continuous value assumption, which I inherited from \cite{Lueker1995}. The first two modifications are trivial: Accommodating negative $u_{1}$ and $a_{1}$ would be simple because all that matters is the difference, $\Delta_{1}(y) = u_{1} - a_{1}'y$. And a simple concentration of measure argument establishes that a constraint that does not bind in the limit has only a $O(1)$ effect on the expected regret because the probability of it binding decreases exponentially fast in $n$. (I incorporated constraints that are slack in the limit in a previous version of the manuscript.) The third change is more noteworthy, because Assumption \ref{a:marginalProb} is more straightforward and flexible. For example, this assumption permits unbounded shadow prices and hence unbounded utilities, and it extends the model to cover \cites{Arlotto2020} specification.
} 
See the appendix for a notation guide and the online supplement for the proofs. 

As before, I will count backward from period $ n \in \mathbbm{N} $ to period $ 1 $. In each period, a customer arrives, and you must decide whether or not to fulfill their demand from your inventory. You begin in period $ n $ with initial inventory endowment $ n b_{n} = n \beta $, for some given $\beta \in \mathbbm{R}_{+}^{m} $, so that you have $ e_{j}'b_{n} $ units of the $ j $th resource budgeted for the ``average" remaining period, where $e_{j}$ is the unit vector indicating the $j$th position. If you satisfy the period--$ n $ customer then you exchange inventory bundle $ a_{n} \in \mathbbm{R}_{+}^{m} $ for utility $ u_{n} $, so that you begin period $ n - 1 $ with resource vector $ b_{n-1} \equiv (n b_{n} - a_{n}) / (n - 1) $. If, on the other hand, you reject the period--$ n $ customer, then you receive no utility and lose no resources, so that you begin period $ n - 1 $ with resource vector $ b_{n-1} \equiv n b_{n} / (n-1) $. And this pattern repeats so that $b_{t -1} \equiv (t b_{t} - a_{t}) / (t - 1)$ if you satisfy the period--$t$ customer and $b_{t -1} \equiv (t b_{t}) / (t - 1)$ otherwise. The problem is dynamic because you don't observe variables $ u_{t} $ and $ a_{t} $ until the beginning of period $ t $. These variables satisfy the following assumptions:

\begin{assumption}\label{a:distribution}
	The customers are \iid: vectors $ \{(u_{t}, a_{t})\}_{t = 1}^{n} $ are drawn independently of one another, from joint distribution $\mu$.
\end{assumption}

\begin{assumption}\label{a:positiveUandA}
	The utilities and resource requirements are non-negative: $ u_{1}, a_{1} \ge 0 $ almost surely.
\end{assumption}

\begin{assumption}\label{a:boundedUtil}
	The utilities have finite expectation: $ \E(u_{1}) < \infty $.
\end{assumption}

\begin{assumption}\label{a:boundedA}
	The resource requirements are bounded: $ a_{1} \le \alpha $, almost surely, for some $ \alpha \in \mathbbm{R}_{+}^{m} $.
\end{assumption}

N.B. that $u_{1}$ can have unbounded support, whereas the other models cited in Section \ref{s:lit}---most notably those of \cite{Lueker1995} and \cite{Li2019b}---restrict $u_{1}$ to a finite range. 

Let $v_{t}^{b}$ denote the utility you receive from period $t$ onwards when you begin that period with resource endowment $tb \in \mathbbm{R}^{m}$. Since you follow the expected-utility-maximizing policy, this variable's expectation satisfies the following Bellman equations:
\begin{align}
	\E(v_{t}^{b}) \equiv & \E\Big(\max_{x_{t} \in \{0, 1\}} x_{t} u_{t} + \E(v_{t-1}^{\psi_{t}^{b}(x_{t}a_{t})}) \quad \stt \quad x_{t}a_{t} \le tb\Big),  \label{eq:firstBellmanEq}\\
	\E(v_{0}^{b}) \equiv & 0, \n
	\aq \psi_{t}^{b}(a) \equiv & 
	\begin{cases}
		(t b - a) / (t-1) & t > 1 , \\
		0 & t = 1 .
	\end{cases} \label{eq:firstBellmanEq2}
\end{align}
To better understand this system, consider the following optimal action:
\begin{align}
	\pi_{t}^{b} \equiv \argmax_{x_{t} \in \{0, 1\}} x_{t} u_{t} + \E(v_{t-1}^{\psi_{t}^{b}(x_{t}a_{t})}) \quad \stt \quad x_{t}a_{t} \le tb .\label{eq:optimalAction}
\end{align}
In other words, you accept the period-$t$ customer (i.e., set $x_{t} = 1$) if you have inventory enough to do so (i.e., $a_{t} \le tb$) and if the total expected utility conditional on satisfying this customer (i.e., $u_{t} + \E(v_{t-1}^{\psi_{t}^{b}(1)})$) exceeds the total expected utility conditional on turning them away (i.e., $\E(v_{t-1}^{\psi_{t}^{b}(0)})$).

Under this policy you garner total value $ v_{n}^{\beta} $ from your initial $n \beta $ resource endowment, where
\begin{align}
	v_{t}^{b} \equiv & \pi_{t}^{b} u_{t} + v_{t-1}^{\psi_{t}^{b}(\pi_{t}^{b}a_{t})} \quad \text{and} \quad v_{0}^{b} \equiv 0 .\label{eq:defOfLittleV}
\end{align}
However, if you could have observed all of the customer attributes before deciding which ones to satisfy, then you would have garnered value $V_{n}^{\beta}$, where 
\begin{align}
	V_{t}^{b} \equiv & \max_{x \in \{0, 1\}^t} \sum_{s=1}^{t} x_{s}u_{s} \quad \stt \quad \sum_{s=1}^{t} x_{s}a_{s} \le tb . \label{eq:IP}
\end{align}

Your regret is the difference between the utility you extract when you observe all customer variables upfront and the utility you extract when you learn these variables on the fly:
\begin{align}
	R_{n} \equiv & V_{n}^{\beta} - v_{n}^{\beta} \label{eq:regretToBound}.
\end{align}
Our objective is to show that $ \E(R_{n}) = \Theta(\log n) $ as $ n \rightarrow \infty $. 

Since expanding your choice set from $ \{0, 1\} $ to $ [0, 1] $ will not make you worse off, we have
\begin{align}
	\bar{V}_{n}^{\beta} \ge & V_{n}^{\beta}, \n
	\wq \bar{V}_{t}^{b} \equiv & \max_{x \in [0, 1]^t} \sum_{s=1}^{t} x_{s}u_{s} \quad \stt \quad \sum_{s=1}^{t} x_{s} a_{s} \le tb \label{eq:primalProgram} \\
	= & \min_{y \in \mathbbm{R}_{+}^{m}} t\Lambda_{t}^{b}(y), \label{eq:ConvexProblem}\\
	\Lambda_{t}^{b}(y) \equiv & b'y + \sum_{s = 1}^{t}\Delta_{s}(y)^{+}/t, \n
	\aq \Delta_{t}(y) \equiv & u_{t} - a_{t}'y .\label{eq:defOfDelta}
\end{align}
Line \eqref{eq:ConvexProblem} is a reconfiguration of the dual of line \eqref{eq:primalProgram}'s linear program \cite[see][]{Li2019b}. (To remember that it is a dual, it helps to think of $\Lambda$ as an upside down $V$.) This dual problem has a not-necessarily-unique shadow price minimizer:
\begin{align}
	y_{t}^{b} \in \argmin_{y \in \mathbbm{R}_{+}^{m}} t\Lambda_{t}^{b}(y).\label{eq:notnecessarilyunique}
\end{align}

Since we initialized $b_{n} = \beta$, the problem in \eqref{eq:ConvexProblem} converges, as $n \rightarrow \infty$, to the following deterministic problem:
\begin{align}
	\min_{y \in \mathbbm{R}_{+}^{m}} \Lambda_{\infty}^{\beta}(y) \quad \text{where} \quad
	\Lambda_{\infty}^{b}(y) \equiv b'y + \E(\Delta_{1}(y)^{+}). \label{eq:infinityLambdaDef}
\end{align}
The following assumption endows this problem with a positive shadow price solution.

\begin{assumption}\label{a:PositiveShadowPrice}
	All resources are consumed in the limit: there exists $y_{\infty}^{\beta} \in \argmin_{y \in \mathbbm{R}_{+}^{m}} \Lambda_{\infty}^{\beta}(y)$ such that $y_{\infty}^{\beta} > 0$.
\end{assumption}

Extending this assumption to accommodate constraints that are strictly slack in the limit is simple. However, it's harder to accommodate constraints that just barely hold in the limit. See \cites{Jiang} recent work for an interesting analysis of the degenerate-limit case.

The final assumption is the multivariate analog of \cites{Lueker1995} local restriction. \citeauthor{Lueker1995} imposed two critical constraints on the joint distribution of $(u_{1}, a_{1})$: a local restriction that holds in a neighborhood of the $u_{1} = a_{1}'y_{\infty}^{\beta}$ level set, and a global restriction that holds across the entire breadth of the distribution. I will need only the former because all the tough calls lie at the margin. For example, the following assumption permits point masses in the distribution, so long as they do not abut the fluid model's accept-reject indifference curve. 
\begin{assumption}\label{a:marginalProb}
	There's a continuum of marginal customers that strain the resources in a linearly independent fashion: the Jacobian matrix $\tfrac{\partial}{\partial y}\E(\indicator{\Delta_{1}(y) > 0} a_{1})$ exists, is full rank, and is continuous in $y$ in a neighborhood of $y_{\infty}^{\beta}$. 
\end{assumption}

This assumption is more straightforward than the second-order growth condition imposed by \cite{Li2019b} and many others. Indeed, it simply states that shadow prices give us complete control over inventories. To see this, note that $\E(\indicator{\Delta_{1}(y) > 0} a_{1})$ is the mean resource consumption rate when we satisfy all customers with positive surplus utility, under shadow price vector $y$. Accordingly, Jacobian matrix $\tfrac{\partial}{\partial y}\E(\indicator{\Delta_{1}(y) > 0} a_{1})$ maps marginal shadow price changes to marginal consumption rate changes. This matrix being full rank ensures that we can control the inventory burn rate in a linearly independent fashion by fine-tuning $y$. For example, marginally shifting the shadow price in the direction of $(\tfrac{\partial}{\partial y}\E(\indicator{\Delta_{1}(y) > 0} a_{1}))^{-1}e_{i}$ would marginally decrease the consumption of the $i$th resource, without changing that of the other resources.

Here's a simple sufficient condition that implies Assumption \ref{a:marginalProb}.
\begin{example}\label{e:sufficientCondition1}
	Suppose that given $a_{1}$, utility $u_{1}$ has bounded and continuous conditional density function $g(u_{1} \colonBreak a_{1})$, which almost surely satisfies $g(a_{1}'y_{\infty}^{\beta}\colonBreak a_{1}) > 0$. Further, suppose that $\E(a_{1}a_{1}')$ is non-singular. 
\end{example}

The following lemma is equivalent to Assumption \ref{a:marginalProb}, so you can consider it an alternative assumption:
\begin{lemma}\label{l:defineDotNew}
	The limiting problem's second derivative is positive and continuous at its minimizer: Hessian matrix $\ddot{\Lambda}_{\infty}(y) \equiv \tfrac{\partial^{2}}{\partial y^{2}} \Lambda_{\infty}^{b}(y) =  - \tfrac{\partial}{\partial y}\E(\indicator{\Delta_{1}(y) > 0} a_{1})$ exists, is positive definite (and hence full rank), and its elements are continuous in $y$ in a neighborhood of $y_{\infty}^{\beta}$.
\end{lemma}
\noindent
Combining Lemma \ref{l:defineDotNew} with Assumption \ref{a:PositiveShadowPrice} yields the following sister lemma via the implicit function theorem.
\begin{lemma}\label{l:lipschitzYinB}
	Limiting shadow prices are locally differentiable in the resource vector: if $b$ is sufficiently close to $\beta$, then $\Lambda_{\infty}^{b}$ has a unique minimizer, $y_{\infty}^{b} > 0$, which is continuously differentiable---and hence Lipschitz continuous---in $b$, with $\tfrac{\partial}{\partial b}y_{\infty}^{b} = - \ddot{\Lambda}_{\infty}(y_{\infty}^{b})^{-1}$.
\end{lemma}
\noindent
Together, Lemmas \ref{l:defineDotNew} and \ref{l:lipschitzYinB} imply that $\ddot{\Lambda}_{\infty}(y_{\infty}^{b})$---the Hessian matrix of $\Lambda_{\infty}^{b}$ at its minimum---is continuous in $b$ in a neighborhood of $\beta$. Accordingly, $\{\omega_{i}^{b}\}_{i \in [m]}$ and $\{\sigma_{i}^{b}\}_{i \in [m]}$ are likewise continuous in $b$, where $\omega_{i}^{b}$ is an eigenvector of $\ddot{\Lambda}_{\infty}(y_{\infty}^{b})$ with eigenvalue $\sigma_{i}^{b}$. Further, since $\ddot{\Lambda}_{\infty}(y_{\infty}^{b})$ is positive definite, we can take $\{\omega_{i}^{b}\}_{i \in [m]}$ to be orthonormal and take $\{\sigma_{i}^{b}\}_{i \in [m]}$ to be real numbers that satisfy $\sigma_{1}^{b} \ge \cdots \ge \sigma_{m}^{b} > 0$ (provided that $b$ is sufficiently close to $\beta$).

Lemma \ref{l:defineDotNew} also implies that
\begin{align}
	\dot{\Lambda}_{\infty}^{b}(y) \equiv \tfrac{\partial}{\partial y}\Lambda_{\infty}^{b}(y) = b - \E(\indicator{\Delta_{1}(y) > 0} a_{1})\label{eq:defOfInfiniteGradient}
\end{align} 
exists and is continuous in $y$ a neighborhood of $y_{\infty}^{\beta}$. Unfortunately, the finite analog, $\Lambda_{t}^{b}$, is not always differentiable, but when it is, its gradient equals subgradient
\begin{align}
	\dot{\Lambda}_{t}^{b}(y) \equiv b - \sum_{s=1}^{t} \indicator{\Delta_{s}(y) > 0} a_{s}/t. \label{eq:defOftGradient}
\end{align}

Our model is now fully characterized. Thus, we are now ready for the primary results. 
\begin{theorem}\label{p:prop1}
	The optimal policy of the online linear program \emph{without learning} yields an expected regret that grows at no more than a $\log n$ asymptotic rate: $\E(R_{n}) = O(\log n) $ as $ n \rightarrow \infty $ when distribution $\mu$ is known to the decision maker.
\end{theorem}

\begin{theorem}\label{p:prop1Hard}
	The optimal policy of the online linear program \emph{with learning} yields an expected regret that grows at no more than a $\log n$ asymptotic rate: $\E(R_{n}) = O(\log n) $ as $ n \rightarrow \infty $ when distribution $\mu$ is unknown to the decision maker.
\end{theorem}

\begin{theorem}\label{p:prop2}
	The optimal policy of the online linear program \emph{without learning} yields an expected regret that grows at no less than a $\log n$ asymptotic rate: $\E(R_{n}) = \Omega(\log n) $ as $ n \rightarrow \infty $ when distribution $\mu$ is known to the decision maker.
\end{theorem}

\begin{corollary}\label{c:prop2}
	The optimal policy of the online linear program \emph{with learning} yields an expected regret that grows at no less than a $\log n$ asymptotic rate: $\E(R_{n}) = \Omega(\log n) $ as $ n \rightarrow \infty $ when distribution $\mu$ is unknown to the decision maker.
\end{corollary}

Since knowing $\mu$ won't decrease your regret, Corollary \ref{c:prop2} follows immediately from Theorem \ref{p:prop2}, and Theorem \ref{p:prop1} follows immediately from Theorem \ref{p:prop1Hard}. However, I don't call Theorem \ref{p:prop1} a corollary because I provide an independent proof for it. Indeed, I will use the proof of Theorem \ref{p:prop1} as a stepping stone to the proof of Theorem \ref{p:prop1Hard}.

Also note that the single-dimensional results of Section \ref{s:bound} and \cite{Lueker1995} imply none of the multi-dimensional results above---the previous findings establish that an online linear program \emph{can} exhibit $\log n$ regret, but not that it \emph{must} do so. Naturally, the regret could be larger for the ``harder" online linear program, but the regret also be \emph{smaller}. Indeed, while an additional restriction cannot increase the objective value, it can decrease the regret by burdening the offline problem more than the online problem. For instance, Examples \ref{e:makeRegretGoToZero} and \ref{e:makeRegretGoToZero2} illustrate that adding a second constraint can reduce the expected regret from $\Theta(\log n)$ to $o(1)$ in the multisecretary problem, and from $\Theta(\log n)$ to $O(1)$ in \cites{Arlotto2020} stochastic knapsack problem. Hence, some constraints negate the $\log n$ regret rate; I must prove that all such negating constraints violate our assumptions.

Further, increasing the dimesionality can lower the regret by helping the period-$t$ inventory vector, $b_{t}$, escape the neighborhood of $\beta$ for which the expected myopic regret is high. Indeed, since we have imposed only local restrictions on the demand function, we can lower bound the myopic regret only when $b_{t}$ lies in some ball $B_{\delta}(\beta)$. And the more dimensions $B_{\delta}(\beta)$ has, the easier it is for $b_{t}$ to break free.

Finally, even if we imposed global demand restrictions so that all $\mathbbm{R}_{>0}^{m}$ had uniformly high myopic regrets, we still couldn't easily collapse the state space to a single dimension. The natural state space collapse would project the resource and inventory vectors onto shadow prices, defining $\tilde{a}_{t} \equiv a_{t}'y_{n}^{\beta}/\vertii{y_{n}^{\beta}}$ and $\tilde{b}_{t} \equiv b_{t}'y_{n}^{\beta}/\vertii{y_{n}^{\beta}}$. But this projection doesn't neatly transform the model into \cites{Lueker1995} framework, because the shadow prices we project onto are a function of the data, which makes the resulting $\tilde{a}_{t}$ values interdependent.

\begin{example}\label{e:makeRegretGoToZero}
	Consider the multisecretary problem of Section \ref{s:boundSetup}, but with an additional payroll budget constraint: now, in addition to the $n \beta$ available positions, you also start with $n\beta/2$ dollars, which you use to pay your workforce. The $t$th man commands wage $u_{t}$, so the applicants all yield the same bang for the buck. By design, you will almost certainly run out of money before you fill all the positions when $n$ is large, both under the optimal online and offline policies. Hence, only your payroll budget constraint is relevant as $n \rightarrow \infty$. But you will never regret how you spend this budget because every dollar yields the same marginal utility. Accordingly, the regret must go to zero, almost surely, as $n \rightarrow \infty$. 
\end{example}

\begin{example}\label{e:makeRegretGoToZero2}
	Suppose you sequentially decide which of $n$ items to add to a knapsack with volume $n/4$. The item volumes are independent uniform random variables, but the item values are all one dollar, so you want to add small items to your backpack. \cites{Lueker1995} theorem indicates that this problem has $\Theta(\log n)$ expected regret. But now suppose you have an A backpack and a B backpack, both with volume $n/4$. And suppose each item comprises an A part, which can be stored only in the A backpack, and a B part, which can be stored only in the B backpack. Furthermore, suppose that the volume of the A part is a uniform random variable and that the volume of the B part is one minus the volume of the A part, and hence also a uniform random variable. Finally, you get a dollar for each part you add to a backpack. Now, if you could put one part of an item in a knapsack and not the other part, your problem would decompose into two independent problems, each with $\Theta(\log n)$ expected regret. But if putting one part of an item into a backpack compels you to put the other part in the other backpack, then your regret is only $O(1)$. To see this, note that your regret will be proportional to the leftover space in one backpack when the other one is filled, and Corollary \ref{c:concentrateMeasure} establishes that this wasted capacity is $O(1)$ under Algorithm \ref{alg:upper}. 
\end{example}

\subsection{Dual Convergence Results}\label{s:convergence}

Everything boils down to shadow prices, so we can only make progress once we understand how $y_{t}^{b}$ converges to $y_{\infty}^{b}$. I will thus begin the analysis by presenting four propositions that crisply characterize the shadow prices' limiting behavior. 

\begin{proposition}\label{l:Gaussian}
	There exists $\delta > 0$ such that $ \sqrt{t}(y_{t}^{b} - y_{\infty}^{b}) \stackrel{d}{\rightarrow} \mathcal{N}(0, \Sigma^{b})$ for all $b \in B_{\delta}(\beta)$, where $ \Sigma^{b} \equiv \ddot{\Lambda}_{\infty}(y_{\infty}^{b})^{-1}\cov(\indicator{\Delta_{1}(y_{\infty}^{b}) > 0}a_{1}) \ddot{\Lambda}_{\infty}(y_{\infty}^{b})^{-1} $ is full rank and continuous in $b \in B_{\delta}(\beta)$.
\end{proposition}
\noindent
Unfortunately, this proposition proved less helpful than I had hoped because the rate of convergence could depend on $b$---i.e., the magnitude of $t$ required to ensure that $\sqrt{t}(y_{t}^{b} - y_{\infty}^{b}) \approx \mathcal{N}(0, \Sigma^{b})$ could be unbounded in any neighborhood of $\beta$. This, unfortunately, won't do because I'll need to invoke my convergence results at a random value of $b_{t}$. Hence, rather than Proposition \ref{l:Gaussian}, I will use the following results, which control the limiting shadow prices \emph{uniformly} across $b \in B_{\delta}(\beta)$.

\begin{proposition}\label{p:withinBallBound}
	There exists $\delta > 0$ such that $\E(\sup_{b \in B_{\delta}(\beta)}\vertii{y_{t}^{b} - y_{\infty}^{b}}^{2}) = O(1/t)$.
\end{proposition}

\begin{proposition}\label{p:OmegaOneOverT}
	There exists $\delta > 0$ such that $ \E(\inf_{b \in B_{\delta}(\beta)}\vertii{y_{t}^{b} - y_{\infty}^{b}}^{2}) = \Omega(1/t)$.
\end{proposition}

\begin{corollary}\label{c:OmegaOneOverT}
	There exists $\delta > 0$ such that the covariance matrix of $y_{t}^{b}$ has a $\Theta(1/t)$ spectral norm, for all $b \in B_{\delta}(\beta)$.
\end{corollary}

Note, positioning the $\sup_{b \in B_{\delta}(\beta)}$ and $\inf_{b \in B_{\delta}(\beta)}$ terms inside of the expectations makes these results especially strong. We'll need this extra strength to bound the regret when $\mu$ is unknown, in which case shadow prices and inventory vectors become tangled. (\cite{Jiang} independently developed a $\sup_{b \in B_{\delta}(\beta)}$-free version of Proposition \ref{p:withinBallBound}.)

Proposition \ref{p:withinBallBound} is a stronger version of \cites{Li2019b} first theorem, which states that $\E(\vertii{y_{t}^{b} - y_{\infty}^{b}}^{2}) = O((\log \log t)/t)$. I had to shave off the repeated logarithms to derive a sharp $\log n$ upper bound. I did so with a new approach. I first bounded the difference between $ y_{t}^{b} $ and $ y_{\infty}^{b} $ with the difference between the limiting gradient, $ \dot{\Lambda}_{\infty}^{b}(\cdot) $, and its finite analog, $ \dot{\Lambda}^{b}_{t}(\cdot) $, evaluated at the shadow price midway point, $ \hat{y}_{t}^{b} \equiv (y_{t}^{b} + y_{\infty}^{b}) / 2 $. But $ \hat{y}_{t}^{b} $ is difficult to work with, so I then bounded the expected value of $\vertii{\dot{\Lambda}^{b}_{t}(\hat{y}_{t}^{b}) - \dot{\Lambda}^{b}_{\infty}(\hat{y}_{t}^{b})}^{2}$ with the expected value of $\sup_{y \in B_{2\epsilon}(y_{\infty}^{\beta})} \vertii{\dot{\Lambda}^{b}_{t}(y) - \dot{\Lambda}^{b}_{\infty}(y)}^{2}$. Finally, I bounded the expected value of this supremum with a classic empirical processes result.

I also used empirical processes to prove Proposition \ref{p:OmegaOneOverT}, which will permit the corresponding lower regret bound. Specifically, I establish this result by showing that $\sqrt{t}(y_{t}^{b} - y_{t}^{b})$ is near $\gamma \in \mathbbm{R}^{m}$ if $\sqrt{t}(\dot{\Lambda}^{\beta}_{t}(y) - \dot{\Lambda}^{\beta}_{\infty}(y))$ is near $\ddot{\Lambda}_{\infty}(y_{\infty}^{\beta})\gamma$ for all $y$ in a neighborhood of $y_{\infty}^{\beta}$, and this latter condition holds because the mapping $(j, y) \mapsto \sqrt{t} e_{j}'(\dot{\Lambda}_{t}^{b}(y) - \dot{\Lambda}_{\infty}^{b}(y))$ converges to a sufficiently well-behaved Gaussian process, indexed by $y$ and $j$. 

While the previous propositions establish that our shadow price variances falls linearly with $t$, the following proposition and corollary show that their tails falls exponentially with $t$.

\begin{proposition} \label{p:outsideBallBound}
	For all $p \ge 0$, there exist $ \delta, C > 0 $ such that $ \E\big(\sup_{b \in B_{\delta}(\beta)}\indicator{y_{t}^{b} \notin B_{\epsilon}(y_{\infty}^{b})} \vertii{y_{t}^{b} - y_{\infty}^{b}}^{p}\big) \le \exp(-C \epsilon^{2} t)$ for all sufficiently small $\epsilon > 0$ and sufficiently large $t$.
\end{proposition}

\begin{corollary} \label{c:outsideBallBound}
	There exist $ \delta, C > 0 $ such that $ \Pr\big(\sup_{b \in B_{\delta}(\beta)} \vertii{y_{t}^{b} - y_{\infty}^{b}} > \epsilon\big) \le \exp(-C \epsilon^{2} t)$ for all sufficiently small $\epsilon > 0$ and sufficiently large $t$.
\end{corollary}

Whereas \cites{Li2019b} third proposition establishes a concentration of measure for random subgradient $\dot{\Lambda}_{t}^{b}(y_{\infty}^{b})$, Corollary \ref{c:outsideBallBound} establishes a concentration of measure for random shadow price $y_{t}^{b}$. This latter result is far harder to prove because $y_{t}^{b}$ is not a sum of \iid\ random variables, unlike $\dot{\Lambda}_{t}^{b}(y_{\infty}^{b})$. I establish the shadow price concentration of measure by projecting the shadow prices onto the subgradient of the dual value function at many points. These projections yield inequalities that describe a small box around $y_{t}^{b}$ and $y_{\infty}^{b}$. This box has random faces, so its walls don't meet at 90-degree angles, but the angles exhibit a concentration of measure, so the probability that the wall's fluctuations undermine the box's integrity falls exponentially fast with $t$. 

\subsection{Upper Bound with Known Distribution} \label{s:upperBoundHard}

I will now prove Theorem \ref{p:prop1} by showing that Algorithm \ref{alg:upper} honors its $O(\log n)$ bound. I will begin by showing that the inventory levels follow a martingale under this algorithm. This martingale property concentrates the distribution of $b_{t}$ to the small neighborhood of $\beta$ for which our lemmas apply. Next, I will express the values received under Algorithm \ref{alg:upper} and those received under the optimal algorithm with Bellman-style recurrence relations. I will then combine these recurrence relations to create an analogous regret recurrence relation, which I will unravel to create a corresponding regret recurrence relation. Finally, I will bound this decomposition's myopic regret with our shadow price convergence results. 

Algorithm \ref{alg:upper} satisfies the period-$t$ customer if and only if (i) there is inventory enough to do so (i.e., $tb_{t} \ge a_{t}$) and (ii) the customer has positive surplus utility under the fluid-approximation shadow prices (i.e., $\Delta_{t}(y_{\infty}^{b_{t}}) > 0$). Under this policy, the inventory vector follows a martingale: for $t > 1$, $b_{t} \ge \alpha / t$, and $b_{t}$ sufficiently close to $\beta$, we have
\begin{align*}
	\E(b_{t-1} \colonBreak b_{t}) & = \E(\psi_{t}^{b_{t}}(x_{t}a_{t}) \colonBreak b_{t})\\
	& = (t b_{t} - \E(\indicator{\Delta_{t}(y_{\infty}^{b_{t}}) > 0}a_{t}\colonBreak b_{t})) / (t-1)\\
	& = (t b_{t} - b_{t} + \dot{\Lambda}_{\infty}^{b_{t}}(y_{\infty}^{b_{t}})) / (t-1)\\
	& = b_{t} + \dot{\Lambda}_{\infty}^{b_{t}}(y_{\infty}^{b_{t}}) / (t-1)\\
	& = b_{t}.
\end{align*}
This martingale property implies the following, via the Azuma–Hoeffding inequality.

\begin{lemma}\label{l:concentrateMeasure}
	The inventory vector abides by a concentration of measure, under Algorithm \ref{alg:upper}: for all $\delta > 0$, there exists $ C > 0 $ such that $\Pr(b_{t} \notin B_{\delta}(\beta)) \le \exp(-C t)$, for all sufficiently large $ t $.
\end{lemma}

This result is stronger than one \cite{Li2019b} used. To see this, let $\tau(\delta)$ represent the first time that $b_{t}$ leaves $B_{\delta}(\beta)$:
\begin{align}
	\tau(\delta) \equiv 
	\begin{cases}
		0 & \{b_{t} \colonBreak t \in [n]\} \subset B_{\delta}(\beta),\\
		\max \{t \colonBreak b_{t} \notin B_{\delta}(\beta)\} & \text{otherwise} .
	\end{cases} \label{eq:defOfTau}
\end{align}
\citeauthor{Li2019b} proved that their algorithm yields $\E(\tau(\delta)) = O(\log n \log \log n)$---i.e., that it constrains the resource vector for all but the last $O(\log n \log \log n)$ periods. But I couldn't use this $O(\log n \log \log n)$ result to derive a $O(\log n)$ regret bound, so I had to sharpen their finding. To my surprise, I managed to tighten it $O(1)$, as the following corollary explains.
\begin{corollary}\label{c:concentrateMeasure}
	The time remaining after the resource vector leaves a given neighborhood of $\beta$ is asymptotically independent of $n$, under Algorithm \ref{alg:upper}: $\E(\tau(\delta)) = O(1)$ as $n \rightarrow \infty$, for all $\delta > 0$.
\end{corollary}

\begin{algo}\label{alg:upper}
	~\\ \vspace{-25pt} 
	\begin{enumerate}
		\item \courier{input $n$, $\beta$, $\{u_{t}\}_{t = 1}^{n}$, $\{a_{t}\}_{t = 1}^{n}$, $\mu$}
		\item \courier{initialize $b_{n} \coloneqq \beta$}
		\item \courier{for $t$ from $n$ to $1$ do}
		\begin{enumerate}
			\item \courier{set $x_{t} \coloneqq \indicator{\Delta_{t}(y_{\infty}^{b_{t}}) > 0} \indicator{tb_{t} \ge a_{t}}$}
			\item \courier{set $b_{t-1} \coloneqq \psi_{t}^{b_{t}}(x_{t}a_{t})$}
		\end{enumerate}
		\item \courier{end for}
		\item \courier{output $\{x_{t}\}_{t = 1}^{n}$}
	\end{enumerate}
\end{algo}

Algorithm \ref{alg:upper} yields total value $\bar{v}_{n}$, where 
\begin{align}
	\bar{v}_{t} &\equiv \indicator{\Delta_{t}(y_{\infty}^{b_{t}}) > 0} \indicator{tb_{t} \ge a_{t}} u_{t} + \bar{v}_{t-1}\label{eq:zCases}\\
	\aq \bar{v}_{0} & \equiv 0. \no
\end{align}
Since the optimal policy is no worse than our martingale policy, we have 
\begin{align}
	\E(\bar{R}_{n}) & \ge \E(R_{n}) \label{eq:myopicPolicyRegret}\\
	\wq \bar{R}_{t} & \equiv \bar{V}_{t}^{b_{t}} - \bar{v}_{t}.\no
\end{align}
Accordingly, it will suffice to show that $\E(\bar{R}_{n}) = O(\log n) $.

I will now present some basic linear programming identities to express regret $\bar{R}_{n}$ in terms of shadow prices. First, note that we can express the problem in \eqref{eq:primalProgram} recursively:
\begin{align}
	\bar{V}_{t}^{b} \equiv & 
	\begin{cases}
		\max_{x_{t} \in [0, 1]} x_{t}u_{t} + \bar{V}_{t-1}^{\psi_{t}^{b}(x_{t}a_{t})} & tb \ge a_{t},\\ 
		\bar{V}_{t-1}^{\psi_{t}^{b}(0)} & tb < a_{t}.
	\end{cases} \label{eq:LPBellman}
\end{align}
Second, since the shadow price weakly decreases with the inventory level, we have the following for $ x \in [0, 1] $ and $tb \ge a_{t}$:

\begin{align}
	(1-x)a_{t}'y_{t-1}^{\psi_{t}^{b}(0)} & \le \bar{V}_{t-1}^{\psi_{t}^{b}(x a_{t})} - \bar{V}_{t-1}^{\psi_{t}^{b}(a_{t})}  \le (1-x)a_{t}'y_{t-1}^{\psi_{t}^{b}(a_{t})} \label{eq:wwy2} \\
	x a_{t}'y_{t-1}^{\psi_{t}^{b}(0)}
	& \le \bar{V}_{t-1}^{\psi_{t}^{b}(0)} - \bar{V}_{t-1}^{\psi_{t}^{b}(xa_{t})} \le x a_{t}'y_{t-1}^{\psi_{t}^{b}(a_{t})} . \label{eq:wwy}
\end{align}
Third, $\Delta_{t}(y_{\infty}^{b_{t}}) > 0$ and $tb_{t} \ge a_{t}$ imply $b_{t-1} = \psi_{t}^{b_{t}}(a_{t})$ and hence $\bar{v}_{t-1} = \bar{V}_{t-1}^{\psi_{t}^{b_{t}}(a_{t})} - \bar{R}_{t-1}$. Accordingly, lines \eqref{eq:zCases}--\eqref{eq:wwy2} yield the following, when $\Delta_{t}(y_{\infty}^{b_{t}}) > 0$ and $tb_{t} \ge a_{t}$:
\begin{align}
	\bar{R}_{t} = & \bar{V}_{t}^{b_{t}} - \bar{v}_{t} \n
	= & \max_{x \in [0, 1]} u_{t} x + \bar{V}_{t-1}^{\psi_{t}^{b_{t}}(xa_{t})} - u_{t} - \bar{v}_{t-1} \n
	= & \max_{x \in [0, 1]} u_{t} (x-1) + \bar{V}_{t-1}^{\psi_{t}^{b_{t}}(xa_{t})} - \bar{V}_{t-1}^{\psi_{t}^{b_{t}}(a_{t})} + \bar{R}_{t-1} \n
	\le & \max_{x \in [0, 1]} u_{t} (x - 1) + (1 - x)a_{t}'y_{t-1}^{\psi_{t}^{b_{t}}(a_{t})} + \bar{R}_{t-1} \n
	= & (a_{t}'y_{t-1}^{\psi_{t}^{b_{t}}(a_{t})} - u_{t})^{+} + \bar{R}_{t-1}\n
	= & \Delta_{t}(y_{t-1}^{\psi_{t}^{b_{t}}(a_{t})})^{-} + \bar{R}_{t-1}. \label{eq:approxRegret1}
\end{align}
Analogously, if $\Delta_{t}(y_{\infty}^{b_{t}}) \le 0$ and $tb_{t} \ge a_{t}$ then \eqref{eq:zCases}, \eqref{eq:LPBellman}, and \eqref{eq:wwy} yield
\begin{align}
	\bar{R}_{t} \le & \Delta_{t}(y_{t-1}^{\psi_{t}^{b_{t}}(0)})^{+} + \bar{R}_{t-1} , \label{eq:approxRegret2}
\end{align}
And if $tb_{t} < a_{t}$ then \eqref{eq:zCases} and \eqref{eq:LPBellman} yield
\begin{align}
	\bar{R}_{t} = \bar{R}_{t-1}. \label{eq:approxRegret3}
\end{align}
Finally, since $ \bar{V}_{t}^{b_{t}} $ can't exceed the sum of the remaining utilities, we must also have
\begin{align}
	\bar{R}_{t} & \le \sum_{s=1}^{t} u_{s} + \bar{R}_{t-1}. \label{eq:approxRegret4}
\end{align}
Now combining inequalities \eqref{eq:approxRegret1}--\eqref{eq:approxRegret4} inductively yields the following, for sufficiently small $\delta > 0$:
\begin{align}
	\bar{R}_{n} \le & \sum_{t=1}^{n} r_{t}, \label{eq:myopicRegretDecomp}\\
	\wq r_{t} & \equiv \indicator{b_{t} \notin B_{\delta/2}(\beta)} \sum_{s=1}^{t} u_{s}\n
	& \quad + \indicator{b_{t} \in B_{\delta/2}(\beta)}\indicator{\Delta_{t}(y_{\infty}^{b_{t}}) > 0}\Delta_{t}(y_{t-1}^{\psi_{t}^{b_{t}}(a_{t})})^{-} \n
	& \quad + \indicator{b_{t} \in B_{\delta/2}(\beta)} \indicator{\Delta_{t}(y_{\infty}^{b_{t}}) \le 0} \Delta_{t}(y_{t-1}^{\psi_{t}^{b_{t}}(0)})^{+} .\no
\end{align}
Note, I condition on $b_{t} \in B_{\delta/2}(\beta)$, because that implies $\psi_{t}^{b_{t}}(0), \psi_{t}^{b_{t}}(a_{t}) \in B_{\delta}(\beta)$, when $t$ is large. 

Finally, combining lines \eqref{eq:myopicPolicyRegret} and \eqref{eq:myopicRegretDecomp} with the following lemma yields Theorem \ref{p:prop1}. 

\begin{lemma}\label{l:MyopicRegretFirst}
	The expected period-$t$ myopic regret is $O(1/t)$ under Algorithm \ref{alg:upper}: there exists $C > 0$ such that $\E(r_{t}) \le C/t$, for all $n \in \mathbbm{N}$ and $t \le n$.
\end{lemma}

To control the first term of the myopic regret, I use the fact that $\E(\sum_{s=1}^{t} u_{s})$ increases linearly in $t$, whereas $\Pr(b_{t} \notin B_{\delta/2}(\beta))$ falls exponentially, by Lemma \ref{l:concentrateMeasure}. To control the second term, I bound $\E(\indicator{\Delta_{t}(y_{\infty}^{b_{t}}) > 0} \Delta_{t}(y_{t-1}^{\psi_{t}^{b_{t}}(a_{t})})^{-})$ in terms of $\E(\sup_{b \in B_{\delta}(\beta)}\vertii{y_{t-1}^{b} - y_{\infty}^{b}}^{2})$, $ \E(\sup_{b \in B_{\delta}(\beta)}\indicator{y_{t-1}^{b} \notin B_{\epsilon}(y_{\infty}^{b})} \vertii{y_{t-1}^{b} - y_{\infty}^{b}})$, and $\Pr(\sup_{b \in B_{\delta}(\beta)} \vertii{y_{t}^{b} - y_{\infty}^{b}} > \epsilon)$, and then apply Propositions \ref{p:withinBallBound} and \ref{p:outsideBallBound} and Corollary \ref{c:outsideBallBound}. And I control the third term similarly since the argument also holds in the mirror image.

\subsection{Upper Bound with Unknown Distribution} \label{s:upperBoundHarder}

I will now prove Theorem \ref{p:prop1Hard} by showing that Algorithm \ref{alg:upperLearning} honors its $O(\log n)$ bound. The only difference between Algorithms \ref{alg:upper} and \ref{alg:upperLearning} is that the former uses limiting shadow price $y_{\infty}^{b_{t}}$, which requires knowledge of $\mu$, whereas the latter uses \emph{look-back shadow price $\underleftarrow{y}_{t}^{b_{t}}$}, which is an estimate of $y_{\infty}^{b_{t}}$ given the data observed up until period $t+1$. More specifically $\underleftarrow{y}_{t}^{b_{t}}$ is a minimizer of the backwards-looking problem
\begin{align}
	\underleftarrow{\Lambda}_{t}^{b}(y) \equiv b'y + \sum_{s = t+1}^{n}\Delta_{s}(y)^{+}/(n-t) . \label{eq:lookBackDual}
\end{align}

Our shadow price convergence results hold for look-back shadow prices but with $(n-t)$-period scaling rather than $t$-period scaling. For example, Proposition \ref{p:withinBallBound} implies that $\E(\indicator{b_{t} \in B_{\delta}(\beta)} \vertii{\underleftarrow{y}_{t}^{b_{t}} - y_{\infty}^{b_{t}}}^{2}) = O(1/(n-t))$. (Note that this would \emph{not} be the case if the proposition positioned the $\sup_{b \in B_{\delta}(\beta)}$ term outside of the expectation, since $b_{t}$ correlates with the random map $b \mapsto \underleftarrow{y}_{t}^{b}$.)

\begin{algo}\label{alg:upperLearning}
	~\\ \vspace{-25pt} 
	\begin{enumerate}
		\item \courier{input $n$, $\beta$, $\{u_{t}\}_{t = 1}^{n}$, $\{a_{t}\}_{t = 1}^{n}$}
		\item \courier{initialize $b_{n} \coloneqq \beta$}
		\item \courier{for $t$ from $n$ to $1$ do}
		\begin{enumerate}
			\item \courier{set $x_{t} \coloneqq \indicator{\Delta_{t}(\underleftarrow{y}_{t}^{b_{t}}) > 0} \indicator{tb_{t} \ge a_{t}}$}
			\item \courier{set $b_{t-1} \coloneqq \psi_{t}^{b_{t}}(x_{t}a_{t})$}
		\end{enumerate}
		\item \courier{end for}
		\item \courier{output $\{x_{t}\}_{t = 1}^{n}$}
	\end{enumerate}
\end{algo}

While the inventory vector does not follow a martingale under Algorithm \ref{alg:upperLearning}, as it does under Algorithm \ref{alg:upper}, we can still exercise control over its trajectory for all but $O(1)$ periods, as the following results establish. 

\begin{lemma}\label{l:concentrateMeasureLearning}
	The inventory vector abides by a concentration of measure, under Algorithm \ref{alg:upperLearning}: for all $\delta > 0$, there exists $ C > 0 $ such that $\Pr(b_{t} \notin B_{\delta}(\beta)) \le \exp(-C \min(t, \sqrt{n}))$, for all sufficiently large $ t \le n $.
\end{lemma}

\begin{corollary}\label{c:concentrateMeasureLearning}
	The time remaining after the resource vector leaves a given neighborhood of $\beta$ is asymptotically independent of $n$, under Algorithm \ref{alg:upperLearning}: $\E(\tau(\delta)) = O(1)$ as $n \rightarrow \infty$, for all $\delta > 0$.
\end{corollary}

The critical insight underlying Lemma \ref{l:concentrateMeasureLearning} is that $b_{t}$ can't escape $B_{\delta/2}(\beta)$ in less than $\Omega(n)$ time, and hence without first generating an $\Omega(n)$-sized sample of training data. This means that by the time the $\{b_{t}\}_{t = n}^{1}$ process has made it halfway out of $B_{\delta}(\beta)$---i.e., departed $B_{\delta/2}(\beta)$---our look-back shadow prices are accurate enough to (almost) guarantee that it can't traverse the second half. This property enables us to restrict attention to the periods with accurate look-back shadow prices (i.e., periods after time $\tau(\delta/2)$). 

But controlling the evolution of $\{b_{t}\}_{t = n}^{1}$ is difficult even when look-back shadow prices are accurate. The problem is that while $b_{t}$ is independent of the mapping $b \mapsto y_{t}^{b}$, it is not independent of the mapping $b \mapsto \underleftarrow{y}_{t}^{b}$. Indeed, the inventory vectors and look-back shadow prices intertwine in a complex dance. To extricate $b_{t}$ from this pas de deux, I decompose it into three parts: $b_{\tau(\delta/2) + 1}$, $\sum_{s = t}^{\tau(\delta/2)} b_{s} - \E(b_{s} \colonBreak b_{s + 1})$, and $\sum_{s = t}^{\tau(\delta/2)} \E(b_{s} \colonBreak b_{s + 1}) - b_{s+1}$. By definition, the first part is within $\delta/2$ of $\beta$. The second part follows a martingale and thus concentrates about zero. And the third part is small, provided that $\underleftarrow{y}_{s}^{b_{s}}$ is near $y_{\infty}^{b_{s}}$, for all $s \in \{t + 1, \cdots, \tau(\delta/2) + 1\}$. Crucially, $\tau(\delta/2)$ will be small enough to ensure that this holds with high probability, provided that $b_{s}$ is near $\beta$ for all $s \in \{t + 1, \cdots, \tau(\delta/2) + 1\}$. And thus, I can inductively establish the result: $b_{s}$ being near $\beta$ for $s \in \{t + 1, \cdots, \tau(\delta/2) + 1\}$ implies that $\underleftarrow{y}_{s}^{b_{s}}$ is near $y_{\infty}^{b_{s}}$ for $s \in \{t + 1, \cdots, \tau(\delta/2) + 1\}$, which implies that $\sum_{s = t}^{\tau(\delta/2)} \E(b_{s} \colonBreak b_{s + 1}) - b_{s+1}$ is small, which implies that $b_{t}$ is near $\beta$.

Having reigned in our inventory vectors, we are ready to decompose our regret. The analysis of the previous section yields the following:
\begin{align}
	\bar{R}_{n} \le & \sum_{t=1}^{n} r_{t}, \label{eq:myopicRegretDecompLearn}\\
	\wq r_{t} & \equiv \indicator{b_{t} \notin B_{\delta/2}(\beta)} \sum_{s=1}^{t} u_{s}\n
	& \quad + \indicator{b_{t} \in B_{\delta/2}(\beta)}\indicator{\Delta_{t}(\underleftarrow{y}_{t}^{b_{t}}) > 0}\Delta_{t}(y_{t-1}^{\psi_{t}^{b_{t}}(a_{t})})^{-} \n
	& \quad + \indicator{b_{t} \in B_{\delta/2}(\beta)} \indicator{\Delta_{t}(\underleftarrow{y}_{t}^{b_{t}}) \le 0} \Delta_{t}(y_{t-1}^{\psi_{t}^{b_{t}}(0)})^{+} .\no
\end{align}
The only differences between lines \eqref{eq:myopicRegretDecomp} and \eqref{eq:myopicRegretDecompLearn} is that $\bar{R}_{n}$ and $b_{t}$ now correspond to Algorithm \ref{alg:upperLearning}, rather than that Algorithm \ref{alg:upper}, and $\underleftarrow{y}_{t}^{b_{t}}$ has replaced $y_{\infty}^{b_{t}}$.

Finally, combining lines \eqref{eq:myopicPolicyRegret} and \eqref{eq:myopicRegretDecompLearn} with the following lemma yields Theorem \ref{p:prop1Hard}. 

\begin{lemma}\label{l:MyopicRegretFirstLearn}
	The expected period-$t$ myopic regret is $O(1/t) + O(1/(n-t))$ under Algorithm \ref{alg:upperLearning}: there exists $C > 0$ such that $\E(r_{t}) \le C/t + C/(n-t)$, for all $n \in \mathbbm{N}$ and $t \le n$.
\end{lemma}

This lemma is the same as Lemma \ref{l:MyopicRegretFirst}, except now both the $O(1/\sqrt{t})$ errors between $y_{t}^{b_{t}}$ and $y_{\infty}^{b_{t}}$ and the $O(1/\sqrt{n - t})$ errors between $\underleftarrow{y}_{t}^{b_{t}}$ and $y_{\infty}^{b_{t}}$ contribute to your regret.

\subsection{Lower Bound with Known Distribution} \label{s:lowerBoundHard}

I will now prove Theorem \ref{p:prop2} by creating a lower-bounding version of the methodology developed in Section \ref{s:upperBoundHard}. For example, the lower-bounding decomposition will depend on $\Delta_{t}(y_{t-1}^{\psi_{t}^{b_{t}}(0)})^{-}$ and $\Delta_{t}(y_{t-1}^{\psi_{t}^{b_{t}}(a_{t})})^{+}$ (as opposed to $\Delta_{t}(y_{t-1}^{\psi_{t}^{b_{t}}(a_{t})})^{-}$ and $\Delta_{t}(y_{t-1}^{\psi_{t}^{b_{t}}(0)})^{+}$); the lower-bounding version of Lemma \ref{l:concentrateMeasure} will ensure the proximity of $b_{t}$ and $\beta$ under the optimal algorithm (as opposed to Algorithm \ref{alg:upper}); and the lower-bounding version of Lemma \ref{l:MyopicRegretFirst} will establish that the expected myopic regret is $\Omega(1/t)$ (as opposed to $O(1/t)$).

In this section $\{b_{t}\}_{t = n}^{1}$ will characterize the inventory levels that correspond to the optimal actions specified in line \eqref{eq:optimalAction}: $b_{n} = \beta$ and $b_{t-1} = \psi_{t}^{b_{t}}(\pi_{t}^{b_{t}}a_{t})$. Unfortunately, we now have little control over $\{b_{t}\}_{t = n}^{1}$, because the optimal policy is unknown. Nevertheless, we can still situate $b_{t}$ near $\beta$ for a substantial time interval.

\begin{lemma}\label{l:probBInControl}
	The inventory vector tends to lie near $\beta$ under the optimal policy for most of the second half of the horizon: For all $\delta > 0$, if $n$ is sufficiently large then $n^{3/4} \le t \le n/2$ implies $\Pr(b_{t} \notin B_{\delta/2}(\beta)) \le n^{-1/2}$.
\end{lemma}

This lemma was the hardest result in this article to prove because the optimal policy is opaque. Generalizing the technique developed in Section \ref{s:lowerMultisecretary}, I argue that the regret incurred when $b_{t}$ strays from $\beta$ is at least as large as the value sacrificed when we chop the linear program into two separate problems, one with horizon $t$ and endowment $t b_{t}$ and the other with horizon $n-t$ and endowment $n \beta - t b_{t}$. The concavity of $\bar{V}_{t}^{b}$ in $b$ ensures that this division is costly when $b_{t}$ meaningfuly differs from $\beta$.

As before, we will benchmark against the offline linear program, line \eqref{eq:primalProgram}, rather than the offline integer program, line \eqref{eq:IP}. The following result will enable us to do so:
\begin{align}
	\E(\underline{R}_{n}) & = \E(R_{n}) + O(1), \label{eq:OmegaEqualsOmega} \\
	\wq \underline{R}_{t} & \equiv \bar{V}_{t}^{b_{t}} - v_{t}^{b_{t}}.\no	
\end{align}
Line \eqref{eq:OmegaEqualsOmega} holds because the linear program has a solution that partially satisfies at most $m$ customers, and thus the integer program must derive at least as much value from resource endowment $\beta$ as the linear program does from resource endowment $\beta- m\alpha/n$: $V_{n}^{\beta} \ge \bar{V}_{n}^{\beta-m\alpha/n}$. And since the shadow price decreases in the inventory level, this implies that $V_{n}^{\beta} \ge \bar{V}_{n}^{\beta} - m\alpha'y_{n}^{\beta-m\alpha/n}$, and hence that $R_{n} \ge \bar{V}_{n}^{\beta} - v_{n}^{\beta} - m\alpha'y_{n}^{\beta-m\alpha/n}$. Finally, Proposition \ref{p:withinBallBound} indicates that $\E(y_{n}^{\beta-m\alpha/n})= O(1)$ as $n \rightarrow \infty$, which establishes the result.

I will now create the lower-bounding version of our regret decomposition. First, note that $\pi_{t}^{b_{t}} = 1$ implies $b_{t-1} = \psi_{t}^{b_{t}}(a_{t})$, and hence $v_{t-1}^{\psi_{t}^{b_{t}}(a_{t})} = \bar{V}_{t-1}^{\psi_{t}^{b_{t}}(a_{t})} - \underline{R}_{t-1}$. Also, $\pi_{t}^{b_{t}} = 1$ implies $tb_{t} \ge a_{t}$, which with lines \eqref{eq:LPBellman} and \eqref{eq:wwy2} yield
\begin{align*}
	\underline{R}_{t} & \equiv \bar{V}_{t}^{b_{t}} - v_{t}^{b_{t}}\\
	& = \max_{x_{t} \in [0, 1]} x_{t}u_{t} + \bar{V}_{t-1}^{\psi_{t}^{b_{t}}(x_{t}a_{t})} - u_{t} - v_{t-1}^{\psi_{t}^{b_{t}}(a_{t})}\\
	& = \max_{x_{t} \in [0, 1]} (x_{t} - 1)u_{t} + \bar{V}_{t-1}^{\psi_{t}^{b_{t}}(x_{t}a_{t})} - \bar{V}_{t-1}^{\psi_{t}^{b_{t}}(a_{t})} + \underline{R}_{t-1}\\
	& \ge \max_{x_{t} \in [0, 1]} (x_{t} - 1)u_{t} + (1-x)a_{t}'y_{t-1}^{\psi_{t}^{b_{t}}(0)} + \underline{R}_{t-1}\\
	& = (a_{t}'y_{t-1}^{\psi_{t}^{b_{t}}(0)} - u_{t})^{+} + \underline{R}_{t-1}\\
	& = \Delta_{t}(y_{t-1}^{\psi_{t}^{b_{t}}(0)})^{-} + \underline{R}_{t-1}.
\end{align*}
Analogously, if $\pi_{t}^{b_{t}} = 0$ and $tb_{t} \ge a_{t}$ then lines \eqref{eq:LPBellman} and \eqref{eq:wwy} yield
\begin{align*}
	\underline{R}_{t} \ge & \Delta_{t}(y_{t-1}^{\psi_{t}^{b_{t}}(a_{t})})^{+} + \underline{R}_{t-1}. 
\end{align*}
Further, we always trivially have
\begin{align*}
	\underline{R}_{t} \ge & \underline{R}_{t-1}.
\end{align*}
Now choose $\delta > 0$ small enough to ensure that $\delta \iota \le \beta$, where $\iota$ is a vector of ones. In this case, $b_{t} \in B_{\delta/2}(\beta)$ implies $tb_{t} \ge a_{t}$ for $t \ge 2\vertii{\alpha}/\delta$, which with our previous three inequalities inductively implies
\begin{align}
	\underline{R}_{n} \ge & \sum_{t=\lceil2\vertii{\alpha}/\delta\rceil}^{n} r_{t},  \label{eq:lowerBoundRegretDecomp}\\
	\wq r_{t} \equiv & \indicator{b_{t} \in B_{\delta/2}(\beta)} \big(\pi_{t}^{b_{t}}\Delta_{t}(y_{t-1}^{\psi_{t}^{b_{t}}(0)})^{-} + (1 - \pi_{t}^{b_{t}})\Delta_{t}(y_{t-1}^{\psi_{t}^{b_{t}}(a_{t})})^{+}\big). \no
\end{align}
Finally, combining lines \eqref{eq:OmegaEqualsOmega} and \eqref{eq:lowerBoundRegretDecomp} with the following lemma yields Theorem \ref{p:prop2}. 

\begin{lemma}\label{l:rIsHighGivenB}
	The expected period-$t$ myopic regret is $\Omega(1/t)$ under the optimal policy for most of the second half of the horizon: There exists $C > 0$ such that $\E(r_{t}) \ge C/t$, for all sufficiently large $t$ that satisfies $n^{3/4} \le t \le n/2$.
\end{lemma}

To establish this result, I show that if $b_{t} \in B_{\delta/2}(\beta)$---which happens with high probability, by Lemma \ref{l:probBInControl}---then $\sqrt{t}(y_{t}^{b_{t}} - y_{\infty}^{b_{t}})$ could be near \emph{any} $\gamma \in \mathbbm{R}^{m}$. Accordingly, both Type I errors---rejecting customers that should have been satisfied---and Type II errors---satisfying customers that should have been rejected---are unavoidable because the shadow price can always be larger or smaller than anticipated. Specifically, I show that there's at least a $\Omega(1/\sqrt{t})$ chance that both the expected Type I and Type II errors are $\Omega(1/\sqrt{t})$. 

\section{Conclusion}

I began this project by investigating what happens when we extend secretary valuations from the finite set $ \{v_1, \cdots, v_{j}\} $ to infinite set $ [0, 1] $ in \cites{Arlotto2019} multisecretary model. I found that this extension invalidates \cites{Arlotto2019} finite regret bound. Why is this? Well, in the former case, secretaries are interchangeable parts, which means that your hiring errors can be ``undone" in the future. For example, you can confidently reject the marginal candidate in \citeauthors{Arlotto2019} setting because you'll almost certainly see his like again (many times over). But no two men are alike when values are drawn from the uniform distribution. So, in this case, your hiring mistakes cannot be undone---they will linger forever, making you ever more regretful.

Alright, fine, so your regret increases. But why does it do so at a $ \log(n) $ rate? Well, suppose there are $ n $ applicants for $ n\beta $ open positions, and the first man you interview has utility $ u_{n} $. Do you classify him as ``hired" or ``not hired"? Like all classification problems, you must balance between the threat of a Type I error and a Type II error. You make a Type I error when you extend a job offer and the ``shadow price" of an open position exceeds $ u_{n} $---in this case, you took the slot of a more deserving applicant. And you make a Type II error when you don't extend an offer and $ u_{n} $ exceeds the ``shadow price" of an open position---in this case, you reserved the slot for an inferior applicant. The ``shadow price" of a position equals the utility of the $ (n\beta) $th best man out of the remaining $ n - 1 $ candidates, which is order statistic $ h_{n-1}^{n\beta} $. Hence, you must try to anticipate whether or not $ u_{n} $ exceeds $ h_{n-1}^{n\beta} $. Fortunately, the standard deviation of $ h_{n-1}^{n\beta} $ drops like $ 1/\sqrt{n} $, so we can think of $ h_{n-1}^{n\beta} $ as residing in a ``confidence interval" whose length is on the order of $ 1/\sqrt{n} $. This has two implications. First, you'll only have around a $ 1/\sqrt{n} $ chance of making a hiring mistake because the optimal action is unambiguous when $ u_{n} $ falls outside the confidence interval. Second, the cost of a hiring mistake is also around $ 1/\sqrt{n} $ because the capability of the man you should have hired and the capability of the man you did hire both reside in the confidence interval. Thus, the expected regret of your period-$ n $ decision will be on the order of $ \underbrace{1/\sqrt{n}}_{\text{mistake probability}} \cdot \underbrace{1/\sqrt{n}}_{\text{mistake cost}} = 1/n $, which means that your total regret grows like the harmonic series. 

The same principle applies to the more general online linear program, but extending the tight $O(\log n)$ and $\Omega(\log n)$ bounds to this problem necessitated more refined control over shadow prices. Hence, I precisely characterize the convergence of dual variable $y_{t}^{b}$ to its deterministic limit, $y_{\infty}^{b}$. Specifically, I develop weak conditions in which
\begin{itemize}
	\item $\sqrt{t}(y_{t}^{b} - y_{\infty}^{b})$ converges to a multivariate normal for all $b \in B_{\delta}(\beta)$, 
	\item $\Pr(\sup_{b \in B_{\delta}(\beta)}\vertii{y_{t}^{b} - y_{\infty}^{b}} > \epsilon)$ falls exponentially fast in $t$, 
	\item $\E(\sup_{b \in B_{\delta}(\beta)}\vertii{y_{t}^{b} - y_{\infty}^{b}}^{2}) = O(1/t)$, and 
	\item $ \E(\inf_{b \in B_{\delta}(\beta)}\vertii{y_{t}^{b} - y_{\infty}^{b}}^{2}) = \Omega(1/t)$.
\end{itemize}
Further, since the $\sup_{b \in B_{\delta}(\beta)}$ and $\inf_{b \in B_{\delta}(\beta)}$ terms lie inside of the expectation, we can apply these bounds to random values of $b_{t}$, even if they correlate with shadow prices. This was the key to accommodating the most interesting aspect of the problem: online learning.

\section*{Acknowledgments}
Many thanks to Siddhartha Banerjee, Itai Gurvich, Ioannis Stamatopoulos, He Wang, and three anonymous reviewers for their very insightful feedback.

\theendnotes

\appendix
\pagebreak
\section*{List of Symbols}\label{s:glossary}
\scriptsize
\begin{tabular}{lll}
	$[x]$  &  the set $\{1, \cdots, x\}$ &\\
	$x \wedge y$  &  vector with $i$th element $\min(x_{i}, y_{i})$ &\\
	$x\vee y$  &  vector with $i$th element is $\max(x_{i}, y_{i})$ &\\
	$x^{+}$  &  $\max(0, x)$ &\\
	$x^{-}$  &  $\max(0, -x)$ &\\
	$e_{j}$  &  unit vector indicating $j$th position &\\
	$\iota$  &  vector of ones &\\
	$\indicator{}$  &  indicator function &\\
	$B_{\delta}(b)$  &  open ball with radius $\delta$ about $b$&\\
	$m$  &  number of resources to manage &\\
	$n$  &  number of time periods&\\
	$t$  &  generic time period&\\
	$b_{t}$  &  inventory vector, defined as period-$t$ inventory holdings divided by $t$&\\
	$b$  &  generic inventory vector&\\
	$\beta$  &  initial inventory vector&\\
	$\tau(\delta)$  &  first time inventory vector leaves $B_{\delta}(\beta)$& line \eqref{eq:defOfTau}\\
	$u_{t}$  &  utility received by satisfying period-$t$ customer&\\
	$a_{t}$  &  resources consumed by satisfying period-$t$ customer&\\
	$\Delta_{t}$  &  surplus utility function &  line \eqref{eq:defOfDelta}\\
	$\mu$  & joint distribution of $(u_{t}, a_{t})$ & Assumption \ref{a:distribution}\\
	$\alpha$  &  upper bound on $a_{t}$ & Assumption \ref{a:boundedA}\\
	$x_{t}$  &  period-$t$ decision variable&\\
	$\psi_{t}^{b}$  &  function determining period-$(t-1)$ inventory vector & line \eqref{eq:firstBellmanEq2}\\
	$\pi_{t}^{b}$  &  optimal action & line \eqref{eq:optimalAction}\\
	$v_{t}^{b}$  &  online objective value  & line \eqref{eq:defOfLittleV}\\
	$\bar{v}_{t}$  & martingale-policy objective value & line \eqref{eq:zCases}\\
	$V_{t}^{b}$  &  offline objective value & line \eqref{eq:IP}\\
	$\bar{V}_{t}^{b}$  &  offline objective value with linear programming relaxation & line \eqref{eq:primalProgram}\\
	$R_{n}$  &  regret & line \eqref{eq:regretToBound}\\
	$\bar{R}_{t}$  &  upper-bound regret relaxation & line \eqref{eq:myopicPolicyRegret}\\
	$\underline{R}_{t}$  &  lower-bound regret relaxation & line \eqref{eq:OmegaEqualsOmega} \\
	$r_{t}$  &  myopic regret & lines \eqref{eq:myopicRegretDecomp}, \eqref{eq:myopicRegretDecompLearn}, and \eqref{eq:lowerBoundRegretDecomp}\\
	$\Lambda_{t}^{b}$  &  dual objective & line \eqref{eq:ConvexProblem}\\
	$\underleftarrow{\Lambda}_{t}^{b}$  & look-back dual objective & line \eqref{eq:lookBackDual}\\
	$\dot{\Lambda}_{t}$  & dual objective subgradient & line \eqref{eq:defOftGradient}\\
	$\Lambda_{\infty}^{b}$  &  limiting dual objective & line \eqref{eq:infinityLambdaDef}\\
	$\dot{\Lambda}_{\infty}$  &  limiting dual gradient & line \eqref{eq:defOfInfiniteGradient}\\
	$\ddot{\Lambda}_{\infty}$  & limiting dual Hessian & Lemma \ref{l:defineDotNew}\\
	$\omega_{i}^{b}$  &  $i$th orthonormal eigenvector of limiting dual Hessian & below Lemma \ref{l:lipschitzYinB}\\
	$\sigma_{i}^{b}$  &  $i$th largest eigenvalue of limiting dual Hessian & below Lemma \ref{l:lipschitzYinB}\\
	$y_{t}^{b}$  & dual optimal solution & line \eqref{eq:notnecessarilyunique}\\
	$\underleftarrow{y}_{t}^{b}$  &  look-back dual optimal solution & before line \eqref{eq:lookBackDual}\\
	$y_{\infty}^{b}$  & limiting dual optimal solution & Assumption \ref{a:PositiveShadowPrice} and Lemma \ref{l:lipschitzYinB}\\
	$y$  &  generic dual solution &\\
\end{tabular}

\bibliography{library.bib}

\normalsize
\pagebreak
\setcounter{page}{1}
\section*{ONLINE APPENDIX OF PROOFS}\label{s:proofs}

\begin{proof}[Lemma \ref{l:defineDotNew} Proof]
	Assumption \ref{a:marginalProb} implies that the event $\Delta_{1}(y) = 0$ and $a_{1} \ne 0$ has measure zero, for $y$ sufficiently close to $y_{\infty}^{\beta}$. Hence, $\Delta_{1}(y)^{+}$ is almost surely differentiable in $y$, which means that
	\begin{align*}
		\tfrac{\partial}{\partial y} \Lambda_{\infty}^{b}(y) & = \tfrac{\partial}{\partial y} \big(b'y + \E(\Delta_{1}(y)^{+})\big)\\
		& = b+ \E\big(\tfrac{\partial}{\partial y} \Delta_{1}(y)^{+}\big)\\
		& = b - \E(\indicator{\Delta_{1}(y) > 0} a_{1}).
	\end{align*} 
	Note we can commute the expectation and differentiation because $a_{1}$ is bounded. Combining the derivative above with Assumption \ref{a:marginalProb} and the convexity of $\Lambda_{\infty}^{b}$ implies the result.
\end{proof}

\begin{proof}[Lemma \ref{l:lipschitzYinB} Proof]
	Assumption \ref{a:PositiveShadowPrice} and Lemma \ref{l:defineDotNew} imply that (i) $\dot{\Lambda}_{\infty}^{\beta}(y_{\infty}^{\beta}) = 0$, (ii) $\ddot{\Lambda}_{\infty}(y_{\infty}^{\beta})$ is non-singular, and (iii) $\dot{\Lambda}_{\infty}^{b}(y)$ is continuously differentiable in $y$ near $y_{\infty}^{\beta}$. Further, $\dot{\Lambda}_{\infty}^{b}(y)$ is continuously differentiable in $b$, since $\tfrac{\partial}{\partial b}\dot{\Lambda}_{\infty}^{b}(y) = I$ (see the proof of Lemma \ref{l:defineDotNew}). Accordingly, the implicit function theorem establishes that each $b$ in a neighborhood of $\beta$ has a corresponding shadow price vector $y_{\infty}^{b}$ that has continuous derivative $\tfrac{\partial}{\partial b} y_{\infty}^{b} = - \tfrac{\partial}{\partial y}\dot{\Lambda}_{\infty}^{b}(y)^{-1} \tfrac{\partial}{\partial b}\dot{\Lambda}_{\infty}^{b}(y)|_{y = y_{\infty}^{b}} = - \ddot{\Lambda}_{\infty}^{b}(y_{\infty}^{b})^{-1}$.  Further, $y_{\infty}^{b}$ must be the unique minimizer of $\Lambda_{\infty}^{b}$ for $b$ near $\beta$, because $\dot{\Lambda}_{\infty}^{b}(y_{\infty}^{b}) = 0$ and $\ddot{\Lambda}_{\infty}(y)$ is positive definite for $y$ near $y_{\infty}^{\beta}$.
\end{proof}

\begin{proof}[Corollary \ref{c:prop2} Proof]
	This follow immediately from Theorem \ref{p:prop2}.
\end{proof}

\begin{proof}[Proposition \ref{l:Gaussian} Proof]
	I will first establish that $\Sigma^{b}$ is continuous and full rank for all $b$ in a neighborhood of $\beta$. Lemmas \ref{l:defineDotNew} and \ref{l:lipschitzYinB} imply the continuity, and Lemma \ref{l:defineDotNew} implies that $\Sigma^{b}$ is full rank if $\cov(\indicator{\Delta_{1}(y_{\infty}^{b}) > 0}a_{1})$ is full rank. If this latter matrix were not full rank, then there would be some $\gamma \ne 0$ that almost surely satisfies $\indicator{\Delta_{1}(y_{\infty}^{b}) > 0}a_{1}'\gamma = \E(\indicator{\Delta_{1}(y_{\infty}^{b}) > 0}a_{1}'\gamma)$, which would imply that either (i) $\Delta_{1}(y_{\infty}^{b}) > 0$, almost surely, or (ii) $\indicator{\Delta_{1}(y_{\infty}^{b}) > 0}a_{1}'\gamma = 0$, almost surely. The former case violates Assumption \ref{a:marginalProb} because it implies that $\E(\indicator{\Delta_{1}(y+dy) > 0} a_{1}'\gamma) = \E(\indicator{\Delta_{1}(y) > 0} a_{1}'\gamma)$ for $dy \le 0$, and the latter case violates Assumption \ref{a:marginalProb} because it implies that $\E(\indicator{\Delta_{1}(y+dy) > 0} a_{1}'\gamma) = \E(\indicator{\Delta_{1}(y) > 0} a_{1}'\gamma)$ for $dy > 0$.
	
	The fact that $ \sqrt{t}(y_{t}^{b} - y_{\infty}^{b}) \stackrel{d}{\rightarrow} \mathcal{N}(0, \Sigma^{b})$ follows directly from theorem 2.13 of \cite{Kosorok2008}, so it will suffice to show that the conditions of this theorem hold. To use follow \citeauthors{Kosorok2008} notation, define functions
	\begin{align}
		m_{y}(u_{1}, a_{1}) \equiv & b'y + \Delta_{1}(y)^{+},\n
		\dot{m}(a_{1}) \equiv & \vertii{b} + \vertii{a_{1}}, \n
		\aq \dot{m}_{\infty}(u_{1}, a_{1}) \equiv & b - \indicator{\Delta_{1}(y_{\infty}^{b}) > 0} a_{1}. \no
	\end{align}
	First, the Hessian matrix of $ \E(m_{y}(u_{1}, a_{1})) $ at $ y = y_{\infty}^{b} $ is $ \ddot{\Lambda}_{\infty}(y_{\infty}^{b}) $, which is non-singular when $ b $ is sufficiently close to $\beta$, by Lemmas \ref{l:defineDotNew} and \ref{l:lipschitzYinB}. Second, Assumption \ref{a:boundedA} establishes that $ \E(\dot{m}(a_{1})^{2}) $ and $ \E(\vertii{\dot{m}_{\infty}(u_{1}, a_{1})}^{2}) $ are finite. Third, functions $ m_{y} $ and $ \dot{m} $ satisfy condition (2.18) of \cite{Kosorok2008}:
	\begin{align}
		\verti{m_{y}(u_{1}, a_{1}) - m_{z}(u_{1}, a_{1})} = & b'y + \Delta_{1}(y)^{+} - b'z - \Delta_{1}(z)^{+} \n
		\le & (\vertii{b} + \vertii{a_{1}}) \vertii{y - z} \n
		= & \dot{m}(a_{1}) \vertii{y - z} . \no
	\end{align}
	Fourth, Assumption \ref{a:marginalProb} ensures that functions $ m_{y} $ and $ \dot{m}_{\infty} $ satisfy condition (2.19) of \cite{Kosorok2008}:
	\begin{align*}
		\E\big(\big(& m_{y}(u_{1}, a_{1}) - m_{y_{\infty}^{b}}(u_{1}, a_{1}) - \dot{m}_{\infty}(u_{1}, a_{1})'(y - y_{\infty}^{b})\big)^{2}\big) \n
		= & \E\big(\big(\Delta_{1}(y)^{+} - \Delta_{1}(y_{\infty}^{b})^{+} + \indicator{\Delta_{1}(y_{\infty}^{b}) > 0} a_{1}'(y - y_{\infty}^{b})\big)^{2}\big) \n
		= & \E\big(\Delta_{1}(y)^{2} \verti{\indicator{\Delta_{1}(y) > 0} - \indicator{\Delta_{1}(y_{\infty}^{b}) > 0}}\big) \n
		\le & \E\big((a_{1}' y - a_{1}' y_{\infty}^{b})^{2} \verti{\indicator{\Delta_{1}(y) > 0} - \indicator{\Delta_{1}(y_{\infty}^{b}) > 0}}\big) \n
		= & \vertii{y - y_{\infty}^{b}}^{2} \E\big(\vertii{\indicator{\Delta_{1}(y) > 0}a_{1} - \indicator{\Delta_{1}(y_{\infty}^{b}) > 0}a_{1}}^{2} \big) \n
		\le & \vertii{\alpha}\vertii{y - y_{\infty}^{b}}^{2} \E\big(\indicator{\Delta_{1}(y \wedge y_{\infty}^{b}) > 0}a_{1} - \indicator{\Delta_{1}(y \vee y_{\infty}^{b}) > 0}a_{1}\big) \n
		\le & \vertii{\alpha}\vertii{y - y_{\infty}^{b}}^{2} O(\vertii{y - y_{\infty}^{b}}) \\
		= & o(\vertii{y - y_{\infty}^{b}}).
	\end{align*}
	Finally, Proposition \ref{p:withinBallBound} establishes that $ \vertii{y_{t}^{b} - y_{\infty}^{b}} \stackrel{p}{\rightarrow} 0 $.
\end{proof}

\begin{proof}[Proposition \ref{p:withinBallBound} Proof]
	Since Proposition \ref{p:outsideBallBound} establishes that $\E(\sup_{b \in B_{\delta}(\beta)}\indicator{y_{t}^{b} \notin B_{\epsilon}(y_{\infty}^{b})} \vertii{y_{t}^{b} - y_{\infty}^{b}}^{2}) = o(1/t)$, it will suffice to show that $\E(\sup_{b \in B_{\delta}(\beta)}\indicator{y_{t}^{b} \in B_{\epsilon}(y_{\infty}^{b})} \vertii{y_{t}^{b} - y_{\infty}^{b}}^{2}) = O(1/t)$, for sufficiently small $\epsilon > 0$. I will establish this result with Theorems 2.14.2 and 2.14.5 of \cite{VanderVaart1996}. However, translating the problem into \citeauthors{VanderVaart1996} empirical processes framework will take some effort. First, I bound the magnitude of $y_{t}^{b} - y_{\infty}^{b}$ in terms of the magnitude of $\dot{\Lambda}^{b}_{\infty}(\hat{y}_{t}^{b}) - \dot{\Lambda}_{t}^{b}(\hat{y}_{t}^{b})$, where $\hat{y}_{t}^{b} \equiv (y_{t}^{b} + y_{\infty}^{b})/2$. Since $\hat{y}_{t}^{b}$ lies between the minimizers of $\Lambda_{\infty}^{b}$ and $\Lambda_{t}^{b}$, the vector $\hat{y}_{t}^{b} - y_{\infty}^{b}$ projects positively onto gradient $\dot{\Lambda}^{b}_{\infty}(\hat{y}_{t}^{b})$ and projects negatively onto subgradient $\dot{\Lambda}_{t}^{b}(\hat{y}_{t}^{b})$. I use this fact to show that $(\hat{y}_{t}^{b} - y_{\infty}^{b})' (\dot{\Lambda}^{b}_{\infty}(\hat{y}_{t}^{b}) - \dot{\Lambda}_{t}^{b}(\hat{y}_{t}^{b}))$ is larger than some fixed multiple of $\vertii{\hat{y}_{t}^{b} - y_{\infty}^{b}}^{2}$, which indicates that $\vertii{\dot{\Lambda}^{b}_{\infty}(\hat{y}_{t}^{b}) - \dot{\Lambda}_{t}^{b}(\hat{y}_{t}^{b})}$ is larger than some fixed multiple of $\vertii{y_{t}^{b} - y_{\infty}^{b}}$. This, in turn, implies that the expectation of the maximum of $\vertii{\dot{\Lambda}^{b}_{t}(y) - \dot{\Lambda}^{b}_{\infty}(y)}^{2}$, across $y$ in some small ball of $y_{\infty}^{\beta}$, is larger than some fixed multiple of the expectation of $\indicator{\vertii{y_{t}^{b} - y_{\infty}^{b}} \le \epsilon}\vertii{y_{t}^{b} - y_{\infty}^{b}}^{2}$. And bounding the expectation of the maximum of $\vertii{\dot{\Lambda}^{b}_{t}(y) - \dot{\Lambda}^{b}_{\infty}(y)}^{2}$ is a classic empirical processes problem.
	
	Now let's get to the proof. First, Lemma \ref{l:lipschitzYinB} establishes that we can choose $\delta$ small enough so that $y_{\infty}^{b} \in B_{\epsilon}(y_{\infty}^{\beta})$ for all $b \in B_{\delta}(\beta)$, in which case $y_{t}^{b} \in B_{\epsilon}(y_{\infty}^{b})$ implies $y_{t}^{b} \in B_{2\epsilon}(y_{\infty}^{\beta})$, which in turn implies $\hat{y}_{t}^{b} \in B_{3\epsilon/2}(y_{\infty}^{b})$, where $\hat{y}_{t}^{b} \equiv (y_{t}^{b} + y_{\infty}^{b})/2$.
	
	Second, let $\sigma_{m}^{b}$ denote the smallest singular value of $\ddot{\Lambda}_{\infty}(y_{\infty}^{b})$. Lemmas \ref{l:defineDotNew} and \ref{l:lipschitzYinB} imply that we can set $\delta$ small enough so that for all $b \in B_{\delta}(\beta)$ we have $\sigma_{m}^{b} \ge \sigma_{m}^{\beta}/2$, and hence
	\begin{align*}
		(\hat{y}_{t}^{b} - y_{\infty}^{b})'\ddot{\Lambda}_{\infty}(y_{\infty}^{b}) (\hat{y}_{t}^{b}- y_{\infty}^{b}) \ge \sigma_{m}^{\beta} \vertii{\hat{y}_{t}^{b}- y_{\infty}^{b}}^{2}/2.
	\end{align*}
	Next, note that $\dot{\Lambda}^{b}_{\infty}(y_{\infty}^{b}) = 0$ implies
	\begin{align*}
		\dot{\Lambda}^{b}_{\infty}(\hat{y}_{t}^{b}) & = \dot{\Lambda}^{b}_{\infty}(\hat{y}_{t}^{b}) - \dot{\Lambda}^{b}_{\infty}(y_{\infty}^{b}) \\
		& = \ddot{\Lambda}_{\infty}(y_{\infty}^{b}) (\hat{y}_{t}^{b} - y_{\infty}^{b}) + o(\vertii{\hat{y}_{t}^{b} - y_{\infty}^{b}}),
	\end{align*}
	where the little-o term holds uniformly across $b \in B_{\delta}(\beta)$. Accordingly, we can set $\epsilon$ small enough so that $ y_{t}^{b} \in B_{2\epsilon}(y_{\infty}^{\beta}) $ implies
	\begin{align*}
		\vertii{\dot{\Lambda}^{b}_{\infty}(\hat{y}_{t}^{b}) - \ddot{\Lambda}_{\infty}(y_{\infty}^{b}) (\hat{y}_{t}^{b} - y_{\infty}^{b})} \le \sigma_{m}^{\beta} \vertii{\hat{y}_{t}^{b} - y_{\infty}^{b}}/4,
	\end{align*}
	for all $b \in B_{\delta}(\beta)$. Now combining these last two results yields the following, for $ y_{t}^{b} \in B_{2\epsilon}(y_{\infty}^{\beta}) $:
	\begin{align*}
		(\hat{y}_{t}^{b} -& y_{\infty}^{b})' \dot{\Lambda}^{b}_{\infty}(\hat{y}_{t}^{b})\\
		& = (\hat{y}_{t}^{b} - y_{\infty}^{b})'\ddot{\Lambda}_{\infty}(y_{\infty}^{b}) (\hat{y}_{t}^{b} - y_{\infty}^{b}) \\
		& \quad + (\hat{y}_{t}^{b} - y_{\infty}^{b})'\big(\dot{\Lambda}^{b}_{\infty}(\hat{y}_{t}^{b}) - \ddot{\Lambda}_{\infty}(y_{\infty}^{b}) (\hat{y}_{t}^{b} - y_{\infty}^{b})\big)\\
		& \ge \sigma_{m}^{\beta} \vertii{\hat{y}_{t}^{b} - y_{\infty}^{b}}^{2}/2 - \vertii{\hat{y}_{t}^{b} - y_{\infty}^{b}}\vertii{\dot{\Lambda}^{b}_{\infty}(\hat{y}_{t}^{b}) - \ddot{\Lambda}_{\infty}(y_{\infty}^{b}) (\hat{y}_{t}^{b} - y_{\infty}^{b})}\\
		& \ge \sigma_{m}^{\beta} \vertii{\hat{y}_{t}^{b} - y_{\infty}^{b}}^{2}/4 .
	\end{align*}
	And combining this with $(\hat{y}_{t}^{b} - y_{\infty}^{b})'\dot{\Lambda}_{t}^{b}(\hat{y}_{t}^{b})= (y_{t}^{b} - \hat{y}_{t}^{b})'\dot{\Lambda}_{t}^{b}(\hat{y}_{t}^{b}) \le 0$, which we get from Lemma \ref{l:yinh}, yields the following, for $ y_{t}^{b} \in B_{2\epsilon}(y_{\infty}^{\beta}) $:
	\begin{align*}
		\vertii{\hat{y}_{t}^{b} - y_{\infty}^{b}}& \vertii{\dot{\Lambda}^{b}_{\infty}(\hat{y}_{t}^{b}) - \dot{\Lambda}_{t}^{b}(\hat{y}_{t}^{b})}\\
		& \ge
		(\hat{y}_{t}^{b} - y_{\infty}^{b})' (\dot{\Lambda}^{b}_{\infty}(\hat{y}_{t}^{b}) - \dot{\Lambda}_{t}^{b}(\hat{y}_{t}^{b})) \\
		& \ge \sigma_{m}^{\beta} \vertii{\hat{y}_{t}^{b} - y_{\infty}^{b}}^{2}/4 .
	\end{align*}
	Hence, $ y_{t}^{b} \in B_{2\epsilon}(y_{\infty}^{\beta}) $ implies 
	\begin{align*}
		\vertii{\dot{\Lambda}^{b}_{\infty}(\hat{y}_{t}^{b}) - \dot{\Lambda}_{t}^{b}(\hat{y}_{t}^{b})}
		& \ge \sigma_{m}^{\beta} \vertii{\hat{y}_{t}^{b} - y_{\infty}^{b}}/4 = \sigma_{m}^{\beta} \vertii{y_{t}^{b} - y_{\infty}^{b}}/8 .
	\end{align*}
	And thus, we have
	\begin{align*}
		\E\Big(&\sup_{b \in B_{\delta}(\beta)}\indicator{y_{t}^{b}\in B_{\epsilon}(y_{\infty}^{b})}\vertii{y_{t}^{b} - y_{\infty}^{b}}^{2}\Big) \\
		& \le \E\Big(\sup_{b \in B_{\delta}(\beta)}\indicator{y_{t}^{b} \in B_{2\epsilon}(y_{\infty}^{\beta})} \vertii{y_{t}^{b} - y_{\infty}^{b}}^{2}\Big) \n
		& \le (8/\sigma_{m}^{\beta})^{2} \E\Big(\sup_{b \in B_{\delta}(\beta)}\indicator{y_{t}^{b} \in B_{2\epsilon}(y_{\infty}^{\beta})} \vertii{\dot{\Lambda}^{b}_{t}(\hat{y}_{t}^{b}) - \dot{\Lambda}^{b}_{\infty}(\hat{y}_{t}^{b})}^{2}\Big) \n 
		& \le (8/\sigma_{m}^{\beta})^{2} \E\Big(\sup_{b \in B_{\delta}(\beta)}\sup_{y \in B_{2\epsilon}(y_{\infty}^{\beta})} \vertii{\dot{\Lambda}^{b}_{t}(y) - \dot{\Lambda}^{b}_{\infty}(y)}^{2}\Big)\\
		& = (8/\sigma_{m}^{\beta})^{2} \E\Big(\sup_{y \in B_{2\epsilon}(y_{\infty}^{\beta})} \vertii{\dot{\Lambda}^{\beta}_{t}(y) - \dot{\Lambda}^{\beta}_{\infty}(y)}^{2}\Big),
	\end{align*}
	where the last line holds because $\dot{\Lambda}^{b}_{t} - \dot{\Lambda}^{b}_{\infty}$ is independent of $b$. Finally, Lemma \ref{l:boundOnSquare} establishes that the expectation in the last line is less than $C/t$, for some universal constant $ C > 0 $.
\end{proof}

\begin{proof}[Proposition \ref{p:OmegaOneOverT} Proof]
	This follows immediately from Lemma \ref{l:ballY}.
\end{proof}

\begin{proof}[Corollary \ref{c:OmegaOneOverT} Proof]
	This follows from Proposition \ref{p:withinBallBound} and Lemma \ref{l:ballY}.
\end{proof}

\begin{proof}[Proposition \ref{p:outsideBallBound} Proof]
	The proof hinges on two key results. The first result is that there exists $ \delta, C > 0 $ such that 
	\begin{align}
		\Pr\big(\sup_{b \in B_{\delta}(\beta)} \vertii{y_{t}^{b} - y_{\infty}^{b}} > \epsilon\big) \le 4m^{2}\exp(-C\epsilon^{2}t), \label{eq:resultToEstablish1}
	\end{align}
	for all $ t \in \mathbbm{N} $ and sufficiently small $ \epsilon > 0 $. The second result is that for all sufficiently large $ \gamma > 0 $ there exists $ \delta, C > 0 $ such that 
	\begin{align}
		\E\big(\sup_{b \in B_{\delta}(\beta)}\indicator{y_{t}^{b} \notin B_{\gamma^{1/p}}(0)}\vertii{y_{t}^{b}}^{p}\big) \le \exp(-Ct), \label{eq:resultToEstablish2}
	\end{align}
	for all sufficiently large $ t $.  
	
	The $p = 0$ case follows immediately from the line \eqref{eq:resultToEstablish1}. Deriving the $p > 0$ case from lines \eqref{eq:resultToEstablish1} and \eqref{eq:resultToEstablish2} will take a bit more work. To that end, choose $\gamma $ large enough so that $\gamma \ge \sup_{b \in B_{\delta}(\beta)} \vertii{y_{\infty}^{b}}^{p}$, and hence $\vertii{y_{t}^{b} - y_{\infty}^{b}} \le \vertii{y_{t}^{b}} + \gamma^{1/p}$ (Lemma \ref{l:lipschitzYinB} establishes that this is possible). And with this, lines \eqref{eq:resultToEstablish1} and \eqref{eq:resultToEstablish2} imply that we can choose $C > 0$ so that we have the following for all sufficiently small $\epsilon$ and large $t$:
	\begin{align*}
		\E\big(&\sup_{b \in B_{\delta}(\beta)}\indicator{y_{t}^{b} \notin B_{\epsilon}(y_{\infty}^{b})} \vertii{y_{t}^{b} - y_{\infty}^{b}}^{p}\big) \n
		& \le \E\big(\sup_{b \in B_{\delta}(\beta)}\indicator{y_{t}^{b} \notin B_{\epsilon}(y_{\infty}^{b})} \indicator{y_{t}^{b} \notin B_{\gamma^{1/p}}(0)} (\vertii{y_{t}^{b}} + \gamma^{1/p})^{p}\big) \n
		& \qquad + \E\big(\sup_{b \in B_{\delta}(\beta)}\indicator{y_{t}^{b} \notin B_{\epsilon}(y_{\infty}^{b})} \indicator{y_{t}^{b} \in B_{\gamma^{1/p}}(0)} (\vertii{y_{t}^{b}} + \gamma^{1/p})^{p}\big)\n
		& \le \E\big(\sup_{b \in B_{\delta}(\beta)}\indicator{y_{t}^{b} \notin B_{\epsilon}(y_{\infty}^{b})} \indicator{y_{t}^{b} \notin B_{\gamma^{1/p}}(0)} 2^{p}\vertii{y_{t}^{b}}^{p}\big) \n
		& \qquad + \E\big(\sup_{b \in B_{\delta}(\beta)}\indicator{y_{t}^{b} \notin B_{\epsilon}(y_{\infty}^{b})} \indicator{y_{t}^{b} \in B_{\gamma^{1/p}}(0)}2^{p}\gamma\big)\n
		& \le 2^{p} \E\big(\sup_{b \in B_{\delta}(\beta)}\indicator{y_{t}^{b} \notin B_{\gamma^{1/p}}(0)} \vertii{y_{t}^{b}}^{p}\big) + 2^{p}\gamma \Pr\big(\sup_{b \in B_{\delta}(\beta)} \vertii{y_{t}^{b} - y_{\infty}^{b}} > \epsilon\big)\n
		& \le 2^{p} \exp(-Ct) + 2^{p+2} \gamma m^{2}\exp(-C\epsilon^{2}t).
	\end{align*}
	The inequality above establishes the $p > 0$ case. Hence, proving lines \eqref{eq:resultToEstablish1} and \eqref{eq:resultToEstablish2} will complete the argument.
	
	Before getting into the math, let me roughly sketch the proof of line \eqref{eq:resultToEstablish1}. The key tool will be Lemma \ref{l:boxThaty}, which is our only means for positioning $y_{t}^{b}$. The lemma corresponds to a set of inequalities that describe a small box, which is roughly aligned with the orthonormal basis $\{\omega_{j}^{b}\}_{i = 1}^{m}$; if these inequalities all hold, then the box is intact, and $y_{t}^{b}$ resides inside of it. I will use this result to bound the distance between $y_{t}^{b}$ and $y_{\infty}^{b}$ with the distances between $\dot{\Lambda}^{b}_{t}(y_{\infty}^{b} + \eta k \omega_{j}^{b})$ and $\eta k \sigma_{j}^{b}\omega_{j}^{b}$, for $j \in m$, $k \in \{-1, 1\}$, and $\eta > 0$ (these latter distances being the constraints that ensure the integrity of the box). This reframing simplifies the problem, because $\dot{\Lambda}^{b}_{t}(y_{\infty}^{b} + \eta k \omega_{j}^{b})$ is a sum of \iid\ bounded variables. The second part of the proof replaces the $\eta k \sigma_{j}^{b} \omega_{j}^{b}$ term in our distance measurements with with $\dot{\Lambda}^{b}_{\infty}(y_{\infty}^{b} + \eta k \omega_{j}^{b})$. This step is useful because $\dot{\Lambda}^{b}_{t}(y_{\infty}^{b} + \eta k \omega_{j}^{b}) - \dot{\Lambda}^{b}_{\infty}(y_{\infty}^{b}+ \eta k \omega_{j}^{b})$ is an empirical process. The final part of the proof invokes a standard empirical process result to establish the desired concentration of measure.
	
	To begin the proof of line \eqref{eq:resultToEstablish1}, note that Lemmas \ref{l:defineDotNew} and \ref{l:lipschitzYinB} imply that we can choose $\delta > 0$ and $\epsilon > 0$ small enough to ensure the existence and continuity of $\ddot{\Lambda}_{\infty}$ between $y_{\infty}^{b}$ and $y_{\infty}^{b} + \eta k \omega_{j}^{b}$, and small enough to ensure that $\sigma_{1}^{b} \le 2 \sigma_{1}^{\beta}$, $\sigma_{m}^{b} \ge \sigma_{m}^{\beta}/2 > 0$, and $y_{\infty}^{b} + \eta k \omega_{j}^{b} \ge 0$, for all $j \in [m]$, $k \in \{-1, 1\}$, $b \in B_{\delta}(\beta)$, and $\eta \equiv \epsilon /(1 + 8\sqrt{m}\sigma_{1}^{\beta}/\sigma_{m}^{\beta})$. Now, with these conditions, we can use Lemma \ref{l:boxThaty} to bound the left-hand side of \eqref{eq:resultToEstablish1} in terms of more amenable subgradients:
	\begin{align}
		\Pr\big(&\sup_{b \in B_{\delta}(\beta)} \vertii{y_{t}^{b} - y_{\infty}^{b}} > \epsilon\big) \n
		&\le \Pr\big(\sup_{b \in B_{\delta}(\beta)} \vertii{y_{t}^{b} - y_{\infty}^{b}} > \eta(1 + 2\sqrt{m}\sigma_{1}^{b}/\sigma_{m}^{b})\big)\n
		& \le \Pr\big(\sup_{b \in B_{\delta}(\beta)}\max_{j \in [m]}\max_{k \in \{-1, 1\}}\vertii{\dot{\Lambda}^{b}_{t}(y_{\infty}^{b} + \eta k \omega_{j}^{b}) - \eta k \sigma_{j}^{b} \omega_{j}^{b}} - \eta\sigma_{m}^{b}/(2\sqrt{m}) > 0\big)\n
		& \le \Pr\big(\sup_{b \in B_{\delta}(\beta)}\max_{j \in [m]}\max_{k \in \{-1, 1\}}\vertii{\dot{\Lambda}^{b}_{t}(y_{\infty}^{b} + \eta k \omega_{j}^{b}) - \eta k \sigma_{j}^{b} \omega_{j}^{b}} - \eta\sigma_{m}^{\beta}/(4\sqrt{m}) > 0\big). \label{eq:moreAmenableSubgradient}
	\end{align}
	
	Now I will frame the last expression above as an empirical process by replacing the $\eta k \sigma_{j}^{b} \omega_{j}^{b}$ term with $\dot{\Lambda}^{b}_{\infty}(y_{\infty}^{b} + \eta k \omega_{j}^{b})$. To this end, note that the mean value theorem indicates that there exists $\xi \in (0, \eta)$ for which
	\begin{align*}
		\dot{\Lambda}^{b}_{\infty}(y_{\infty}^{b} + \eta k \omega_{j}^{b}) & = \dot{\Lambda}^{b}_{\infty}(y_{\infty}^{b} + \eta k \omega_{j}^{b}) - 0 \\
		& = \dot{\Lambda}^{b}_{\infty}(y_{\infty}^{b} + \eta k \omega_{j}^{b}) - \dot{\Lambda}^{b}_{\infty}(y_{\infty}^{b}) \\
		& = \eta k \ddot{\Lambda}_{\infty}(y_{\infty}^{b} + \xi k \omega_{j}^{b}) \omega_{j}^{b} \\
		& = \eta k \ddot{\Lambda}_{\infty}(y_{\infty}^{b}) \omega_{j}^{b} + \eta k(\ddot{\Lambda}_{\infty}(y_{\infty}^{b} + \xi k \omega_{j}^{b}) - \ddot{\Lambda}_{\infty}(y_{\infty}^{b}))\omega_{j}^{b}\\
		& = \eta k \sigma_{j}^{b} \omega_{j}^{b} + o(\eta),
	\end{align*}
	where the little-o term holds uniformly across $b \in B_{\delta}(\beta)$.  Accordingly, we can set $\epsilon$ small enough so that $\sup_{b \in B_{\delta}(\beta)}\vertii{\dot{\Lambda}^{b}_{\infty}(y_{\infty}^{b} + \eta k \omega_{j}^{b}) - \eta k \sigma_{j}^{b} \omega_{j}^{b}} \le \eta\sigma_{m}^{\beta}/(8 \sqrt{m})$, in which case we have
	\begin{align*}
		&\vertii{\dot{\Lambda}^{b}_{t}(y_{\infty}^{b} + \eta k \omega_{j}^{b}) - \eta k \sigma_{j}^{b} \omega_{j}^{b}} \\
		& \quad \le 
		\vertii{\dot{\Lambda}^{b}_{t}(y_{\infty}^{b} + \eta k \omega_{j}^{b}) - \dot{\Lambda}^{b}_{\infty}(y_{\infty}^{b}+ \eta k \omega_{j}^{b})} + \vertii{\dot{\Lambda}^{b}_{\infty}(y_{\infty}^{b} + \eta k \omega_{j}^{b}) - \eta k \sigma_{j}^{b} \omega_{j}^{b}}\\
		& \quad \le 	\vertii{\dot{\Lambda}^{b}_{t}(y_{\infty}^{b} + \eta k \omega_{j}^{b}) - \dot{\Lambda}^{b}_{\infty}(y_{\infty}^{b}+ \eta k \omega_{j}^{b})} + \eta\sigma_{m}^{\beta}/(8 \sqrt{m}).
	\end{align*}
	And, finally, combining this with line \eqref{eq:moreAmenableSubgradient} and the fact that $\dot{\Lambda}_{t}^{b} - \dot{\Lambda}_{t}^{b} = \dot{\Lambda}_{t}^{\beta} - \dot{\Lambda}_{t}^{\beta}$ yields the following:
	\begin{align*}
		\Pr&\big(\sup_{b \in B_{\delta}(\beta)} \vertii{y_{t}^{b} - y_{\infty}^{b}} > \epsilon\big)\\
		& \le \Pr\big(\sup_{b \in B_{\delta}(\beta)}\max_{j \in [m]}\max_{k \in \{-1, 1\}}\vertii{\dot{\Lambda}^{b}_{t}(y_{\infty}^{b} + \eta k \omega_{j}^{b}) - \eta k \sigma_{j}^{b} \omega_{j}^{b}}> \eta\sigma_{m}^{\beta}/(4\sqrt{m})\big) \\
		& \le \Pr\big(\sup_{b \in B_{\delta}(\beta)}\max_{j \in [m]}\max_{k \in \{-1, 1\}}\vertii{\dot{\Lambda}^{b}_{t}(y_{\infty}^{b} + \eta k \omega_{j}^{b}) - \dot{\Lambda}^{b}_{\infty}(y_{\infty}^{b}+ \eta k \omega_{j}^{b})}> \eta\sigma_{m}^{\beta}/(8\sqrt{m})\big)\\
		& \le \sum_{i = 1}^{m} \sum_{j = 1}^{m}\sum_{k \in \{-1, 1\}}\Pr\big(\sup_{b \in B_{\delta}(\beta)}\verti{e_{i}'\dot{\Lambda}^{b}_{t}(y_{\infty}^{b} + \eta k \omega_{j}^{b}) - e_{i}'\dot{\Lambda}^{b}_{\infty}(y_{\infty}^{b}+ \eta k \omega_{j}^{b})} \ge \eta\sigma_{m}^{\beta}/(8m)\big)\\
		& \le \sum_{i = 1}^{m} \sum_{j = 1}^{m}\sum_{k \in \{-1, 1\}}\Pr\big(\sup_{y \in B_{\nu}(y_{\infty}^{\beta})}\verti{e_{i}'\dot{\Lambda}^{\beta}_{t}(y) - e_{i}'\dot{\Lambda}^{\beta}_{\infty}(y)} \ge \eta\sigma_{m}^{\beta}/(8m)\big),
	\end{align*}
	where $\nu > 0$ is a constant that's large enough to ensure that $y_{\infty}^{b} + \eta k \omega_{j}^{b} \in B_{\nu}(y_{\infty}^{\beta})$ for all $b \in B_{\delta}(\beta)$. Finally, Theorem 2.14.9 of \cite{VanderVaart1996} implies that this last expression falls exponentially fast in $t$ (see the proof of lemma \ref{l:boundOnSquare} for confirmation of this theorem's hypothesis.) 
	
	This establishes line \eqref{eq:resultToEstablish1}, which establishes the $p = 0$ case. I will now prove line \eqref{eq:resultToEstablish2}, assuming $p > 0$. The proof will proceed as follows: First, I will bound the probability that $\vertii{y_{t}^{b}}^{p}$ exceeds some $\gamma > 0$ with the probability that $e_{j}'\dot{\Lambda}_{t}^{b}( e_{j} \gamma^{1/p}/\sqrt{m})$ is negative. This latter random variable is easier to work with because it is a sum of \iid\ random variables. Second, I will lower bound $e_{j}'\dot{\Lambda}_{t}^{b}( e_{j}\gamma^{1/p}/\sqrt{m})$ with a binomial random variable, with success probability $\rho_{\gamma} \equiv \Pr(u_{1} > \eta \gamma^{1/p} / (2\sqrt{m}))$. This characterization will enable me to use the binomial Chernoff bound to establish that $\Pr(\vertii{y_{t}^{b}}^{p} > \gamma)$ falls exponentially fast in $t$. And finally, I will integrate over this tail bound to create a corresponding expectation bound.
	
	To begin the proof, I will show that $ e_{j}'y_{t}^{b} > \omega $ implies $ e_{j}'\dot{\Lambda}_{t}^{b}(\omega e_{j}) \le 0 $, for $\omega \in \mathbbm{R}$. To see this, take $e_{j}'y_{t}^{b} \ge \omega$ and $ \hat{y} \equiv y_{t}^{b} - e_{j} (e_{j}'y_{t}^{b} - \omega)/2 $, and apply Lemma \ref{l:yinh}: 
	\begin{align*}
		0 \ge&  (y_{t}^{b} - \hat{y})'\dot{\Lambda}_{t}^{b}(\hat{y}) \\
		= &  ((e_{j}'y_{t}^{b} - \omega)/2) e_{j}'\dot{\Lambda}_{t}^{b}(\hat{y}) \\
		= & ((e_{j}'y_{t}^{b} - \omega)/2)e_{j}'\big(b - \sum_{s=1}^{t} \indicator{\Delta_{s}(\hat{y}) > 0} a_{s}/t\big) \n
		\ge &((e_{j}'y_{t}^{b} - \omega)/2) e_{j}'\big(b - \sum_{s=1}^{t} \indicator{u_{s} > a_{s}'e_{j}e_{j}' \hat{y}}a_{s}/t\big) \n
		\ge &((e_{j}'y_{t}^{b} - \omega)/2) e_{j}'\big(b - \sum_{s=1}^{t}\indicator{u > a'e_{j}\omega}a_{s}/t\big) \n
		= & ((e_{j}'y_{t}^{b} - \omega)/2) e_{j}'\dot{\Lambda}_{t}^{b}(\omega e_{j}) .
	\end{align*}
	Since $e_{j}'y_{t}^{b} - \omega$ is positive, by assumption, it follows that $e_{j}'\dot{\Lambda}_{t}^{b}(\omega e_{j})$ must be non-positive. 
	
	And now, I'll use this result to replace the shadow price with a simpler subgradient:
	\begin{align*}
		\Pr\big(\sup_{b \in B_{\delta}(\beta)}\vertii{y_{t}^{b}}^{p} > \gamma\big) & \le \sum_{j=1}^{m} \Pr\big(\sup_{b \in B_{\delta}(\beta)}e_{j}'y_{t}^{b} > \gamma^{1/p}/\sqrt{m}\big) \n
		& \le \sum_{j=1}^{m}\Pr\big(\sup_{b \in B_{\delta}(\beta)}e_{j}'\dot{\Lambda}_{t}^{b}( e_{j} \gamma^{1/p}/\sqrt{m}) \le 0\big). 
	\end{align*}	
	
	Next, we will bound the complex random variable in the last probability above with a simple binomial random variable. To that end, choose $\delta, \eta > 0$ so that $\eta \le e_{j}'b$ for all $b \in B_{\delta}(\beta)$, in which case we have the following:
	\begin{align*}
		\sup_{b \in B_{\delta}(\beta)} & e_{j}'\dot{\Lambda}_{t}^{b}( e_{j}\gamma^{1/p}/\sqrt{m}) \\
		& = \sup_{b \in B_{\delta}(\beta)} e_{j}'b - \sum_{s=1}^{t}\indicator{u_{s} > a_{s}'e_{j} \gamma^{1/p}/\sqrt{m})} e_{j}'a_{s}/t \n
		& \ge \sup_{b \in B_{\delta}(\beta)} e_{j}'b - \sum_{s=1}^{t} \Big(\indicator{e_{j}'a_{s} \le e_{j}'b/2} (e_{j}'b/2)/t + \indicator{e_{j}'a_{s} > e_{j}'b/2}\indicator{u_{s} > a_{s}'e_{j} \gamma^{1/p}/\sqrt{m})}\alpha/t\Big) \n
		& \ge \sup_{b \in B_{\delta}(\beta)} e_{j}'b/2 - \sum_{s=1}^{t} \indicator{u_{s} > e_{j}'b \gamma^{1/p}/(2\sqrt{m})} \alpha/t\\
		& \ge \eta/2 - \xi_{t} \alpha/t,
	\end{align*}
	where $\xi_{t} \equiv \sum_{s=1}^{t} \indicator{u_{s} > \eta \gamma^{1/p}/(2\sqrt{m})}$ is a binomial$(t, \rho_{\gamma})$, with $\rho_{\gamma} \equiv \Pr(u_{1} > \eta \gamma^{1/p}/(2\sqrt{m}))$. Further, since $\E(u_{1}) \le \infty$, we must have $\rho_{\gamma} \le \gamma^{-1/p}$, for sufficiently large $\gamma$. Hence, combining the previous two results with the binomial Chernoff bound yields the following for sufficiently large $\gamma$:
	\begin{align*}
		\Pr\big(\sup_{b \in B_{\delta}(\beta)}\vertii{y_{t}^{b}}^{p} > \gamma\big) & \le \sum_{j=1}^{m} \Pr(\eta/2 - \xi_{t} \alpha/t \le 0) \\
		& = m \Pr(\xi_{t} \ge t\eta/(2\alpha))\\
		& \le m \exp\Big(-\frac{t\eta}{2\alpha}\Big(\log\frac{\eta}{2 \alpha\rho_{\gamma}} - 1\Big)\Big) \\
		& \le m \exp\Big(-\frac{t\eta}{4\alpha}\log\frac{\eta \gamma^{1/p}}{2 \alpha}\Big) \\
		& = m (\eta/(2\alpha))^{\frac{-t\eta}{4\alpha}}\gamma^{\frac{-t\eta}{4p\alpha}},
	\end{align*}
	where the penultimate line supposes that $\gamma$ is large enough to satisfy $\log(\frac{\eta \gamma^{1/p}}{2 \alpha})/2 \ge 1$. Now choosing $\gamma$ large enough to satisfy the previous result and large enough to ensure that $\vertii{y_{t}^{b}}^{p} \ge \gamma$ implies $y_{t}^{b} \notin B_{\epsilon}(y_{\infty}^{b})$ yields the following:
	\begin{align*}
		\E(&\sup_{b \in B_{\delta}(\beta)}\indicator{y_{t}^{b} \notin B_{\gamma^{1/p}}(0)}\vertii{y_{t}^{b}}^{p}) \\
		& \le \gamma\Pr\big(\sup_{b \in B_{\delta}(\beta)}\vertii{y_{t}^{b}}^{p} \ge \gamma\big) + \int_{x=\gamma}^{\infty} \Pr\big(\sup_{b \in B_{\delta}(\beta)}\vertii{y_{t}^{b}}^{p} > x\big) dx \n
		& \le \gamma\Pr\big(\sup_{b \in B_{\delta}(\beta)}\vertii{y_{t}^{b} - y_{\infty}^{b}} \ge \epsilon\big)  +  \int_{x=\gamma}^{\infty} m (\eta/(2\alpha))^{\frac{-t\eta}{4\alpha}}x^{\frac{-t\eta}{8\alpha}} dx \\
		& \le \gamma\Pr\big(\sup_{b \in B_{\delta}(\beta)}\vertii{y_{t}^{b} - y_{\infty}^{b}} \ge \epsilon\big) + \frac{m (\eta/(2\alpha))^{\frac{-t\eta}{4\alpha}}\gamma^{1 - \frac{t\eta}{8\alpha}}}{\frac{t\eta}{8\alpha} - 1}.
	\end{align*}
	The last expression above falls exponentially fast in $t$, by line \eqref{eq:resultToEstablish1}, so this establishes line \eqref{eq:resultToEstablish2}.
\end{proof}	

\begin{proof}[Corollary \ref{c:outsideBallBound} Proof]
	This is the $p = 0$ case of Proposition \ref{p:outsideBallBound}.
\end{proof}

\begin{proof}[Lemma \ref{l:concentrateMeasure} Proof]
	Consider an alternative martingale $\{\hat{b}_{t}\}_{t = n}^{1}$ in which $\hat{b}_{n} \equiv b_{n}$ and
	\begin{align*}
		\hat{b}_{t} \equiv 
		\begin{cases}
			b_{t} & \hat{b}_{t+1} \in B_{\delta}(\beta),\\
			\hat{b}_{t+1} & \hat{b}_{t+1} \notin B_{\delta}(\beta).
		\end{cases} 
	\end{align*}
	In other words, $\hat{b}_{t}$ tracks $b_{t}$ until the first time that $b_{t}$ departs $B_{\delta}(\beta)$, at which point $\hat{b}_{t}$ remains frozen in place. By design, $\hat{b}_{t} \in B_{\delta}(\beta)$ implies $b_{t} \in B_{\delta}(\beta)$, and hence $\Pr(b_{t} \notin B_{\delta}(\beta)) \le \Pr(\hat{b}_{t} \notin B_{\delta}(\beta))$.
	
	And, with this, the result follows from the Azuma–Hoeffding inequality, since $\vertii{\hat{b}_{t} - \hat{b}_{t+1}} \le (\vertii{\beta} + \delta + \vertii{\alpha})/t$:
	\begin{align}
		\Pr(& b_{t} \notin B_{\delta}(\beta)) \n
		& \le \sup_{N \ge t} \Pr(\hat{b}_{t} \notin B_{\delta}(\beta))) \n
		& \le \sup_{N \ge t} \sum_{j=1}^{m} \Pr\big(\verti{e_{j}'\hat{b}_{t} - e_{j}'b_{n}} \ge \delta/\sqrt{m}\big) \n
		& \le \sup_{N \ge t} 2 m\exp\Big(-\frac{\delta^{2}/m}{2\sum_{s=t}^{N-1} (\vertii{\beta} + \delta + \vertii{\alpha})^{2} / s^2}\Big) \n
		& < 2 m\exp\Big(-\frac{\delta^{2}}{2m (\vertii{\beta} + \delta + \vertii{\alpha})^{2} \int_{s=t-1}^{\infty} ds s^2} \Big) \n
		& < 2 m\exp\Big(-\frac{\delta^{2}(t-1)}{2m(\vertii{\beta} + \delta + \vertii{\alpha})^{2}} \Big). \label{eq:lastLineOfLemma}
	\end{align}
\end{proof}

\begin{proof}[Corollary \ref{c:concentrateMeasure} Proof]
	Let $\{\hat{b}_{t}\}_{t = n}^{1}$ be the alternative martingale defined in the proof of Lemma \ref{l:concentrateMeasure}. Note that we have $\hat{b}_{t} \notin B_{\delta}(\beta)$ if and only if $t\le \tau(\delta)$. And, with this, line \eqref{eq:lastLineOfLemma} implies the result:	
	\begin{align*}
		\E(\tau(\delta)) & = \sum_{s = 1}^{n} \Pr(\tau(\delta) \ge s)\\
		& = \sum_{t = 1}^{n} \Pr(\hat{b}_{t} \notin B_{\delta}(\beta))\\
		& < \sum_{t = 1}^{\infty}2 m\exp\Big(-\frac{\delta^{2}(t-1)}{2m(\vertii{\beta} + \delta + \vertii{\alpha})^{2}} \Big)\\
		& = O(1).
	\end{align*}
\end{proof}

\begin{proof}[Lemma \ref{l:MyopicRegretFirst} Proof]
	Bounding the first term of the myopic regret is simple: Since $\indicator{b_{t} \notin B_{\delta/2}(\beta)}$ is independent of $\sum_{s=1}^{t} u_{s}$, Lemma \ref{l:concentrateMeasure} indicates that there exists $ C > 0 $ for which
	\begin{align*}
		\E\big(& \indicator{b_{t} \notin B_{\delta/2}(\beta)}\sum_{s=1}^{t} u_{s}\big) \\
		& = \Pr(b_{t} \notin B_{\delta/2}(\beta)) \sum_{s=1}^{t} \E(u_{s})\\
		& \le \exp(-Ct) t \E(u_{1}) \\
		& = o(1/t).
	\end{align*}
	
	I will now bound the second term of the myopic regret with Propositions \ref{p:withinBallBound} and \ref{p:outsideBallBound}, and Corollary \ref{c:outsideBallBound}. Doing so will take several steps. First, the independence of $b_{t}$ from the random mapping $b \mapsto \indicator{\Delta_{t}(y_{\infty}^{b}) > 0}\Delta_{t}(y_{t-1}^{\psi_{t}^{b}(a_{t})})^{-}$ establishes that
	\begin{align*}
		\E\big(&\indicator{b_{t} \in B_{\delta/2}(\beta)}\indicator{\Delta_{t}(y_{\infty}^{b_{t}}) > 0}\Delta_{t}(y_{t-1}^{\psi_{t}^{b_{t}}(a_{t})})^{-}\big)\\
		& = \E\big(\indicator{b_{t} \in B_{\delta/2}(\beta)}\E\big(\indicator{\Delta_{t}(y_{\infty}^{b}) > 0}\Delta_{t}(y_{t-1}^{\psi_{t}^{b}(a_{t})})^{-}\big)\big|_{b = b_{t}}\big) . 
	\end{align*}
	Accordingly, it will suffice to show that 
	\begin{align}
		\sup_{b \in B_{\delta/2}(\beta)}\E\big(\indicator{\Delta_{t}(y_{\infty}^{b}) > 0}\Delta_{t}(y_{t-1}^{\psi_{t}^{b}(a_{t})})^{-}\big) = O(1/t) . \label{eq:ddddddd}
	\end{align}
	
	Next, I will remove the random $a_{t}$ in the superscript of $y_{t-1}^{\psi_{t}^{b}(a_{t})}$ by showing that the following holds, for all $b \in B_{\delta/2}(\beta)$ and large $t$:
	\begin{align}
		\indicator{\Delta_{t}(y_{\infty}^{b}) > 0} \Delta_{t}(y_{t-1}^{\psi_{t}^{b}(a_{t})})^{-} & \le \indicator{\Delta_{t}(y_{\infty}^{b}) > 0} \Delta_{t}(y_{\infty}^{b} + \xi_{t-1} \iota)^{-}, \label{eq:lineToCombine}
	\end{align}
	where $\xi_{t-1} \equiv \sup_{b \in B_{\delta}(\beta)}\vertii{y_{t-1}^{b} - y_{\infty}^{b}} + C/t$, for some constant $C$ that has yet to be defined. Note that the first term of $\xi_{t-1}$ appears compatible with our shadow price convergence results. Moreover, $\xi_{t-1}$, unlike $y_{t-1}^{\psi_{t}^{b}(a_{t})}$, is independent of the random function $\Delta_{t}$, which makes the right-hand side of \eqref{eq:lineToCombine} congruent with Lemma \ref{l:BoundEindicatorDelta2}. This lemma will convert $\xi_{t-1}$ into something like $\sup_{b \in B_{\delta}(\beta)}\vertii{y_{t}^{b} - y_{\infty}^{b}}^{2}$, which is precisely what we need to invoke Proposition \ref{p:withinBallBound}. But there will also be a few stray terms, which we'll bound with Proposition \ref{p:outsideBallBound} and Corollary \ref{c:outsideBallBound}.
	
	I will now derive line \eqref{eq:lineToCombine}. First, note that Assumption \ref{a:boundedA} and Lemma \ref{l:lipschitzYinB} imply there exits $C > 0$ that satisfies the following, almost surely:
	\begin{align*}
		\sup_{b \in B_{\delta/2}(\beta)} \vertii{y_{\infty}^{\psi_{t}^{b}(a_{t})} - y_{\infty}^{b}} \le C/t	.
	\end{align*}
	Accordingly, if we choose $t$ large enough so that $b \in B_{\delta/2}(\beta)$ implies $\psi_{t}^{b}(a_{t}) \in B_{\delta/2}(\beta)$, then we have the following:
	\begin{align*}
		\vertii{y_{t-1}^{\psi_{t}^{b}(a_{t})} - y_{\infty}^{b}} & \le \vertii{y_{t-1}^{\psi_{t}^{b}(a_{t})} - y_{\infty}^{\psi_{t}^{b}(a_{t})}} + \vertii{y_{\infty}^{\psi_{t}^{b}(a_{t})} - y_{\infty}^{b}}\\
		& \le \sup_{b \in B_{\delta}(\beta)}\vertii{y_{t-1}^{b} - y_{\infty}^{b}} + C/t\\
		& \equiv \xi_{t-1}. 
	\end{align*}
	And, by design, this new variable we've defined satisfies $y_{t-1}^{\psi_{t}^{b}(a_{t})} \le y_{\infty}^{b} + \xi_{t-1}\iota$, for $b \in B_{\delta/2}(\beta)$ and large $t$, which establishes line \eqref{eq:lineToCombine}.
	
	Next, I will apply Lemma \ref{l:BoundEindicatorDelta2} to the right-hand side of \eqref{eq:lineToCombine}. I can only do so when $y_{\infty}^{b}$ and $y_{\infty}^{b} + \xi_{t-1} \iota$ reside in the $\epsilon$-ball of $y_{\infty}^{\beta}$, for some sufficiently small $\epsilon$. Lemma \ref{l:lipschitzYinB} guarantees that the first vector can be thus situated, when $b \in B_{\delta/2}(\beta)$ and $\delta$ is small, but we have no such guarantee for the second vector. So we will have to consider the $y_{\infty}^{b} + \xi_{t-1} \iota \in B_{\epsilon}(y_{\infty}^{\beta})$ and $y_{\infty}^{b} + \xi_{t-1} \iota \notin B_{\epsilon}(y_{\infty}^{\beta})$ cases separately:
	\begin{align*}
		\E\big(&\indicator{\Delta_{t}(y_{\infty}^{b}) > 0} \Delta_{t}(y_{\infty}^{b} + \xi_{t-1} \iota)^{-} \big)\\
		& = \E\big(\E\big(\indicator{y_{\infty}^{b} + \xi_{t-1} \iota \in B_{\epsilon}(y_{\infty}^{\beta})}\indicator{\Delta_{t}(y_{\infty}^{b}) > 0} \Delta_{t}(y_{\infty}^{b} + \xi_{t-1} \iota)^{-}  \colonBreak \xi_{t-1}\big)\big) \\
		& \quad + \E\big(\indicator{y_{\infty}^{b} + \xi_{t-1} \iota \notin B_{\epsilon}(y_{\infty}^{\beta})}\indicator{\Delta_{t}(y_{\infty}^{b}) > 0} \Delta_{t}(y_{\infty}^{b} + \xi_{t-1} \iota)^{-}\big) \\
		& \le \E\big(\indicator{y_{\infty}^{b} + \xi_{t-1} \iota \in B_{\epsilon}(y_{\infty}^{\beta})}2\sigma_{1}^{\beta}\vertii{\xi_{t-1} \iota}^{2}\big) \\
		& \quad + \E\big(\indicator{y_{\infty}^{b} + \xi_{t-1} \iota \notin B_{\epsilon}(y_{\infty}^{\beta})} (\vertii{\alpha}\vertii{y_{\infty}^{b}} + \vertii{\alpha} \vertii{\xi_{t-1} \iota})\big) \\
		& \le 2\sigma_{1}^{\beta}\vertii{\iota}^{2} \E(\xi_{t-1}^{2}) + \vertii{\alpha}\vertii{y_{\infty}^{b}} \Pr(y_{\infty}^{b} + \xi_{t-1} \iota \notin B_{\epsilon}(y_{\infty}^{\beta})) + 
		\\
		& \quad + \vertii{\alpha} \vertii{\iota}
		\E\big(\indicator{y_{\infty}^{b} + \xi_{t-1} \iota \notin B_{\epsilon}(y_{\infty}^{\beta})}\xi_{t-1}\big)\\
		& \le 2\sigma_{1}^{\beta}\vertii{\iota}^{2} \E(\xi_{t-1}^{2}) + 2\vertii{\alpha}\vertii{y_{\infty}^{\beta}} \Pr(\xi_{t-1} > \epsilon/(2 \sqrt{m})) + 
		\\
		& \quad + \vertii{\alpha} \vertii{\iota}
		\E\big(\indicator{\xi_{t-1} > \epsilon/(2 \sqrt{m})}\xi_{t-1}\big).
	\end{align*}
	For the last line, I suppose that $\delta$ is small enough so that (i) $\sup_{b \in B_{\delta/2}(\beta)} \vertii{y_{\infty}^{b}} \le 2 \vertii{y_{\infty}^{\beta}}$ and (ii) $\{y_{\infty}^{b} \colonBreak b \in B_{\delta/2}(\beta)\} \subset B_{\epsilon/2}(y_{\infty}^{\beta})$, in which case $y_{\infty}^{b} + \xi_{t-1} \iota \notin B_{\epsilon}(y_{\infty}^{\beta})$ implies $\xi_{t-1} > \epsilon/(2\sqrt{m})$. Further, Proposition \ref{p:withinBallBound} establishes that the first term in the last line above is $O(1/t)$, Corollary \ref{c:outsideBallBound} establishes that the second term is $o(1/t)$, and Proposition \ref{p:outsideBallBound} establishes that the third term is $o(1/t)$. Hence, the last expression---which holds for all $b \in B_{\delta/2}(\beta)$---is $O(1/t)$, which with line \eqref{eq:lineToCombine} establishes line \eqref{eq:ddddddd}. 
	
	Finally, the same argument yields the same bound for the third term of myopic regret. 
\end{proof}

\begin{proof}[Lemma \ref{l:concentrateMeasureLearning} Proof]
	Let $\{\hat{b}_{t}\}_{t = n}^{1}$ denote the inventory process defined in the proof of Lemma \ref{l:concentrateMeasure}, but derived from from Algorithm \ref{alg:upperLearning}'s $b_{t}$ values. Just to remind you, the $\{\hat{b}_{t}\}_{t = n}^{1}$ process tracks the $\{b_{t}\}_{t = n}^{1}$ process until time $\tau(\delta)$---i.e., until Algorithm \ref{alg:upperLearning}'s $b_{t}$ values first depart $B_{\delta}(\beta)$---at which point the process freezes in place. The $\{\hat{b}_{t}\}_{t = n}^{1}$ process will be easier to study because a constant multiple of $t$ bounds its innovations. And since $\hat{b}_{t} \in B_{\delta}(\beta)$ implies $b_{t} \in B_{\delta}(\beta)$, it will suffice to establish the concentration of measure for $\hat{b}_{t}$.
	
	I will bound the distance between $\hat{b}_{t}$ and $\beta$ with the following inequality:
	\begin{align}
		\vertii{\hat{b}_{t} - \beta} & \le \vertii{\hat{b}_{\tau(\delta/2) + 1} - \beta} + \vertii{\xi_{t}} + \sum_{s=t}^{\tau(\delta/2)} \vertii{\E(\hat{b}_{s} \colonBreak \hat{b}_{s + 1}) - \hat{b}_{s+1}}, \label{eq:betaBDecomp}\\
		\wq \xi_{t} & \equiv  \sum_{s=t}^{\tau(\delta/2)} \hat{b}_{s} - \E(\hat{b}_{s} \colonBreak \hat{b}_{s + 1}). \no
	\end{align}
	I cap the sums at time $\tau(\delta/2)$ to give our look-back shadow prices a sufficiently large sample. Indeed, a sample with $n - \tau(\delta/2)$ observations will comprise enough data to ensure that the look-back shadow prices---and hence the $\hat{b}_{t}$ values---are well-behaved.  More specifically, I will show that $n - \tau(\delta/2) = \Theta(n)$ by showing that there exists $\gamma < 1$ that satisfies
	\begin{align}
		\tau(\delta/2) + 1\le \gamma n.\label{eq:tauBound}
	\end{align}
	To see this, note that period-$t$'s resource vector satisfies
	\begin{align*}
		(n \beta - (n - t)\alpha) / t \le \underbrace{\big(n \beta - \sum_{s = t}^{n}x_{s}a_{s}\big)/ t}_{= b_{t}} & \le n \beta / t,
	\end{align*}
	where the lower bound is within $\delta/2$ of $\beta$ unless $t \le \frac{n}{1 + \delta/(2\vertii{\alpha - \beta})}$, and the upper bound is within $\delta/2$ of $\beta$ unless $t \le \frac{n}{1 + \delta /(2\vertii{\beta})}$. Hence, if $\vertii{\alpha - \beta} \ge \vertii{\beta}$, which we can suppose without loss of generality, then $b_{t} \notin B_{\delta/2}(\beta)$ implies $t \le \frac{n}{1 + \delta/(2\vertii{\alpha - \beta})}$.
	
	I will now use \eqref{eq:betaBDecomp} to inductively prove that there exists $C > 0$ such that
	\begin{align}
		\Pr\big(\max_{s = t}^{n} \vertii{\hat{b}_{s} - \beta} > \delta\big) \le (\tau(\delta/2) + 1 - t) \big(2\exp(-Ct) + 2n\exp(-C (1 - \gamma) \sqrt{n})\big), \label{eq:inductiveHypothsis}
	\end{align}
	for all sufficiently large $t \le n$. Initializing our induction will be simple: by definition, we have $\Pr(\hat{b}_{t} \in B_{\delta}(\beta)) = 1$ for $t \ge \tau(\delta/2) + 1$, which establishes the base case. However, establishing the inductive step will require unraveling the knotty relationship between look-back shadow prices and inventory vectors. Specifically, showing that $\vertii{\hat{b}_{t} - \beta}$ is small for $t \le \tau(\delta/2)$ will require showing that $\vertii{\E(\hat{b}_{s} \colonBreak \hat{b}_{s+1}) - \hat{b}_{s+1}}$ is small for all $s \in \{t, \cdots, \tau(\delta/2)\}$, which in turn will require showing that $\vertii{\underleftarrow{y}_{s}^{\hat{b}_{s}} - y_{\infty}^{\hat{b}_{s}}}$ is small for all $s \in \{t+1, \cdots, \tau(\delta/2)+1\}$, which in turn will require showing that $\vertii{\hat{b}_{s} - \beta}$ is small for all $s \in \{t+1, \cdots, \tau(\delta/2)+1\}$. 
	
	I will now that if \eqref{eq:inductiveHypothsis} holds for sufficiently large $t + 1 \le \tau(\delta/2) + 1$, then there a suitably high probability that 
	\begin{align}
		\vertii{\hat{b}_{\tau(\delta/2) + 1} - \beta} & \le \delta/2,\n
		\vertii{\xi_{t}} & \le \delta/4, \label{eq:thisIsADisplay}\\
		\aq \sum_{s=t}^{\tau(\delta/2)} \vertii{\E(\hat{b}_{s} \colonBreak \hat{b}_{s + 1}) - \hat{b}_{s+1}} & \le \delta/4,\no
	\end{align}
	which with line \eqref{eq:betaBDecomp} will establish induction. Note that the first inequality in display \eqref{eq:thisIsADisplay} holds by the definition of $\tau(\delta/2)$, so we will only have to concern ourselves with the latter two inequalities.
	
	I will now show that the second inequality in display \eqref{eq:thisIsADisplay} holds with high probability, conditional on $\hat{b}_{s} \in B_{\delta}(\beta)$ for all $s \in \{t+1, \cdots, \tau(\delta/2)+1\}$. Since $\{\xi_{t}\}_{t = \tau(\delta/2)}^{1}$ is a martingale that satisfies $\vertii{\xi_{t} - \xi_{t+1}} = \vertii{\hat{b}_{t} - \E(\hat{b}_{t} \colonBreak \hat{b}_{t + 1})} \le (\vertii{\beta} + \delta + \vertii{\alpha})/t$, by design, the argument underlying line \eqref{eq:lastLineOfLemma} analogously implies that there exists $C > 0$ such that $\Pr(\vertii{\xi_{t}} > \delta/4) \le \exp(-Ct)$, for all sufficiently large $t$. And since $\Pr(A|B) = \Pr(A \cap B)/\Pr(B) \le \Pr(A)/\Pr(B)$, it follows that 
	\begin{align}
		\Pr\big(&\vertii{\xi_{t}} > \delta/4 \colonBreak \max_{s = t+1}^{\tau(\delta/2) + 1} \vertii{\hat{b}_{s} - \beta} \le \delta\big) \n
		& \le \frac{\Pr(\vertii{\xi_{t}} > \delta/4) }{\Pr\big(\max_{s = t+1}^{\tau(\delta/2) + 1} \vertii{\hat{b}_{s} - \beta} \le \delta\big)}\n
		& \le 2\exp(-Ct). \label{eq:expCTxi}
	\end{align}
	Note, the last line holds because $\Pr(\max_{s = t+1}^{\tau(\delta/2) + 1} \vertii{\hat{b}_{s} - \beta} \le \delta) \ge 1/2$, by our inductive hypothesis. 
	
	I will now show that the third inequality in display \eqref{eq:thisIsADisplay} holds with high probability, conditional on $\hat{b}_{s} \in B_{\delta}(\beta)$ for all $s \in \{t+1, \cdots, \tau(\delta/2)+1\}$. This step will take more work. First note that $\hat{b}_{s+1} \in B_{\delta}(\beta)$ implies $\hat{b}_{s+1} = b_{s+1}$ and $\hat{b}_{s} = b_{s}$, and thus implies
	\begin{align*}
		\E(\hat{b}_{s}& \colonBreak \hat{b}_{s+1}) - \hat{b}_{s+1} \n
		& = \E(b_{s} \colonBreak b_{s+1}) - b_{s+1}\n
		& = ((s+1) b_{s+1} - \E(\indicator{\Delta_{s+1}(\underleftarrow{y}_{s+1}^{b_{s+1}}) > 0}a_{s+1}\colonBreak b_{s+1})) / s - b_{s+1}\n
		& = ((s+1) b_{s+1} - b_{s+1} + \dot{\Lambda}_{\infty}^{b_{s+1}}(\underleftarrow{y}_{s+1}^{b_{s+1}})) / s - b_{s+1}\n
		& = \dot{\Lambda}_{\infty}^{b_{s+1}}(\underleftarrow{y}_{s+1}^{b_{s+1}}) / s\n
		& = \ddot{\Lambda}_{\infty}(y_{\infty}^{b_{s+1}})(\underleftarrow{y}_{s+1}^{b_{s+1}} - y_{\infty}^{b_{s+1}})/s + o(\vertii{\underleftarrow{y}_{s+1}^{b_{s+1}} - y_{\infty}^{b_{s+1}}})/s \n
		& = \ddot{\Lambda}_{\infty}(y_{\infty}^{\hat{b}_{s+1}})(\underleftarrow{y}_{s+1}^{\hat{b}_{s+1}} - y_{\infty}^{\hat{b}_{s+1}})/s + o(\vertii{\underleftarrow{y}_{s+1}^{\hat{b}_{s+1}} - y_{\infty}^{\hat{b}_{s+1}}})/s, 
	\end{align*}
	where the penultimate line holds by Lemma \ref{l:defineDotNew}, since $\dot{\Lambda}_{\infty}^{b_{s+1}}(y_{\infty}^{b_{s+1}}) = 0$ when $b_{s+1} \in B_{\delta}(\beta)$ and $\delta$ is small. Thus, we can choose $n$ sufficiently small so that $\max_{s = t + 1}^{\tau(\delta/2) + 1} \vertii{\hat{b}_{s} - \beta} \le \delta$ and $\max_{s = t+1}^{\tau(\delta/2) + 1}\vertii{\underleftarrow{y}_{s}^{\hat{b}_{s}} - y_{\infty}^{\hat{b}_{s}}} \le n^{-1/4}$ imply
	\begin{align}
		\sum_{s=t}^{\tau(\delta/2)}&\vertii{\E(\hat{b}_{s} \colonBreak \hat{b}_{s+1}) - \hat{b}_{s+1}} \n
		& \le \sum_{s=t}^{\tau(\delta/2)} \vertii{\ddot{\Lambda}_{\infty}(y_{\infty}^{\hat{b}_{s+1}})(\underleftarrow{y}_{s+1}^{\hat{b}_{s+1}} - y_{\infty}^{\hat{b}_{s+1}})}/s + o(\vertii{\underleftarrow{y}_{s+1}^{\hat{b}_{s+1}} - y_{\infty}^{\hat{b}_{s+1}}})/s \n
		& \le \sum_{s=t}^{\tau(\delta/2)} 2 \sigma_{1}^{\beta} \vertii{\underleftarrow{y}_{s+1}^{\hat{b}_{s+1}} - y_{\infty}^{\hat{b}_{s+1}}}/s + o(\vertii{\underleftarrow{y}_{s+1}^{\hat{b}_{s+1}} - y_{\infty}^{\hat{b}_{s+1}}})/s \n
		& \le \sum_{s=t}^{\tau(\delta/2)} 3 \sigma_{1}^{\beta}n^{-1/4}/s \n
		& \le \delta/4. \label{eq:deltaOver4}
	\end{align}
	Note, the third line holds because the largest singular value of $\ddot{\Lambda}_{\infty}(y_{\infty}^{\hat{b}_{s+1}})$ is less than twice the largest singular value of $\ddot{\Lambda}_{\infty}(y_{\infty}^{\beta})$ when $\hat{b}_{s+1} \in B_{\delta}(\beta)$ and $\delta$ is small, and the fourth line holds because the little-o term is less than $\sigma_{1}^{\beta}n^{-1/4}$ when $\vertii{\underleftarrow{y}_{s+1}^{\hat{b}_{s+1}} - y_{\infty}^{\hat{b}_{s+1}}} \le n^{-1/4}$ and $n$ is large.
	
	Further, we can use Corollary \ref{c:outsideBallBound} upper bound the probability that $\max_{s = t+1}^{\tau(\delta/2) + 1}\vertii{\underleftarrow{y}_{s}^{\hat{b}_{s}} - y_{\infty}^{\hat{b}_{s}}} > n^{-1/4}$, conditional on $\max_{s = t + 1}^{\tau(\delta/2) + 1} \vertii{\hat{b}_{s} - \beta} \le \delta$. Specifically, combining this corollary with line \eqref{eq:tauBound} and our inductive hypothesis yields the following, for some $C > 0$ and all sufficiently large $n$:
	\begin{align}
		\Pr\big(&\max_{s = t+1}^{\tau(\delta/2) + 1}\vertii{\underleftarrow{y}_{s}^{\hat{b}_{s}} - y_{\infty}^{\hat{b}_{s}}} > n^{-1/4} \colonBreak \max_{s = t + 1}^{\tau(\delta/2) + 1} \vertii{\hat{b}_{s} - \beta} \le \delta \big) \n
		& \le \sum_{s = t+1}^{\tau(\delta/2) + 1}\Pr\big(\sup_{b \in B_{\delta}(\beta)} \vertii{\underleftarrow{y}_{s}^{b} - y_{\infty}^{b}} > n^{-1/4} \colonBreak \max_{s = t+1}^{\tau(\delta/2) + 1} \vertii{\hat{b}_{s} - \beta} \le \delta\big)\n
		& \le \sum_{s = t+1}^{\tau(\delta/2) + 1} \frac{\Pr\big(\sup_{b \in B_{\delta}(\beta)} \vertii{\underleftarrow{y}_{s}^{b} - y_{\infty}^{b}} > n^{-1/4}\big)}{\Pr\big(\max_{s = t+1}^{\tau(\delta/2)} \vertii{\hat{b}_{s} - \beta} \le \delta\big)}\n
		& \le \sum_{s = t+1}^{\tau(\delta/2) + 1} \frac{\exp(-C n^{-1/2} (n - s))}{1/2} \n
		& \le 2n\exp(-C n^{-1/2} (n - \tau(\delta/2) - 1)) \n
		& \le 2n\exp(-C (1 - \gamma) \sqrt{n} ). \label{eq:c1gsqrtn}
	\end{align}
	
	And now, finally, we can combine lines \eqref{eq:betaBDecomp}, \eqref{eq:expCTxi}, \eqref{eq:deltaOver4}, and \eqref{eq:c1gsqrtn} to establish that
	\begin{align*}
		\Pr(&\hat{b}_{t} \notin B_{\delta}(\beta) \colonBreak \max_{s = t + 1}^{n} \vertii{\hat{b}_{s} - \beta} \le \delta) \\
		& \le \Pr\big(\vertii{\xi_{t}} > \delta/4 \colonBreak \max_{s = t + 1}^{n} \vertii{\hat{b}_{s} - \beta} \le \delta\big) \\
		& \qquad + \Pr\big(\sum_{s=t}^{\tau(\delta/2)}\vertii{\E(\hat{b}_{s} \colonBreak \hat{b}_{s+1}) - \hat{b}_{s+1}} > \delta/4 \colonBreak \max_{s = t + 1}^{n} \vertii{\hat{b}_{s} - \beta} \le \delta\big) \\
		& \le 2\exp(-Ct) + \Pr\big(\max_{s = t+1}^{\tau(\delta/2) + 1}\vertii{\underleftarrow{y}_{s}^{\hat{b}_{s}} - y_{\infty}^{\hat{b}_{s}}} > n^{-1/4} \colonBreak \max_{s = t + 1}^{n} \vertii{\hat{b}_{s} - \beta} \le \delta\big)\\
		& \le 2\exp(-Ct) + 2n\exp(-C (1 - \gamma) \sqrt{n}) .
	\end{align*}
	And with our inductive hypothesis, this implies that
	\begin{align*}
		\Pr\big(&\max_{s = t}^{n} \vertii{\hat{b}_{s} - \beta} > \delta\big) \\
		& = \Pr\big(\max_{s = t+1}^{n} \vertii{\hat{b}_{s} - \beta} > \delta\big) + \Pr(\hat{b}_{t} \notin B_{\delta}(\beta) \colonBreak \max_{s = t + 1}^{n} \vertii{\hat{b}_{s} - \beta} \le \delta)\\
		& \le (\tau(\delta/2) + 1 - t-1) (2\exp(-C(t+1)) + 2n\exp(-C (1 - \gamma) \sqrt{n})) \\
		& \qquad + 2\exp(-Ct) + 2n\exp(-C (1 - \gamma) \sqrt{n})\\
		& \le (\tau(\delta/2) + 1 - t) \big(2\exp(-Ct) + 2n\exp(-C (1 - \gamma) \sqrt{n})\big).
	\end{align*}
\end{proof}	

\begin{proof}[Lemma \ref{l:MyopicRegretFirstLearn} Proof]
	This proof will closely follow the Lemma \ref{l:MyopicRegretFirst} proof. That proof's argument establishes that the first term of the myopic regret is $o(1/t)$, so I will begin with the second term of the myopic regret. First, define $\xi_{t-1}\equiv \sup_{b \in B_{\delta}(\beta)}\vertii{y_{t-1}^{b} - y_{\infty}^{b}} + C/t$, for the $C > 0$ used in the proof of Lemma \ref{l:MyopicRegretFirst}, and analogously define $\underleftarrow{\xi}_{t+1} \equiv \sup_{b \in B_{\delta/2}(\beta)}\vertii{\underleftarrow{y}_{t}^{b} - y_{\infty}^{b}}$. Note, I subscript this latter variable with $t + 1$ because $\underleftarrow{y}_{t}^{b}$ is determined by that time. Next, the argument underlying line \eqref{eq:lineToCombine} establishes the following, for large $t$:
	\begin{align*}
		\E\big(&\indicator{b_{t} \in B_{\delta/2}(\beta)}\indicator{\Delta_{t}(\underleftarrow{y}_{t}^{b_{t}}) > 0}\Delta_{t}(y_{t-1}^{\psi_{t}^{b_{t}}(a_{t})})^{-}\big)\\
		& \le \E\big(\indicator{b_{t} \in B_{\delta/2}(\beta)}\indicator{\Delta_{t}(y_{\infty}^{b_{t}}- \underleftarrow{\xi}_{t+1}\iota) > 0} \Delta_{t}(y_{\infty}^{b_{t}} + \xi_{t-1} \iota)^{-}\big)\\
		& = \E\big(\indicator{b_{t} \in B_{\delta/2}(\beta)}\E(\indicator{\Delta_{t}(y) > 0} \Delta_{t}(\bar{y})^{-})\big|_{y = y_{\infty}^{b_{t}}- \underleftarrow{\xi}_{t+1}\iota,\ \bar{y} =  y_{\infty}^{b_{t}} + \xi_{t-1} \iota}\big).
	\end{align*}
	The last line holds because the random function $\Delta_{t}$ is independent of $b_{t}$, $\underleftarrow{\xi}_{t+1}$, and $\xi_{t-1}$.
	
	Now let $\mathcal{E}$ represent the event in which $y_{\infty}^{b} + \xi_{t-1} \iota \in B_{\epsilon}(y_{\infty}^{\beta})$ and let $\underleftarrow{\mathcal{E}}$ represent the event in which $y_{\infty}^{b}- \underleftarrow{\xi}_{t}\iota \in B_{\epsilon}(y_{\infty}^{\beta})$, where $\epsilon$ is the constant defined in Lemma \ref{l:BoundEindicatorDelta2}. Further, let $\delta$ be small enough so that (i) $\sup_{b \in B_{\delta/2}(\beta)} \vertii{y_{\infty}^{b}} \le 2 \vertii{y_{\infty}^{\beta}}$ and (ii) $\{y_{\infty}^{b} \colonBreak b \in B_{\delta/2}(\beta)\} \subset B_{\epsilon/2}(y_{\infty}^{\beta})$, in which case $\mathcal{E}^{c}$ implies $\xi_{t-1} > \epsilon/(2\sqrt{m})$ and $\underleftarrow{\mathcal{E}^{c}}$ implies $\underleftarrow{\xi}_{t+1} > \epsilon/(2\sqrt{m})$. And with this, we can use Lemma \ref{l:BoundEindicatorDelta2} to continue where we left off:
	\begin{align*}
		\E\big(&\indicator{b_{t} \in B_{\delta/2}(\beta)}\indicator{\Delta_{t}(\underleftarrow{y}_{t}^{b_{t}}) > 0}\Delta_{t}(y_{t-1}^{\psi_{t}^{b_{t}}(a_{t})})^{-}\big)\\
		& \le \E\big(\indicator{b_{t} \in B_{\delta/2}(\beta)}\indicator{\mathcal{E} \cap \underleftarrow{\mathcal{E}}}\E(\indicator{\Delta_{t}(y) > 0} \Delta_{t}(\bar{y})^{-})\big|_{y = y_{\infty}^{b_{t}}- \underleftarrow{\xi}_{t+1}\iota,\ \bar{y} =  y_{\infty}^{b_{t}} + \xi_{t-1} \iota}\big)\\
		& \qquad + \E\big(\indicator{b_{t} \in B_{\delta/2}(\beta)}\indicator{\mathcal{E}^{c} \cup \underleftarrow{\mathcal{E}^{c}}}\E(\indicator{\Delta_{t}(y) > 0} \Delta_{t}(\bar{y})^{-})\big|_{y = y_{\infty}^{b_{t}}- \underleftarrow{\xi}_{t+1}\iota,\ \bar{y} =  y_{\infty}^{b_{t}} + \xi_{t-1} \iota}\big)\\
		& \le \E\big(\indicator{b_{t} \in B_{\delta/2}(\beta)}\indicator{\mathcal{E} \cap \underleftarrow{\mathcal{E}}}2 \sigma_{1}^{\beta}\vertii{\xi_{t-1}\iota + \underleftarrow{\xi}_{t+1}\iota}^{2}\big) \\
		& \qquad + \E\big(\indicator{b_{t} \in B_{\delta/2}(\beta)}\indicator{\mathcal{E}^{c} \cup \underleftarrow{\mathcal{E}^{c}}}\vertii{\alpha}\vertii{y_{\infty}^{b_{t}} + \xi_{t-1}\iota}\big)\\
		& \le 4 m \sigma_{1}^{\beta}\E(\xi_{t-1}^{2} + \underleftarrow{\xi}_{t+1}^{2})\\
		& \qquad + 2\vertii{\alpha}\vertii{y_{\infty}^{\beta}}\Pr\big(\xi_{t-1} > \epsilon/(2\sqrt{m}) \cup \underleftarrow{\xi}_{t+1} > \epsilon/(2\sqrt{m})\big) \\
		& \qquad + 2\sqrt{m}\vertii{\alpha}\vertii{y_{\infty}^{\beta}}\E\big(\indicator{\xi_{t-1} > \epsilon/(2\sqrt{m}) \cup \underleftarrow{\xi}_{t+1} > \epsilon/(2\sqrt{m})}\xi_{t-1}\big),
	\end{align*}
	where the last line holds because $(x + y)^{2} \le 2x^{2} + 2 y^{2}$ and $\vertii{\iota} = \sqrt{m}$. Finally, Proposition \ref{p:withinBallBound}, Proposition \ref{p:outsideBallBound}, and Corollary \ref{c:outsideBallBound} imply that this last expression is less than $C/t + C/(n-t)$, for some $C > 0$.
	
	Finally, the same argument yields the same bound for the third term of myopic regret. 
\end{proof}

\begin{proof}[Corollary \ref{c:concentrateMeasureLearning} Proof]
	Copy the proof of Corollary \ref{c:concentrateMeasure}.
\end{proof}	

\begin{proof}[Lemma \ref{l:probBInControl} Proof]
	I will begin with a high-level plan of attack. The main idea of the proof is that allocating $t (\beta + \xi)$ units of inventory to the first $t$ periods leaves us with only $(n-t)(\beta - \frac{t}{n-t} \xi)$ units for the last $n - t$ periods, so $b_{t} = \beta +\xi$ must imply
	\begin{align}
		v_{n}^{\beta} & \le \bar{V}_{t}^{\beta +\xi} + \underleftarrow{\bar{V}}_{t}^{\beta - \frac{t}{n - t}\xi}\n
		& = t \Lambda_{t}^{\beta +\xi}(y_{t}^{\beta +\xi}) + (n-t) \underleftarrow{\Lambda}_{t}^{\beta - \frac{t}{n - t}\xi}(\underleftarrow{y}_{t}^{\beta - \frac{t}{n - t}\xi})\n
		& \equiv \bar{\Lambda}_{t}^{\xi} \label{eq:xitDef}
	\end{align}
	where $\underleftarrow{\bar{V}}_{t}^{b}$, $\underleftarrow{\Lambda}_{t}^{b}$, and $\underleftarrow{y}_{t}^{b}$ and are equivalent to $\bar{V}_{n - t}^{b}$, $\Lambda_{n-t}^{b}$, and $y_{n-t}^{b}$, but with the order of the customers reversed. Accordingly, it follows that 
	\begin{align}
		\underline{R}_{n}& \ge \bar{V}_{n}^{\beta} - \bar{\Lambda}_{t}^{b_{t} - \beta} = n\Lambda_{n}^{\beta}(y_{n}^{\beta}) - \bar{\Lambda}_{t}^{b_{t} - \beta}. \label{eq:regretInTermsOfBarLambda}
	\end{align}
	And the expression on the right should be large when $b_{t}$ meaningfully deviates from $\beta$ since, in the limit, we have
	\begin{align}
		n\Lambda_{\infty}^{\beta}(y_{\infty}^{\beta}) - \bar{\Lambda}_{\infty}^{\xi} & \ge C t \min(\vertii{\xi}^{2}, 1),\label{eq:costOfSplittingB}
	\end{align}
	for some $C > 0$, where\endnote{
		Technically, $\bar{\Lambda}_{\infty}^{\xi}$ should  be expressed $\bar{\Lambda}_{nt\infty}^{\xi}$, but I suppress the first two subscripts for simplicity. Likewise, $\bar{\Lambda}_{t}^{\xi}$ should technically be expressed $\bar{\Lambda}_{nt}^{\xi}$.
	}
	\begin{align}
		\bar{\Lambda}_{\infty}^{\xi} & \equiv
		t\Lambda_{\infty}^{\beta + \xi}(y_{\infty}^{\beta + \xi}) + (n-t) \Lambda_{\infty}^{\beta - \frac{t}{n - t}\xi}(y_{\infty}^{\beta - \frac{t}{n - t}\xi}). \label{eq:infiniteHorizonAnalog}
	\end{align}
	Line \eqref{eq:costOfSplittingB} suggests that reserving $t(\beta + \xi)$ units of inventory for the first $t$ periods should sacrifice $\Omega(t)$ units of value when $\xi$ non-negligible. I will leverage this fact to show that $\underline{R}_{n}$ is almost always large when $\vertii{b_{t} - \beta}$ is non-negligible. And since $\E(\underline{R}_{n})$ is relatively small, by Theorem \ref{p:prop1}, it follows that $\vertii{b_{t} - \beta}$ is usually negligible. 
	
	Before delving into the details, I will provide a more thorough proof sketch. The proof will have five steps. The first derives limiting bound \eqref{eq:costOfSplittingB}, our only tool for establishing the cost of $b_{t}$ diverging from $\beta$. Now with \eqref{eq:costOfSplittingB}, it's relatively easy to lower bound the regret when $b_{t} = \beta + \xi$, for some specific $\xi$ that lies outside of a ball of the origin. But that's not enough, as we must lower bound this regret when $b_{t} = \beta + \xi$, for \emph{any} $\xi$ that lies outside of a ball of the origin. To create such a uniform result, the second part of the proof bounds $\bar{\Lambda}_{t}^{\xi}$ for all $\xi \in \mathbbm{R}^{m}$ in terms of $\bar{\Lambda}_{t}^{\zeta}$, $y_{t}^{\beta + \zeta}$, and $\underleftarrow{y}_{t}^{\beta - \frac{t}{n - t}\zeta}$, for some given given $\zeta \in \mathbbm{R}^{m}$, and the third part uses this bound to show that $\sup_{\xi \notin B_{2\sqrt{m}\epsilon}(0)}\bar{\Lambda}_{t}^{\xi}$ is usually smaller than $\max_{k \in \{-1, 1\}} \max_{j \in [m]}\bar{\Lambda}_{t}^{k\epsilon \omega_{j}^{\beta}}$, where $\epsilon$ is a small positive number and $\{\omega_{i}^{\beta}\}_{i \in [m]}$ are the orthonormal eigenvectors of $\ddot{\Lambda}_{\infty}(y_{\infty}^{\beta})^{-1}$. Hence, the second and third steps of the proof collapse the relevant domain of $b_{t} - \beta$ from the infinite set $\mathbbm{R}^{m} \setminus B_{2\sqrt{m}\epsilon}(0)$ to the finite set $\{k\epsilon \omega_{j}^{\beta} \colonBreak k \in \{-1, 1\} \times j \in [m]\}$. The fourth step of the proof uses our shadow price convergence results to show that $n\Lambda_{n}^{\beta}(y_{n}^{\beta}) - \max_{k \in \{-1, 1\}} \max_{j \in [m]} \bar{\Lambda}_{t}^{k\epsilon \omega_{j}^{\beta}}$ is usually very large, and the last step combines this with the previous results to establish that $\xi = b_{t} - \beta$ must rarely fall outside of $B_{2\sqrt{m}\epsilon}(0)$.
	
	To begin the proof, note that Lemmas \ref{l:defineDotNew} and \ref{l:lipschitzYinB} imply that $\Lambda_{\infty}^{b}(y_{\infty}^{b})$ is concave in $b$, since $\tfrac{\partial^{2}}{\partial b^{2}} \Lambda_{\infty}^{b}(y_{\infty}^{b}) = \tfrac{\partial}{\partial b} y_{\infty}^{b} = - \ddot{\Lambda}_{\infty}(y_{\infty}^{b})^{-1}$ is negative definite. This concavity implies that we can restrict attention to small $\xi$ vectors, since $\bar{\Lambda}_{\infty}^{\xi}$ decreases in the magnitude of $\xi$. But, more importantly, the concavity implies line \eqref{eq:costOfSplittingB}, as I will now show.
	
	Let $i \in [m]$ denote the index of the largest element of $\xi$, so that either $e_{i}'\xi = \vertii{\xi}_{\infty}$ or $-e_{i}'\xi = \vertii{\xi}_{\infty}$. Since the minus sign doesn't meaningfully affect the analysis, I will henceforth suppose $e_{i}'\xi = \vertii{\xi}_{\infty} \equiv \gamma $, in which case 
	\begin{align}
		\bar{\Lambda}_{\infty}^{\xi} \le \sup_{\{\zeta \in \mathbbm{R}^{m} \colonBreak e_{i}'\zeta = \gamma\}} \bar{\Lambda}_{\infty}^{\zeta} .\label{eq:VVVxi}
	\end{align}
	The solution to this optimization problem satisfies the following first-order conditions for some Lagrange multiplier $\lambda$:
	\begin{align*}
		0 = & \tfrac{\partial}{\partial \zeta} \Big(t \Lambda_{\infty}^{\beta +\zeta}(y_{\infty}^{\beta +\zeta}) + (n-t) \Lambda_{\infty}^{\beta - \frac{t}{n - t}\zeta}(y_{\infty}^{\beta - \frac{t}{n - t}\zeta}) - \lambda (e_{i}' \zeta - \gamma)\Big) \n
		= & t (y_{\infty}^{\beta + \zeta} - y_{\infty}^{\beta - \frac{t}{n - t}\zeta}) - \lambda e_{i}.
	\end{align*} 
	Now we'll use Lemma \ref{l:lipschitzYinB} to differentiate this with respect to $\gamma$:
	\begin{align*}
		0 & = \tfrac{\partial}{\partial \gamma} 0 \big|_{\gamma = 0}\\
		& = \tfrac{\partial}{\partial \gamma}\Big( t (y_{\infty}^{\beta + \zeta} - y_{\infty}^{\beta - \frac{t}{n - t}\zeta}) - \lambda e_{i}\Big)\big|_{\gamma = 0}\\
		& = - \Big(t\ddot{\Lambda}_{\infty}(y_{\infty}^{\beta + \zeta})^{-1} + \frac{t^{2}}{n-t} \ddot{\Lambda}_{\infty}(y_{\infty}^{\beta - \frac{t}{n - t}\zeta})^{-1}\Big) \tfrac{\partial}{\partial \gamma} \zeta - e_{i} \tfrac{\partial}{\partial \gamma} \lambda\big|_{\gamma = 0}\\
		& = - \frac{nt}{n - t}\ddot{\Lambda}_{\infty}(y_{\infty}^{\beta})^{-1}\tfrac{\partial}{\partial \gamma} \zeta - e_{i}\tfrac{\partial}{\partial \gamma} \lambda\big|_{\gamma = 0}.
	\end{align*}
	The last line holds because $\gamma = 0$ implies $\zeta = 0$. Combining the $e_{i}'\zeta = \gamma$ constraint with the expression above yields
	\begin{align*}
		- \frac{n - t}{nt}e_{i}'\ddot{\Lambda}_{\infty}(y_{\infty}^{\beta}) e_{i} \tfrac{\partial}{\partial \gamma} \lambda\big|_{\gamma = 0} = 
		\tfrac{\partial}{\partial \gamma} e_{i}'\zeta \big|_{\gamma = 0} = \tfrac{\partial}{\partial \gamma} \gamma\big|_{\gamma = 0} = 1,
	\end{align*}
	which implies that
	\begin{align*}
		\tfrac{\partial}{\partial \gamma} \lambda\big|_{\gamma = 0} & = \frac{-nt}{(n-t) e_{i}'\ddot{\Lambda}_{\infty}(y_{\infty}^{\beta}) e_{i}}.
	\end{align*}
	Further, since $\lambda = 0$ when $\gamma = 0$, by the concavity of $\Lambda_{\infty}^{b}(y_{\infty}^{b})$ in $b$, it follows that for sufficiently small $\gamma$ we have
	\begin{align}
		\lambda \le \frac{-nt\gamma}{2(n - t)e_{i}' \ddot{\Lambda}_{\infty}(y_{\infty}^{\beta})e_{i}} . \no
	\end{align}
	By definition, our Lagrange multiplier also satisfies $\tfrac{\partial}{\partial \gamma} \sup_{\{\zeta \in \mathbbm{R}^{m} \colonBreak e_{i}'\zeta = \gamma\}} \bar{\Lambda}_{\infty}^{\zeta} = \lambda$, which with the result above yields the following, for sufficiently small $\gamma$:
	\begin{align*}
		n\Lambda_{\infty}^{\beta}(y_{\infty}^{\beta}) &- \sup_{\{\zeta \in \mathbbm{R}^{m} \colonBreak e_{i}'\zeta = \gamma\}} \bar{\Lambda}_{\infty}^{\zeta} \\
		& = - \big(\sup_{\{\zeta \in \mathbbm{R}^{m} \colonBreak e_{i}'\zeta = \gamma\}} \bar{\Lambda}_{\infty}^{\zeta} - \sup_{\{\zeta \in \mathbbm{R}^{m} \colonBreak e_{i}'\zeta = 0\}} \bar{\Lambda}_{\infty}^{\zeta}\big)\n
		& = - \int_{g = 0}^{\gamma} \tfrac{\partial}{\partial g} \sup_{\{\zeta \in \mathbbm{R}^{m} \colonBreak e_{i}'\zeta = g\}} \bar{\Lambda}_{\infty}^{\zeta} dg \n
		& \ge - \int_{g = 0}^{\gamma} \frac{-ntg}{2(n - t)e_{i}' \ddot{\Lambda}_{\infty}(y_{\infty}^{\beta})e_{i}} dg\n
		& = \frac{nt\gamma^{2}}{4(n - t)e_{i}' \ddot{\Lambda}_{\infty}(y_{\infty}^{\beta})e_{i}}\\
		& \ge \frac{t(\vertii{\xi}/\sqrt{m})^{2}}{4\max_{j \in [m]} e_{j}' \ddot{\Lambda}_{\infty}(y_{\infty}^{\beta})e_{j}}.
	\end{align*}
	Note, the first line above holds because the concavity of $\Lambda_{\infty}^{b}(y_{\infty}^{b})$ in $b$ implies that $n\Lambda_{\infty}^{\beta}(y_{\infty}^{\beta}) = \sup_{\{\zeta \in \mathbbm{R}^{m} \colonBreak e_{i}'\zeta = 0\}} \bar{\Lambda}_{\infty}^{\zeta}$, and the last line holds because $\gamma = \vertii{\xi}_{\infty} \ge \vertii{\xi}/\sqrt{m}$. Finally, combining the result above with line \eqref{eq:VVVxi} yields line \eqref{eq:costOfSplittingB}.
	
	Second, I will now bound the difference between $\bar{\Lambda}_{t}^{\xi}$ and $\bar{\Lambda}_{t}^{\zeta}$ in terms of $y_{t}^{\beta + \zeta}$ and $\underleftarrow{y}_{t}^{\beta - \frac{t}{n - t}\zeta}$, which will enable us to invoke our shadow price convergence results. To this end, first note that 
	\begin{align*}
		\bar{\Lambda}_{t}^{\xi} & = \max_{x \in [0, 1]^n} \sum_{s=1}^{n} x_{s}u_{s}  \quad  \\
		\stt & \quad \sum_{s=1}^{t} x_{s} a_{s} \le t(\beta +\xi),\\
		& \quad \sum_{s=t+1}^{n} x_{s} a_{s} \le (n-t)\big(\beta - \frac{t}{n - t}\xi\big).
	\end{align*}
	Since this linear program is concave in its constraints, $\bar{\Lambda}_{t}^{\xi}$ must be concave in $\xi$. Accordingly, the $\bar{\Lambda}_{t}^{\xi}$ function lies below the hyperplane characterized by supergradient $\tfrac{\partial}{\partial \zeta} \bar{\Lambda}_{t}^{\zeta} \equiv t y_{t}^{\beta +\zeta} \tfrac{\partial}{\partial \zeta} (\beta +\zeta) + (n-t) \underleftarrow{y}_{t}^{\beta - \frac{t}{n - t}\zeta} \tfrac{\partial}{\partial \zeta} \big(\beta - \frac{t}{n - t}\zeta\big) = t (y_{t}^{\beta + \zeta} - \underleftarrow{y}_{t}^{\beta - \frac{t}{n - t}\zeta})$: 
	\begin{align}
		\bar{\Lambda}_{t}^{\xi} - \bar{\Lambda}_{t}^{\zeta} & \le  t (\xi- \zeta)' \big(y_{t}^{\beta + \zeta} - \underleftarrow{y}_{t}^{\beta - \frac{t}{n - t}\zeta}\big). \label{eq:XiXixixi}
	\end{align}

	Third, I will use the preceding inequality to establish that
	\begin{align}
		\Pr\big(\sup_{\xi \notin B_{2\sqrt{m}\epsilon}(0)}\bar{\Lambda}_{t}^{\xi} \le \max_{k \in \{-1, 1\}} \max_{j \in [m]}\bar{\Lambda}_{t}^{k\epsilon \omega_{j}^{\beta}}\big) \ge 3/4, \label{eq:PrLambdaLambdaHalf}
	\end{align}
	for all sufficiently small $\epsilon > 0$ and large $t$. This bound is crucial, as it enables us to replace the infinite continuum of $\bar{\Lambda}_{t}^{\xi}$ values for all $\xi \notin B_{2\sqrt{m}\epsilon}(0)$, with the largest of the $2 m$ values of $\bar{\Lambda}_{t}^{k\epsilon \omega_{j}^{\beta}}$. To begin, note that Lemma \ref{l:lipschitzYinB} yields the following, for $k \in \{-1, 1\}$ and small $\epsilon > 0$:
	\begin{align*}
		y_{\infty}^{\beta + k\epsilon \omega_{i}} - y_{\infty}^{\beta} =  - k\epsilon\ddot{\Lambda}_{\infty}(y_{\infty}^{\beta})^{-1} \omega_{i}^{\beta} + o(\epsilon) = - k\epsilon \omega_{i}^{\beta}/\sigma_{i}^{\beta} + o(\epsilon).
	\end{align*}
	Combining this with \eqref{eq:XiXixixi} yields the following, for $t \le n/2$:
	\begin{align}
		\bar{\Lambda}_{t}^{\xi} - \bar{\Lambda}_{t}^{k\epsilon \omega_{i}^{\beta}} & \le  t (\xi - k\epsilon\omega_{i}^{\beta})' \big(y_{t}^{\beta + k\epsilon \omega_{i}^{\beta}} - \underleftarrow{y}_{t}^{\beta - \frac{t}{n - t}k\epsilon \omega_{i}^{\beta}}\big)\n
		& = t (\xi - k\epsilon\omega_{i}^{\beta})' \big(y_{\infty}^{\beta + k\epsilon \omega_{i}^{\beta}} - y_{\infty}^{\beta} - y_{\infty}^{\beta - \frac{t}{n - t}k\epsilon \omega_{i}^{\beta}} + y_{\infty}^{\beta}\big) \n
		& \qquad + t (\xi - 
		k\epsilon\omega_{i}^{\beta})'(y_{t}^{\beta + k\epsilon \omega_{i}^{\beta}} - y_{\infty}^{\beta + k\epsilon \omega_{i}^{\beta}}) - t (\xi - k\epsilon\omega_{i}^{\beta})'(\underleftarrow{y}_{t}^{\beta - \frac{t}{n - t}k\epsilon\omega_{i}} - y_{\infty}^{\beta - \frac{t}{n - t}k\epsilon\omega_{i}}) \n
		& = \frac{-ntk\epsilon}{n-t}(\xi - 
		k\epsilon\omega_{i}^{\beta})'\omega_{i}^{\beta}/\sigma_{i}^{\beta} + t \vertii{\xi - k\epsilon\omega_{i}^{\beta}} (o(\epsilon) + O_{p}(t^{-1/2})), \label{eq:manyLittleOmega}
	\end{align} 
	where the last line holds because $\vertii{y_{t}^{\beta + \epsilon \omega_{i}^{\beta}} - y_{\infty}^{\beta + \epsilon \omega_{i}^{\beta}}}$ and $\vertii{\underleftarrow{y}_{t}^{\beta - \frac{t}{n - t}\epsilon\omega_{i}} - y_{\infty}^{\beta - \frac{t}{n - t}\epsilon\omega_{i}}}$ are $O_{p}(t^{-1/2})$ when $t \le n/2$, by Proposition \ref{p:withinBallBound}. 
	
	Now, to derive \eqref{eq:PrLambdaLambdaHalf} from \eqref{eq:manyLittleOmega}, let $\gamma_{i} \equiv \xi'\omega_{i}^{\beta}$, so that $\xi = \sum_{i = 1}^{m} \gamma_{i} \omega_{i}^{\beta}$, and let $j = \argmax_{i \in [m]} \verti{\gamma_{i}}$ and $k = \sign(\gamma_{j})$, so that $k\gamma_{j} \ge \vertii{\xi}/\sqrt{m}$. Further, choose $\xi \notin B_{2\sqrt{m}\epsilon}$, in which case $\epsilon \le \vertii{\xi}/(2\sqrt{m})$, and hence 
	\begin{align*}
		k(\xi - k\epsilon\omega_{j}^{\beta})'\omega_{j}^{\beta} & = k\gamma_{j} - \epsilon \ge \vertii{\xi}/(2\sqrt{m})\\
		\aq \vertii{\xi - k\epsilon\omega_{i}^{\beta}} &\le 2 \vertii{\xi}.
	\end{align*}
	Finally, set $\epsilon$ small enough so that the $o(\epsilon)$ term in \eqref{eq:manyLittleOmega} is less than $\max_{i \in [m]} \epsilon/(16 \sqrt{m} \sigma_{i}^{\beta})$, and choose $t$ large enough so that the $O_{p}(t^{-1/2})$ term is less than $\max_{i \in [m]} \epsilon/(16 \sqrt{m} \sigma_{i}^{\beta})$, with at least three-quarters probability. When this last event happens, the previous two inequalities and line \eqref{eq:manyLittleOmega} yield the following, for all $\xi \notin B_{2\sqrt{m}\epsilon}(0)$:
	\begin{align*}
		\bar{\Lambda}_{t}^{\xi} - \bar{\Lambda}_{t}^{k\epsilon \omega_{j}^{\beta}} & \le  \frac{-nt\epsilon}{n-t}\vertii{\xi}/(2\sqrt{m}\sigma_{j}^{\beta}) + 2t \vertii{\xi} (o(\epsilon) + O_{p}(t^{-1/2}))\\
		& \le -t\epsilon\vertii{\xi}/(2\sqrt{m}\sigma_{j}^{\beta}) + 2t \vertii{\xi} (\epsilon/(16 \sqrt{m} \sigma_{i}^{\beta}) + \epsilon/(16 \sqrt{m} \sigma_{j}^{\beta})) \\
		& \le -t\epsilon\vertii{\xi}/(4\sqrt{m}\sigma_{j}^{\beta})\\
		& \le 0.
	\end{align*}
	This establishes line \eqref{eq:PrLambdaLambdaHalf}.
	
	Fourth, I will use \eqref{eq:costOfSplittingB} to show that 
	\begin{align}
		\Pr\big(n\Lambda_{n}^{\beta}(y_{n}^{\beta}) - \max_{k \in \{-1, 1\}} \max_{j \in [m]} \bar{\Lambda}_{t}^{k\epsilon \omega_{j}^{\beta}} \ge n^{2/3}\big) \ge 3/4. \label{eq:threeQuarterBound}
	\end{align}
	This expression implies that there is probably at least one combination of $k \in \{-1, 1\}$ and $j \in [m]$ for which allocating $t (\beta + k \epsilon \omega_{j}^{\beta})$ units of inventory to the first $t$ periods is very costly. And with \eqref{eq:PrLambdaLambdaHalf}, this will imply that there's a decent chance that allocating $t (\beta + \xi)$ units of inventory to the first $t$ periods will be very costly, for \emph{any} $\xi \notin B_{2\sqrt{m}\epsilon}(0)$. 
	
	To begin, a nasty series of triangle inequalities yields the following, for $\zeta \equiv k\epsilon \omega_{j}^{\beta}$:
	\begin{align*}
		n\Lambda_{n}^{\beta}(y_{n}^{\beta}) - \bar{\Lambda}_{t}^{\zeta} & \ge
		n\Lambda_{\infty}^{\beta}(y_{\infty}^{\beta}) - t \Lambda_{\infty}^{\beta +\zeta}(y_{\infty}^{\beta +\zeta}) + (n-t) \Lambda_{\infty}^{\beta - \frac{t}{n - t}\zeta}(y_{\infty}^{\beta - \frac{t}{n - t}\zeta}) \\
		& \quad -n\verti{\Lambda_{\infty}^{\beta}(y_{n}^{\beta}) - \Lambda_{\infty}^{\beta}(y_{\infty}^{\beta})} \\
		& \quad -t\verti{\Lambda_{\infty}^{\beta+\zeta}(y_{t}^{\beta + \zeta}) - \Lambda_{\infty}^{\beta+\zeta}(y_{\infty}^{\beta + \zeta})} \\
		& \quad -(n-t)\verti{\Lambda_{\infty}^{\beta - \frac{t}{n - t}\zeta}(\underleftarrow{y}_{t}^{\beta - \frac{t}{n - t}\zeta}) - \Lambda_{\infty}^{\beta - \frac{t}{n - t}\zeta}(y_{\infty}^{\beta - \frac{t}{n - t}\zeta})} \\
		& \quad - t\vertib{\Lambda_{t}^{\beta}(y_{n}^{\beta}) - \Lambda_{\infty}^{\beta}(y_{n}^{\beta}) - \Lambda_{t}^{\beta +\zeta}(y_{t}^{\beta +\zeta}) + \Lambda_{\infty}^{\beta +\zeta}(y_{t}^{\beta +\zeta})} \\
		& \quad - (n-t) \vertib{\underleftarrow{\Lambda}_{t}^{\beta}(y_{n}^{\beta}) - \Lambda_{\infty}^{\beta}(y_{n}^{\beta}) - \underleftarrow{\Lambda}_{t}^{\beta - \frac{t}{n - t}\zeta}(\underleftarrow{y}_{t}^{\beta - \frac{t}{n - t}\zeta}) + \Lambda_{\infty}^{\beta - \frac{t}{n - t}\zeta}(\underleftarrow{y}_{t}^{\beta - \frac{t}{n - t}\zeta}) }.
	\end{align*}
	Each term on the right is bounded in probability: First, line \eqref{eq:costOfSplittingB} establishes that $n\Lambda_{\infty}^{\beta}(y_{\infty}^{\beta}) - t \Lambda_{\infty}^{\beta +\zeta}(y_{\infty}^{\beta +\zeta}) + (n-t) \Lambda_{\infty}^{\beta - \frac{t}{n - t}\zeta}(y_{\infty}^{\beta - \frac{t}{n - t}\zeta}) = \Omega(t)$. Second, since $\Lambda_{\infty}^{\beta}$ is differentiable and since $\vertii{y_{n}^{\beta} - y_{\infty}^{\beta}} = O_{p}(n^{-1/2})$, by Proposition \ref{p:withinBallBound}, we have $n\verti{\Lambda_{\infty}^{\beta}(y_{n}^{\beta}) - \Lambda_{\infty}^{\beta}(y_{\infty}^{\beta})}= O_{p}(n^{1/2})$. Likewise, $t\verti{\Lambda_{\infty}^{\beta+\zeta}(y_{t}^{\beta + \zeta}) - \Lambda_{\infty}^{\beta+\zeta}(y_{\infty}^{\beta + \zeta})}$ and $(n-t)\verti{\Lambda_{\infty}^{\beta - \frac{t}{n - t}\zeta}(\underleftarrow{y}_{t}^{\beta - \frac{t}{n - t}\zeta}) - \Lambda_{\infty}^{\beta - \frac{t}{n - t}\zeta}(y_{\infty}^{\beta - \frac{t}{n - t}\zeta})}$ are $O_{p}(t^{1/2})$ and $O_{p}((n-t)^{1/2})$, respectfully. Finally, Proposition \ref{p:withinBallBound} and Lemma \ref{l:GaussianProcessLambda} imply that the last two terms are also $O_p(t^{1/2})$ and $O_p((n - t)^{1/2})$. Accordingly, we can set $n$ large enough so that if $n^{3/4} \le t \le n/2$ then there is at least a 75\% chance that (i) the $\Omega(t)$ term exceeds $2n^{2/3}$ and (ii) the sum of the $O_{p}(n^{1/2})$, $O_{p}(t^{1/2})$, and $O_{p}((n-t)^{1/2})$ terms are no more than $n^{2/3}$, for all combinations of $k \in \{-1, 1\}$ and $j \in [m]$. And this establishes \eqref{eq:threeQuarterBound}.
	
	Finally, combining \eqref{eq:regretInTermsOfBarLambda}, \eqref{eq:PrLambdaLambdaHalf}, and \eqref{eq:threeQuarterBound} yields the following, for sufficiently small $\epsilon$, sufficiently large $n$, and $n^{3/4} \le t \le n/2$:
	\begin{align*}
		\Pr(\underline{R}_{n}& \ge n^{2/3} \colonBreak b_{t} \notin B_{2\sqrt{m} \epsilon}(\beta)) \n
		& \ge \Pr(n\Lambda_{n}^{\beta}(y_{n}^{\beta}) - \bar{\Lambda}_{t}^{b_{t} - \beta} \ge n^{2/3}\colonBreak b_{t} \notin B_{2\sqrt{m} \epsilon}(\beta))\n
		& \ge \Pr\big(n\Lambda_{n}^{\beta}(y_{n}^{\beta}) - \sup_{\xi \notin B_{2\sqrt{m} \epsilon}(0)}\bar{\Lambda}_{t}^{\xi} \ge n^{2/3}\big)\n
		& \ge \Pr\big(n\Lambda_{n}^{\beta}(y_{n}^{\beta}) - \max_{k \in \{-1, 1\}} \max_{j \in [m]} \bar{\Lambda}_{t}^{k\epsilon \omega_{j}^{\beta}} \ge n^{2/3}\ \cap\ \sup_{\xi \notin B_{2\sqrt{m}\epsilon}(0)}\bar{\Lambda}_{t}^{\xi} \le \max_{k \in \{-1, 1\}} \max_{j \in [m]}\bar{\Lambda}_{t}^{k\epsilon \omega_{j}^{\beta}}\big)\\
		& \ge \Pr\big(n\Lambda_{n}^{\beta}(y_{n}^{\beta}) - \max_{k \in \{-1, 1\}} \max_{j \in [m]} \bar{\Lambda}_{t}^{k\epsilon \omega_{j}^{\beta}} \ge n^{2/3}\big) + \Pr\big(\sup_{\xi \notin B_{2\sqrt{m}\epsilon}(0)}\bar{\Lambda}_{t}^{\xi} \le \max_{k \in \{-1, 1\}} \max_{j \in [m]}\bar{\Lambda}_{t}^{k\epsilon \omega_{j}^{\beta}}\big) - 1 \\
		& \ge 3/4 + 3/4 - 1\\
		& = 1/2.
	\end{align*}
	And now, since $\E(\underline{R}_{n}) = O(\log n)$, by Theorem \ref{p:prop1} and line \eqref{eq:OmegaEqualsOmega}, we have
	\begin{align*}
		O(\log n) & = \E(\underline{R}_{n}) \\
		& \ge n^{2/3}\Pr(b_{t} \notin B_{2\sqrt{m} \epsilon}(\beta))\Pr(\underline{R}_{n}\ge n^{2/3} \colonBreak b_{t} \notin B_{2\sqrt{m} \epsilon}(\beta))\\
		& \ge n^{2/3}\Pr(b_{t} \notin B_{2\sqrt{m} \epsilon}(\beta))/2,
	\end{align*}
	which implies the result.
\end{proof}

\begin{proof}[Lemma \ref{l:rIsHighGivenB} Proof]
	I will show that there exists $C > 0$ that satisfies 
	\begin{align}
		\inf_{b \in B_{\delta/2}(\beta)}\E\big(&\pi_{t}^{b}\Delta_{t}(y_{t-1}^{\psi_{t}^{b}(0)})^{-} + (1-\pi_{t}^{b})\Delta_{t}(y_{t-1}^{\psi_{t}^{b}(\alpha)})^{+}\big) \ge Cm \sigma_{m}^{\beta}/(2t), \label{eq:rIsHighGivenBKeyResult}
	\end{align}
	for all sufficiently large $t$. Combining this result with Lemma \ref{l:probBInControl} yields the desired result:
	\begin{align*}
		\E(r_{t}) & = 
		\E\big(\indicator{b_{t} \in B_{\delta/2}(\beta)} \big(\pi_{t}^{b_{t}}\Delta_{t}(y_{t-1}^{\psi_{t}^{b_{t}}(0)})^{-} + (1 - \pi_{t}^{b_{t}})\Delta_{t}(y_{t-1}^{\psi_{t}^{b_{t}}(a_{t})})^{+}\big)\big)\\
		& = \E\big(\indicator{b_{t} \in B_{\delta/2}(\beta)} \E\big(\pi_{t}^{b}\Delta_{t}(y_{t-1}^{\psi_{t}^{b}(0)})^{-} + (1 - \pi_{t}^{b})\Delta_{t}(y_{t-1}^{\psi_{t}^{b}(a_{t})})^{+}\big)\big|_{b = b_{t}}\big)\\
		& \ge \Pr(b_{t} \in B_{\delta/2}(\beta)) \inf_{b \in B_{\delta/2}(\beta)}\E\big(\pi_{t}^{b}\Delta_{t}(y_{t-1}^{\psi_{t}^{b}(0)})^{-} + (1-\pi_{t}^{b})\Delta_{t}(y_{t-1}^{\psi_{t}^{b}(\alpha)})^{+}\big) \\
		& \ge (1 - n^{-1/2}) Cm \sigma_{m}^{\beta}/(2t),
	\end{align*}
	where the second line holds because $b_{t}$ is independent of the random mapping $b \mapsto \pi_{t}^{b}\Delta_{t}(y_{t-1}^{\psi_{t}^{b}(0)})^{-} + (1 - \pi_{t}^{b})\Delta_{t}(y_{t-1}^{\psi_{t}^{b}(a_{t})})^{+}$, and the last line holds because we're given a large $t$ that satisfies $n^{3/4} \le t \le n/2$. 
	
	Let me briefly outline how we will establish line \eqref{eq:rIsHighGivenBKeyResult}. First, Lemma \ref{l:ballY} implies that there's a $O(1)$ chance that we underestimate the shadow price by at least $4\iota / \sqrt{t}$. I use this fact to establish that there exists some constant $C> 0$ that satisfies the following for sufficiently large $t$:
	\begin{align*}
		\E\big(&(1 - \pi_{t}^{b})\Delta_{t}(y_{t-1}^{\psi_{t}^{b}(a_{t})})^{+}\big)\\
		& \ge C \E\big((1 -\pi_{t}^{b})\big(\indicator{\Delta_{t}(y_{\infty}^{b} - \iota / \sqrt{t}) > 0} - \indicator{\Delta_{t}(y_{\infty}^{b} + \iota / \sqrt{t}) > 0}\big)a_{t}'\iota/\sqrt{t}\big).
	\end{align*}
	This lower bound looks nasty, but it's almost exactly in the form we need to apply our one remaining tool: Assumption \ref{a:marginalProb}. However, before applying this assumption, I must eliminate the pesky $1 - \pi_{t}^{b}$ term. Fortunately, $\E\big(\pi_{t}^{b}\Delta_{t}(y_{t-1}^{\psi_{t}^{b}(0)})^{-}\big)$ honors the same bound, except with $\pi_{t}^{b}$ replacing $1 - \pi_{t}^{b}$, which means that $\E\big(\pi_{t}^{b}\Delta_{t}(y_{t-1}^{\psi_{t}^{b}(0)})^{-} + (1-\pi_{t}^{b})\Delta_{t}(y_{t-1}^{\psi_{t}^{b}(\alpha)})^{+}\big)$ has a corresponding $\pi_{t}^{b}$-free bound, which makes it amenable to Assumption \ref{a:marginalProb}. Finally, the last part of the proof combines this assumption with the fundamental theorem of calculus and Lemma \ref{l:defineDotNew} to express this expectation as an integral over $\ddot{\Lambda}_{\infty}$. 
	
	To begin the proof, let $t$ be large enough so that $b \in B_{\delta/2}(\beta)$ implies $\psi_{t}^{b}(a_{t}) \in B_{\delta}(\beta)$. In this case Lemma \ref{l:ballY} establishes that there exists $C > 0$ that satisfies the following, for all $b \in B_{\delta/2}(\beta)$:
	\begin{align*}
		\Pr\big(&\vertii{\sqrt{t}(y_{t-1}^{\psi_{t}^{b}(a_{t})} - y_{\infty}^{\psi_{t}^{b}(a_{t})}) + 4\iota} \le 1\big) \\
		& \ge \Pr\big(\sup_{b \in B_{\delta}(\beta)} \vertii{\sqrt{t}(y_{t-1}^{b} - y_{\infty}^{b}) + 4\iota} \le 1\big)\\
		& = \Pr\big(\sup_{b \in B_{\delta}(\beta)} \vertii{\sqrt{t}(y_{t-1}^{b} - y_{\infty}^{b}) + 4\iota} \le 1\big)\\
		&> C.
	\end{align*}
	Furthermore, $\vertii{\sqrt{t}(y_{t-1}^{\psi_{t}^{b}(a_{t})} - y_{\infty}^{\psi_{t}^{b}(a_{t})}) + 4\iota} \le 1$ implies the following, when $t$ is large:
	\begin{align*}
		y_{t-1}^{\psi_{t}^{b}(a_{t})} - y_{\infty}^{b} = & (y_{t-1}^{\psi_{t}^{b}(a_{t})} - y_{\infty}^{\psi_{t}^{b}(a_{t})}) + (y_{\infty}^{\psi_{t}^{b}(a_{t})} - y_{\infty}^{b})\\
		& \le -3 \iota/\sqrt{t}  + \iota/\sqrt{t}\\
		& = -2 \iota/\sqrt{t},
	\end{align*}
	where the second line follows because $\vertii{y_{\infty}^{\psi_{t}^{b}(a_{t})} - y_{\infty}^{b}} = o(1/\sqrt{t}) $, by Assumption \ref{a:boundedA} and Lemma \ref{l:lipschitzYinB}. Now combining the previous two results yields the following, for $b \in B_{\delta/2}(\beta)$ and $t$ large:
	\begin{align*}
		\E\big(&(1 - \pi_{t}^{b})\Delta_{t}(y_{t-1}^{\psi_{t}^{b}(a_{t})})^{+}\big)\\
		& \ge \E\big(\indicator{\vertii{\sqrt{t}(y_{t-1}^{\psi_{t}^{b}(a_{t})} - y_{\infty}^{\psi_{t}^{b}(a_{t})}) + 4\iota} \le 1}(1 -\pi_{t}^{b})\Delta_{t}(y_{\infty}^{b} - 2\iota / \sqrt{t})^{+}\big)\\
		& \ge C \E\big((1 -\pi_{t}^{b})\Delta_{t}(y_{\infty}^{b} - 2\iota / \sqrt{t})^{+}\big) \\		
		& \ge C \E\big((1 -\pi_{t}^{b})\indicator{\Delta_{t}(y_{\infty}^{b} - \iota / \sqrt{t}) \ge 0}\Delta_{t}(y_{\infty}^{b} - 2\iota / \sqrt{t})^{+}\big)\\
		& \ge C \E\big((1 -\pi_{t}^{b})\indicator{\Delta_{t}(y_{\infty}^{b} - \iota / \sqrt{t}) \ge 0}a_{t}'\iota/\sqrt{t}\big)\\
		& \ge C \E\big((1 -\pi_{t}^{b})\big(\indicator{\Delta_{t}(y_{\infty}^{b} - \iota / \sqrt{t}) > 0} - \indicator{\Delta_{t}(y_{\infty}^{b} + \iota / \sqrt{t}) > 0}\big)a_{t}'\iota/\sqrt{t}\big).
	\end{align*}
	Note, the third line above holds because $y_{t-1}^{\psi_{t}^{b}(a_{t})}$ is independent of $\Delta_{t}$ and $\pi_{t}$, and the fifth line holds because $\Delta_{t}(y_{\infty}^{b} - \iota / \sqrt{t}) \ge 0$ implies $u_{t} \ge a_{t}'y_{\infty}^{b} - a_{t}'\iota/\sqrt{t}$ and hence implies $\Delta_{t}(y_{\infty}^{b} - 2\iota / \sqrt{t})^{+} \ge a_{t}'\iota/\sqrt{t}$. 
	
	Next, an analogous argument implies that we can set $C$ small enough to satisfy the following, for $b \in B_{\delta/2}(\beta)$ and large $t$:
	\begin{align*}
		\E\big(&\pi_{t}^{b}\Delta_{t}(y_{t-1}^{\psi_{t}^{b}(0)})^{-}\big) \\
		& \ge C\E\big(\pi_{t}^{b}\big(\indicator{\Delta_{t}(y_{\infty}^{b} - \iota / \sqrt{t}) > 0} - \indicator{\Delta_{t}(y_{\infty}^{b} + \iota / \sqrt{t}) > 0}\big)a_{t}'\iota/\sqrt{t}\big).
	\end{align*}
	Finally, adding our two bounds establishes line \eqref{eq:rIsHighGivenBKeyResult}:
	\begin{align*}
		\E\big(&\pi_{t}^{b}\Delta_{t}(y_{t-1}^{\psi_{t}^{b}(0)})^{-} + (1-\pi_{t}^{b})\Delta_{t}(y_{t-1}^{\psi_{t}^{b}(\alpha)})^{+} \colonBreak b \in B_{\delta/2}(\beta) \big)\\
		& = C\E\big(\big(\indicator{\Delta_{t}(y_{\infty}^{b} - \iota / (2\sqrt{t})) > 0} - \indicator{\Delta_{t}(y_{\infty}^{b} + \iota / (2\sqrt{t})) > 0}\big)a_{t}'\iota/\sqrt{t}\big)\\
		& = C /\sqrt{t} \int_{\gamma = -1}^{1} \iota' \tfrac{\partial}{\partial \gamma}\E\big(\indicator{\Delta_{1}(y_{\infty}^{b} - \gamma \iota / (2\sqrt{t})) > 0} a_{1}\big) d\gamma \\
		& = C /\sqrt{t} \int_{\gamma = -1}^{1} \iota'\ddot{\Lambda}_{\infty}(y_{\infty}^{b} - \gamma \iota / \sqrt{t})\iota / (2\sqrt{t})d\gamma\\
		& \ge C /\sqrt{t} \int_{\gamma = -1}^{1} \iota'\iota \sigma_{m}^{\beta}/(4\sqrt{t})d\gamma\\
		& \ge Cm \sigma_{m}^{\beta}/(2t).
	\end{align*}
	The penultimate line above holds because the smallest singular value of $\ddot{\Lambda}_{\infty}(y_{\infty}^{b} - \gamma \iota / \sqrt{t})$ is at least half of the smallest singular value of $\ddot{\Lambda}_{\infty}(y_{\infty}^{\beta})$, when $b$ is near $\beta$ and $t$ is large.
\end{proof}	

\begin{lemma} \label{l:yinh}
	$ (y_{t}^{b} - y)'\dot{\Lambda}_{t}^{b}(y) \le 0 $ for all $ t \in \mathbbm{N} $, $ b \in \mathbbm{R}^{m}_{+} $, and $ y \in \mathbbm{R}^{m}_{+} $.
\end{lemma}
\begin{proof}
	Since $\dot{\Lambda}_{t}^{b}$ is a subgradient, it satisfies $ \Lambda_{t}^{b}(y_{t}^{b}) - \Lambda_{t}^{b}(y) \ge (y_{t}^{b} - y)'\dot{\Lambda}_{t}^{b}(y) $. And since  $\Lambda_{t}^{b}(y_{t}^{b}) \le \Lambda_{t}^{b}(y)$, this yields the result.
\end{proof}

\begin{lemma}\label{l:ballY}
	For all $\gamma \in \mathbbm{R}^{m}$ and $\epsilon > 0$ there exist $\delta, C > 0$ such that $\Pr(\sup_{b \in B_{\delta}(\beta)} \vertii{\sqrt{t}(y_{t}^{b} - y_{\infty}^{b}) - \gamma} \le \epsilon) \ge C$ for all sufficiently large $t$.
\end{lemma}

\begin{proof}
	I will begin with a brief proof sketch. Our primary tool for positioning $y_{t}^{b}$ is is Lemma \ref{l:boxThaty}, which maintains that $y_{t}^{b}$ will be close to $y_{\infty}^{b} + \gamma/\sqrt{t}$ for all $b \in B_{\delta}(\beta)$ if $\dot{\Lambda}^{b}_{t}(y_{\infty}^{b} + (\gamma + \eta k \omega_{j}^{b})/\sqrt{t})$ is close to $\eta k \sigma_{j}^{b}\omega_{j}^{b}$, for all $b \in B_{\delta}(\beta)$, $j \in [m]$, and $k \in \{-1, 1\}$. And with a few triangle inequalities and some basic calculus, I show that this condition holds when $\sqrt{t}(\dot{\Lambda}^{\beta}_{t}(y) - \dot{\Lambda}^{\beta}_{\infty}(y))$ is near $\ddot{\Lambda}_{\infty}(y_{\infty}^{\beta})\gamma$ for all $y$ in the $\nu$-ball of $y_{\infty}^{\beta}$, for some $\nu > 0$. Finally, I use Lemma \ref{l:GaussianProcess} to show that there's an $\Theta(1)$ chance of this happening. This lemma maintains that the mapping $(j, y) \mapsto \sqrt{t} e_{j}'(\dot{\Lambda}_{t}^{\beta}(y) - \dot{\Lambda}_{\infty}^{\beta}(y))$ converges to a Gaussian process whose mean is near $\ddot{\Lambda}_{\infty}(y_{\infty}^{\beta})\gamma$ when we condition on $\sqrt{t}(\dot{\Lambda}^{\beta}_{t}(y_{\infty}^{\beta}) - \dot{\Lambda}^{\beta}_{\infty}(y_{\infty}^{\beta}))$ being near $\ddot{\Lambda}_{\infty}(y_{\infty}^{\beta})\gamma$. 
	
	Lemmas \ref{l:defineDotNew} and \ref{l:lipschitzYinB} imply that we can choose $\delta$ small enough so that $\sigma_{1}^{b} \le 2 \sigma_{1}^{\beta}$ and $\sigma_{m}^{b} \ge \sigma_{m}^{\beta}/2 > 0$ for all $b \in B_{\delta}(\beta)$. And these lemmas also imply that we can choose $t$ large enough to ensure that $y_{\infty}^{b} + (\gamma + \eta k \omega_{j}^{b})/\sqrt{t} \ge 0$ for a given $\eta > 0$ and all $j \in [m]$, $k \in \{-1, 1\}$, and $b \in B_{\delta}(\beta)$. With this, Lemma \ref{l:boxThaty} indicates that $\sup_{b \in B_{\delta}(\beta)} \vertii{\sqrt{t}(y_{t}^{b} - y_{\infty}^{b}) -\gamma} \le \epsilon$ if 
	\begin{align}
		\sup_{b \in B_{\delta}(\beta)}\max_{j \in [m]}\max_{k \in \{-1, 1\}}\vertii{\sqrt{t}\dot{\Lambda}^{b}_{t}(y_{\infty}^{b} + (\gamma + \eta k \omega_{j}^{b})/\sqrt{t}) - \eta k \sigma_{j}^{b}\omega_{j}^{b}} \le \kappa, \label{eq:ballY0}
	\end{align}
	where $\eta \equiv \epsilon/(1 + 8\sqrt{m}\sigma_{1}^{\beta}/\sigma_{m}^{\beta})$ and $\kappa \equiv \eta\sigma_{m}^{\beta}/(4\sqrt{m})$. Further, this inequality holds when the following inequalities hold for all $b \in B_{\delta}(\beta)$, $j \in [m]$, and $k \in \{-1, 1\}$:
	\begin{align}
		\vertii{\sqrt{t}(\dot{\Lambda}^{\beta}_{t}(y_{\infty}^{b} + (\gamma  + \eta k \omega_{j}^{b})/\sqrt{t}) - \dot{\Lambda}^{\beta}_{\infty}(y_{\infty}^{b} + (\gamma  + \eta k \omega_{j}^{b})/\sqrt{t})) + \ddot{\Lambda}_{\infty}(y_{\infty}^{\beta})\gamma} & \le \kappa/3, \label{eq:ballY1}\\
		\vertii{\sqrt{t}\dot{\Lambda}^{b}_{t}(y_{\infty}^{b} + (\gamma  + \eta k \omega_{j}^{b})/\sqrt{t}) - \ddot{\Lambda}_{\infty}(y_{\infty}^{b})\gamma - \eta k \sigma_{j}^{b}\omega_{j}^{b}} & \le \kappa/3,\label{eq:ballY2}\\
		\aq \vertii{\ddot{\Lambda}_{\infty}(y_{\infty}^{b})\gamma - \ddot{\Lambda}_{\infty}(y_{\infty}^{\beta})\gamma} & \le \kappa/3.\label{eq:ballY3}
	\end{align}
	Lines \eqref{eq:ballY1}--\eqref{eq:ballY3} imply line \eqref{eq:ballY0} by the triangle inequality, and by the fact that $\dot{\Lambda}_{t}^{b}(y) - \dot{\Lambda}_{\infty}^{b}(y)$ is independent of $b$, which enables me to change the superscripts in line \eqref{eq:ballY1} from $b$ to $\beta$. I will now show that there's a non-negligible chance that these inequalities hold universally across $b$, $j$, and $k$ in their respective domains.
	
	First, since $B_{\delta}(\beta)$ is compact and  $\ddot{\Lambda}_{\infty}(y_{\infty}^{b})$ is continuous in $b$, by Lemmas \ref{l:defineDotNew} and \ref{l:lipschitzYinB}, it follows that we can set $\delta$ small enough to make inequality \eqref{eq:ballY3} hold universally.
	
	Second, since $\dot{\Lambda}^{b}_{\infty}(y_{\infty}^{b}) = 0$ and $\ddot{\Lambda}_{\infty}$ is locally continuous near $y_{\infty}^{\beta}$, the mean value theorem indicates that there exists $\xi_{tjk}^{b} \in (0, 1)$ for which
	\begin{align*}
		\sqrt{t}\dot{\Lambda}^{b}_{\infty}&(y_{\infty}^{b} + (\gamma  + \eta k \omega_{j}^{b})/\sqrt{t})\\
		& = \sqrt{t}(\dot{\Lambda}^{b}_{\infty}(y_{\infty}^{b} + (\gamma  + \eta k \omega_{j}^{b})/\sqrt{t}) - \dot{\Lambda}^{b}_{\infty}(y_{\infty}^{b})) \\
		& = \ddot{\Lambda}_{\infty}(y_{\infty}^{b} + \xi_{tjk}^{b} (\gamma + \eta k \omega_{j}^{b})/\sqrt{t}) (\gamma + \eta k \omega_{j}^{b})\\
		& = \ddot{\Lambda}_{\infty}(y_{\infty}^{b})\gamma + \eta k \sigma_{j}^{b}\omega_{j}^{b} + \zeta_{tjk}^{b}(\gamma + \eta k \omega_{j}^{b}) ,\\
		\wq \zeta_{tjk}^{b} & \equiv \ddot{\Lambda}_{\infty}(y_{\infty}^{b} + \xi_{tjk}^{b} (\gamma + \eta k \omega_{j}^{b})/\sqrt{t}) - \ddot{\Lambda}_{\infty}(y_{\infty}^{b}) .\no
	\end{align*}
	And the continuity of $\ddot{\Lambda}_{\infty}$ implies that we can set $\delta$ small enough so that
	\begin{align*}
		\sup_{b \in B_{\delta}(\beta)}\max_{j \in [m]}\max_{k \in \{-1, 1\}} \vertii{\zeta_{tjk}^{b}(\gamma + \eta k \omega_{j}^{b})} \le \kappa/3,
	\end{align*}
	for all sufficiently large $t$. Hence, inequality \eqref{eq:ballY2} will hold universally for all sufficiently large $t$ and small $\delta$.
	
	Finally, I will show that for all sufficiently large $t$ the probability that inequality \eqref{eq:ballY1} holds universally across $b \in B_{\delta}(\beta)$, $j \in [m]$, and $k \in \{-1, 1\}$ exceeds some $C > 0$. Since $y_{\infty}^{b}$ and $\omega_{j}^{b}$ are continuous in $b$, by Lemmas \ref{l:defineDotNew} and \ref{l:lipschitzYinB}, it will suffice to show that there exist $\nu > 0$ such that
	\begin{align*}
		\liminf_{t \rightarrow \infty} \Pr\Big(\sup_{y \in B_{\nu}(y_{\infty}^{\beta})}\vertii{\sqrt{t}(\dot{\Lambda}^{\beta}_{t}(y) - \dot{\Lambda}^{\beta}_{\infty}(y)) + \ddot{\Lambda}_{\infty}(y_{\infty}^{\beta})\gamma}  \le \kappa/3\Big) > 0. 
	\end{align*} 
	I will prove this inequality with Lemma \ref{l:GaussianProcess}, which with proposition 3.13 of \cite{eaton1983multivariate} implies that conditional on $\zeta_{t} \equiv \sqrt{t}(\dot{\Lambda}^{\beta}_{t}(y_{\infty}^{\beta}) - \dot{\Lambda}^{\beta}_{\infty}(y_{\infty}^{\beta}))$, the random map $(j, y) \mapsto \sqrt{t} e_{j}'(\dot{\Lambda}_{t}^{\beta}(y) - \dot{\Lambda}_{\infty}^{\beta}(y))$ weakly converges to a Gaussian process with domain $[m] \times B_{\nu}(y_{\infty}^{\beta})$, mean function $\rho^{\zeta_{t}}$, and covariance function $\Xi$, where
	\begin{align*}
		\rho_{j}^{\zeta_{t}}(y) & \equiv e_{j}'\Omega(y, y_{\infty}^{\beta})\Omega(y_{\infty}^{\beta}, y_{\infty}^{\beta})^{-1} \zeta_{t},\\
		\Xi_{j\bar{j}}(y, \bar{y}) & \equiv e_{j}'\Omega(y, \bar{y}) e_{\bar{j}} - e_{j}'\Omega(y, y_{\infty}^{\beta}) \Omega(y_{\infty}^{\beta}, y_{\infty}^{\beta})^{-1} \Omega(y_{\infty}^{\beta}, \bar{y})e_{\bar{j}},\\
		\aq \Omega(y, \bar{y}) & \equiv \E(\indicator{\Delta_{1}(y) > 0}\indicator{\Delta_{1}(\bar{y}) > 0}a_{1}a_{1}') - \E(\indicator{\Delta_{1}(y) > 0}a_{1})\E(\indicator{\Delta_{1}(\bar{y}) > 0}a_{1}').
	\end{align*}
	Since $\Xi$ is independent of $\zeta_{t}$, the random map $(j, y) \mapsto \sqrt{t} e_{j}'(\dot{\Lambda}_{t}^{\beta}(y) - \dot{\Lambda}_{\infty}^{\beta}(y)) - e_{j}'\rho_{j}^{\zeta_{t}}(y)$ is assymptotically independent of $\zeta_{t}$, and hence assymptotically independent of the random map $y \mapsto \rho^{\zeta_{t}}(y)$. Accordingly, for sufficiently large $t$, we have
	\begin{align*}
		\Pr&\Big(\sup_{y \in B_{\nu}(y_{\infty}^{\beta})}\vertii{\sqrt{t}(\dot{\Lambda}^{\beta}_{t}(y) - \dot{\Lambda}^{\beta}_{\infty}(y)) + \ddot{\Lambda}_{\infty}(y_{\infty}^{\beta})\gamma}  \le \kappa/3\Big)\\
		& \ge \Pr\Big(\sup_{y \in B_{\nu}(y_{\infty}^{\beta})}\vertii{\sqrt{t}(\dot{\Lambda}^{\beta}_{t}(y) - \dot{\Lambda}^{\beta}_{\infty}(y)) - \rho^{\zeta_{t}}(y)}  \le \kappa/6\\
		& \qquad\qquad\ \cap\ \sup_{y \in B_{\nu}(y_{\infty}^{\beta})}\vertii{\rho^{\zeta_{t}}(y) - \ddot{\Lambda}_{\infty}(y_{\infty}^{\beta})\gamma} \le \kappa/6\Big)\\
		& \ge p_{t}^{1}p_{t}^{2}/2,\\
		\wq p_{t}^{1} & \equiv \Pr\big(\sup_{y \in B_{\nu}(y_{\infty}^{\beta})}\vertii{\sqrt{t}(\dot{\Lambda}^{\beta}_{t}(y) - \dot{\Lambda}^{\beta}_{\infty}(y)) - \rho^{\zeta_{t}}(y)}  \le \kappa/6\big)\\
		\aq p_{t}^{2} & \equiv \Pr\big(\sup_{y \in B_{\nu}(y_{\infty}^{\beta})}\vertii{\rho^{\zeta_{t}}(y) - \ddot{\Lambda}_{\infty}(y_{\infty}^{\beta})\gamma} \le \kappa/6\big).
	\end{align*}
	
	I will now lower bound probability $p_{t}^{2}$. It is straightforward to confirm that $\vertii{\zeta_{t} - \rho^{\zeta_{t}}(y)} = O(\vertii{y - y_{\infty}^{\beta}})O(\vertii{\zeta_{t}})$, which implies that we can choose $\nu$ small enough so that $\vertii{\zeta_{t} - \ddot{\Lambda}_{\infty}(y_{\infty}^{\beta})\gamma} \le \kappa/12$ implies $\vertii{\zeta_{t} - \rho^{\zeta_{t}}(y)} \le \kappa/12$ for all $y \in B_{\nu}(y_{\infty}^{\beta})$. And this, in turn, implies that 
	\begin{align*}
		p_{t}^{2} & \ge \Pr\Big(\vertii{\zeta_{t} - \ddot{\Lambda}_{\infty}(y_{\infty}^{\beta})\gamma} \le \kappa/12\ \cap\ \sup_{y \in B_{\nu}(y_{\infty}^{\beta})} \vertii{\zeta_{t} - \rho^{\zeta_{t}}(y)} \le \kappa/12 \Big)\\
		& = \Pr\Big(\vertii{\zeta_{t} - \ddot{\Lambda}_{\infty}(y_{\infty}^{\beta})\gamma} \le \kappa/12 \big).
	\end{align*}
	Finally, the limit inferior of this last probability is strictly positive, as $t \rightarrow \infty$, because $\zeta_{t}$ converges to a multivariate normal with a full-rank covariance matrix, by Lemma \ref{l:GaussianProcess}.
	
	I will now lower bound probability $p_{t}^{1}$. First, $\vertii{\zeta_{t} - \rho^{\zeta_{t}}(y)} = O(\vertii{y - y_{\infty}^{\beta}})O(\vertii{\zeta_{t}})$ implies that for a given $M > 0$ we can set $\nu$ small enough so that $\vertii{\zeta_{t}} \le M$ implies $\vertii{\zeta_{t} - \rho^{\zeta_{t}}(y)} \le \kappa/12$ for all $y \in B_{\nu}(y_{\infty}^{\beta})$. And since $\zeta_{t}$ converges to a multivariate normal, we can choose $M$ large enough so that the last equality below holds for all sufficiently large $t$:
	\begin{align*}
		\Pr\big(&\sup_{y \in B_{\nu}(y_{\infty}^{\beta})}\vertii{\sqrt{t}(\dot{\Lambda}^{\beta}_{t}(y) - \dot{\Lambda}^{\beta}_{\infty}(y)) - \rho^{\zeta_{t}}(y)}  \le \kappa/6\big)\\
		& \ge \Pr\big(\sup_{y \in B_{\nu}(y_{\infty}^{\beta})}\vertii{\sqrt{t}(\dot{\Lambda}^{\beta}_{t}(y) - \dot{\Lambda}^{\beta}_{\infty}(y)) -\zeta_{t}} \le \kappa/12\ \cap\ \sup_{y \in B_{\nu}(y_{\infty}^{\beta})} \vertii{\zeta_{t} - \rho^{\zeta_{t}}(y)} \le \kappa/12\big)\\
		& \ge \Pr\big(\sup_{y \in B_{\nu}(y_{\infty}^{\beta})}\vertii{\sqrt{t}(\dot{\Lambda}^{\beta}_{t}(y) - \dot{\Lambda}^{\beta}_{\infty}(y)) -\zeta_{t}}  \le \kappa/12\ \cap\ \vertii{\zeta_{t}} < M\big)\\
		& \ge \Pr\big(\sup_{y \in B_{\nu}(y_{\infty}^{\beta})}\vertii{\sqrt{t}(\dot{\Lambda}^{\beta}_{t}(y) - \dot{\Lambda}^{\beta}_{\infty}(y)) -\zeta_{t}}  \le \kappa/12\big)/2 .
	\end{align*}
	Further, Lemma \ref{l:boundOnSquare2} implies that we can set $\nu$ small enough so that the last inequality in the expression below holds:
	\begin{align*}
		\Pr\big(&\sup_{y \in B_{\nu}(y_{\infty}^{\beta})}\vertii{\sqrt{t}(\dot{\Lambda}^{\beta}_{t}(y) - \dot{\Lambda}^{\beta}_{\infty}(y)) -\zeta_{t}}  > \kappa/12\big) \\
		& = \Pr\big(\sup_{y \in B_{\nu}(y_{\infty}^{\beta})}\vertii{\sqrt{t}(\dot{\Lambda}^{\beta}_{t}(y) - \dot{\Lambda}^{\beta}_{\infty}(y)) -\zeta_{t}}^{2}  > \kappa^{2}/144\big) \\
		& \le \E\big(\sup_{y \in B_{\nu}(y_{\infty}^{\beta})}\vertii{\sqrt{t}(\dot{\Lambda}^{\beta}_{t}(y) - \dot{\Lambda}^{\beta}_{\infty}(y)) -\zeta_{t}}^{2}\big) / (\kappa^{2}/144)\\
		& \le 1/2.
	\end{align*}
	Accordingly, we can set $\nu$ small enough so that $p_{t}^{1} \ge 1/4$ for all sufficiently large $t$.
\end{proof}

\begin{lemma}\label{l:boxThaty}
	If $b$ is close enough to $\beta$ to ensure that $\{\omega_{i}^{b}\}_{i \in [m]}$ and $\{\sigma_{i}^{b}\}_{i \in [m]}$ exist, and if $y \in \mathbbm{R}_{> 0}^{m}$ and $\eta > 0$ satisfy $y + \eta k \omega_{j}^{b} \ge 0$ and $\vertii{\dot{\Lambda}^{b}_{t}(y + \eta k \omega_{j}^{b}) - \eta k \sigma_{j}^{b}\omega_{j}^{b}} \le \eta\sigma_{m}^{b}/(2\sqrt{m})$ for all $j \in [m]$ and $k \in \{-1, 1\}$ then $y_{t}^{b} \in B_{\eta(1 + 2\sqrt{m}\sigma_{1}^{b}/\sigma_{m}^{b})}(y)$.
\end{lemma}
\begin{proof}
	Combining Lemma \ref{l:yinh} with the hypotheses of the current lemma implies the following:
	\begin{align*}
		0 &\ge (y_{t}^{b} - y - \eta k \omega_{j}^{b})'\dot{\Lambda}_{t}^{b}(y + \eta k \omega_{j}^{b}) \\
		& = (y_{t}^{b} - y - \eta k \omega_{j}^{b})'  \eta k\sigma_{j}^{b}\omega_{j}^{b} + (y_{t}^{b} - y - \eta k \omega_{j}^{b})'(\dot{\Lambda}_{t}^{b}(y + \eta k \omega_{j}^{b}) - \eta k \sigma_{j}^{b} \omega_{j}^{b}) \\
		& \ge \eta k\sigma_{j}^{b} (y_{t}^{b} - y)'  \omega_{j}^{b} - \eta^{2} k^{2} \sigma_{j}^{b} \omega_{j}^{b\prime}\omega_{j}^{b} - \vertii{y_{t}^{b} - y - \eta k \omega_{j}^{b}} \vertii{\dot{\Lambda}_{t}^{b}(y + \eta k \omega_{j}^{b}) - \eta k \sigma_{j}^{b} \omega_{j}^{b}}\\
		& \ge \eta k\sigma_{m}^{b} (y_{t}^{b} - y)'  \omega_{j}^{b} - \eta^{2} \sigma_{1}^{b} - (\vertii{y_{t}^{b} - y} + \eta) \eta\sigma_{m}^{b}/(2\sqrt{m}).
	\end{align*}
	And since $ \omega_{1}^{b}, \cdots, \omega_{m}^{b}$ are orthonormal, there must be at least one $j \in [m]$ and one $k \in \{-1, 1\}$ for which $k (y_{t}^{b} - y)'  \omega_{j}^{b} \ge \vertii{y_{t}^{b} - y}/\sqrt{m}$. And thus, we must have
	\begin{align*}
		0 \ge \eta \sigma_{m}^{b} \vertii{y_{t}^{b} - y}/\sqrt{m} - \eta^{2} \sigma_{1}^{b} - \big(\vertii{y_{t}^{b} - y} + \eta\big)\eta\sigma_{m}^{b}/(2\sqrt{m}).
	\end{align*}
	Finally, rearranging the terms yields the result.
\end{proof}

\begin{lemma} \label{l:BoundEindicatorDelta2}
	There exists $\epsilon > 0$ such that if $y, \bar{y} \in B_{\epsilon}(y_{\infty}^{\beta})$ then $\E(\indicator{\Delta_{1}(y) > 0}\big) \Delta_{1}(\bar{y})^{-}) \le 2\sigma_{1}^{\beta}\vertii{\bar{y} - y}^{2}$. 
\end{lemma}

\begin{proof}
	I will first consider the case in which $\bar{y} \ge y$. To begin, note that  $\indicator{\Delta_{1}(y + dy) > 0} \ne \indicator{\Delta_{1}(y) > 0}$ implies that $u_{1} = a_{1}'(y+O(dy))$, in which case $\Delta_{1}(\bar{y})^{-} = a_{1}'(\bar{y} - y + O(dy))$. And with this, Assumption \ref{a:marginalProb} and Lemma \ref{l:defineDotNew} imply the following, for $y$ near $y_{\infty}^{\beta}$:
	\begin{align*}
		\E\big(\big(&\indicator{\Delta_{1}(y + dy) > 0} - \indicator{\Delta_{1}(y) > 0}\big) \Delta_{1}(\bar{y})^{-}\big)\\
		& = \E\big(\big(\indicator{\Delta_{1}(y + dy) > 0} - \indicator{\Delta_{1}(y) > 0}\big)a_{1}\big)'(\bar{y} - y + O(dy)) \\
		& = \big(\tfrac{\partial}{\partial y}\E(\indicator{\Delta_{1}(y) > 0} a_{1}) dy + o(dy)\big)'(\bar{y} - y + O(dy))\\
		& = (\bar{y} - y)'\tfrac{\partial}{\partial y}\E(\indicator{\Delta_{1}(y) > 0} a_{1}) dy + o(dy)\\
		& = (y - \bar{y})'\ddot{\Lambda}(y) dy + o(dy).
	\end{align*}
	Accordingly, for $y$ near $y_{\infty}^{\beta}$ we have $\tfrac{\partial}{\partial y}\E(\indicator{\Delta_{1}(y) > 0}\big) \Delta_{1}(\bar{y})^{-}) = (y - \bar{y})'\ddot{\Lambda}(y)$. And thus, for $y$ and $\bar{y}$ sufficiently close to $y_{\infty}^{\beta}$ we have 
	\begin{align*}
		\E(&\indicator{\Delta_{1}(y) > 0}\big) \Delta_{1}(\bar{y})^{-})\\
		& = \E(\indicator{\Delta_{1}(y) > 0}\big) \Delta_{1}(\bar{y})^{-}) - \E(\indicator{\Delta_{1}(\bar{y}) > 0}\big) \Delta_{1}(\bar{y})^{-})\\
		& = \int_{\gamma = 0}^{1} \tfrac{\partial}{\partial \gamma} \E(\indicator{\Delta_{1}(\bar{y} + \gamma (y - \bar{y})) > 0}\Delta_{1}(\bar{y})^{-}\big) d\gamma\\
		& = \int_{\gamma = 0}^{1} (y - \bar{y})'\ddot{\Lambda}(\bar{y} + \gamma (y - \bar{y}))(y - \bar{y}) d\gamma \\
		& \le \int_{\gamma = 0}^{1} 2\sigma_{1}^{\beta} \vertii{y - \bar{y}}^{2} d\gamma \\
		& = 2 \sigma_{1}^{\beta}\vertii{y - \bar{y}}^{2},
	\end{align*}
	where the penultimate line holds because the largest singular value of $\ddot{\Lambda}(\bar{y} + \gamma (y - \bar{y}))$ is smaller than twice the largest singular value of $y_{\infty}^{\beta}$, when $y$ and $\bar{y}$ are sufficiently close to $y_{\infty}^{\beta}$, by Lemma \ref{l:defineDotNew}.
	
	Now we can use what we've just established to prove the $\bar{y} \ngeq y$ case, since
	\begin{align*}
		\E(&\indicator{\Delta_{1}(y) > 0}\big) \Delta_{1}(\bar{y})^{-})\\
		& \le \E(\indicator{\Delta_{1}(y) > 0}\big) \Delta_{1}(\bar{y} \vee y)^{-})\\
		& \le 2 \sigma_{1}^{\beta}\vertii{y - \bar{y} \vee y}^{2}\\
		& \le 2 \sigma_{1}^{\beta}\vertii{y - \bar{y}}^{2}.
	\end{align*}
\end{proof}

\begin{lemma}\label{l:boundOnSquare}
	There exists $C > 0$ such that $ \E\big(\sup_{b \in \mathbbm{R}_{+}^{m}} \sup_{y \in \mathbbm{R}_{+}^{m}} \vertii{\dot{\Lambda}^{b}_{t}(y) - \dot{\Lambda}^{b}_{\infty}(y)}^{2}\big) \le C/t$ for all $t \in \mathbbm{N}$.
\end{lemma}
\begin{proof}
	I will show that the conditions of \cites{VanderVaart1996} theorems 2.14.2 and 2.14.5 are satisfied. First, to translate the problem into \citeauthors{VanderVaart1996} format, note that
	\begin{align*}
		e_{j}'(\dot{\Lambda}^{b}_{t}(y) - \dot{\Lambda}^{b}_{\infty}(y))
		& = \sum_{s=1}^{t}\lambda_{j}^{y}(u_{s}, a_{s})/t - \E(\lambda_{j}^{y}(u_{1}, a_{1})),\\
		\wq \lambda_{j}^{y}(u_{1}, a_{1}) & \equiv \indicator{\Delta_{1}(y) > 0} e_{j}'a_{1}.
	\end{align*}
	The $\lambda_{j}^{y}$ functions lie under an upper envelope---since $\verti{\lambda_{j}^{y}(u_{1}, a_{1})} \le \vertii{\alpha}$---and so it will suffice to show that \cites{VanderVaart1996} bracketing integral is finite for the set $\{\lambda_{j}^{y}\}_{y \in \mathbbm{R}_{+}^{m}, j \in [m]}$.
	
	To streamline the argument, I will suppose that $a_{1} > 0$, almost surely, and that the conditional distribution of $u_{1}$ given $a_{1}$ is characterized by density function $f$, which is bounded by some $M \in \mathbbm{N}$. These assumptions are not necessary, but the argument is messy without them. 
	
	For a given $\nu > 0$, define $m$-dimensional grid $ G \equiv \gamma \mathbbm{Z}^{m}$, where $\gamma \equiv \nu/(M\vertii{\alpha}_{1}^{2})$. Next, let $ \ell(y) \equiv \max \{g \in G \colonBreak g \le y\} $ represent the largest gridpoint that's weakly less than $ y \in \mathbbm{R}^{m}_{+} $ and let $ h(y) \equiv \min \{g \in G \colonBreak g > \ell(y)\} $ represent the smallest gridpoint that's strictly larger than $ \ell(y) $, so that $h(y) - \ell(y) = \gamma\iota$. Finally, define $U \equiv F_{u}^{-1}(1 - \nu/(2\vertii{\alpha}))$, $A \equiv F_{a}^{-1}(\nu/(2\vertii{\alpha}))$, and $Y \equiv U/A$, where $F_{u}$ is the CDF of $u_{1}$ and $F_{a}$ is the CDF of the smallest element of $a_{1}$ (which we've assumed to be larger than zero).
	
	I will now show that the pair $(\indicator{\vertii{y}_{\infty} \le Y} \lambda_{j}^{h(y)}, \lambda_{j}^{\ell(y \wedge Y)})$ is a $\nu$-bracket that contains $\lambda_{j}^{y}$. First, if $\vertii{y}_{\infty} \le Y$ then 
	\begin{align*}
		\E\big((\lambda_{j}^{\ell(y \wedge Y)}&(u_{1}, a_{1}) - \indicator{\vertii{y}_{\infty} \le Y} \lambda_{j}^{h(\bar{y})}(u_{1}, a_{1}))^{2}\big)^{1/2} \\
		& \le \vertii{\alpha} \Pr\big(u_{1} \in (\ell(y)'a_{1}, \ell(y)'a_{1} + \gamma\iota'a_{1}]\big) \\
		& \le \vertii{\alpha} \E\Big(\Pr\big(u_{1} \in (\ell(y)'a_{1}, \ell(y)'a_{1} + \gamma\vertii{\alpha}_{1}] \colonBreak a_{1}\big)\Big)\\
		& \le \gamma M\vertii{\alpha}_{1}^{2}\\
		& = \nu.
	\end{align*}
	Next, if $e_{i}'y > Y$ then
	\begin{align*}
		\E\big((\lambda_{j}^{\ell(y \wedge Y)}&(u_{1}, a_{1}) - \indicator{\vertii{y}_{\infty} \le Y} \lambda_{j}^{h(\bar{y})}(u_{1}, a_{1}))^{2}\big)^{1/2}\\
		& \le \vertii{\alpha} \Pr\big(u_{1} > a_{1}'(y \wedge Y)\big)\\
		& \le \vertii{\alpha} \Pr(u_{1} > Y a_{1}'e_{i})\\
		& \le \vertii{\alpha} \big(\Pr(a_{1}'e_{i} \le A) + \Pr(u_{1} > Y A)\big)\\
		& \le \nu .
	\end{align*}
	
	Finally, the set $\{(\indicator{\vertii{y}_{\infty} \le Y} \lambda_{j}^{h(y)}, \lambda_{j}^{\ell(y \wedge Y)})\}_{y \in \mathbbm{R}_{+}^{m}}$ has $N_{\nu} = (\lfloor Y/\gamma \rfloor + 1)^{m}$ elements. Note that $E(u_{1}) < \infty$ implies $U < 2 \vertii{\alpha}/\nu$, for all sufficiently small $\nu$. Hence, for small enough $\nu$ we have $N_{\nu}  < (\lfloor 2 \vertii{\alpha}/(\gamma \nu A) \rfloor + 1)^{m} \le (\lfloor 2 \vertii{\alpha}_{1}^{3} M/(\nu^{2} A) \rfloor + 1)^{m} \le C /\nu^{2m}$, which implies that $\int_{\nu = 0}^{1} \sqrt{\log N_{\nu}} d\nu < \infty$.
\end{proof}

\begin{lemma}\label{l:boundOnSquare2}
	For all $\delta > 0$ there exist $\epsilon > 0$ such that 
	$\E\big(\sup_{y \in B_{\epsilon}(y_{\infty}^{\beta})} \vertii{\dot{\Lambda}^{b}_{t}(y) - \dot{\Lambda}^{b}_{\infty}(y) - \dot{\Lambda}^{b}_{t}(y_{\infty}^{\beta}) + \dot{\Lambda}^{b}_{\infty}(y_{\infty}^{\beta})}^{2}\big) \le \delta/t$ for all $t \in \mathbbm{N}$.
\end{lemma}
\begin{proof}
	The proof is similar to the proof of Lemma \ref{l:boundOnSquare}, except with 
	\begin{align*}
		\lambda_{j}^{y}(u_{1}, a_{1}) & \equiv (\indicator{\Delta_{1}(y) > 0} - \indicator{\Delta_{1}(y_{\infty}^{\beta}) > 0}) e_{j}'a_{1}.
	\end{align*}
	Modifying the proof of Lemma \ref{l:boundOnSquare} establishes that the bracketing integral of $\{\lambda_{j}^{y}\}_{y \in B_{\epsilon}(y_{\infty}^{\beta}), j \in [m]}$ is uniformly bounded in $\epsilon \in [0, 1]$. Further, $\lambda_{j}^{y}$ is bounded by envelope function $\bar{\lambda}_{j}^{y}$, where
	\begin{align*}
		\bar{\lambda}_{j}^{y}(u_{1}, a_{1}) & \equiv (\indicator{\Delta_{1}(\underline{y}) > 0} - \indicator{\Delta_{1}(\bar{y}) > 0}) e_{j}'a_{1},\\
		\underline{y} & \equiv y \wedge y_{\infty}^{\beta}, \\
		\aq \bar{y} & \equiv y \vee y_{\infty}^{\beta}.
	\end{align*}
	I will now show that the second moment of this envelope can be made arbitrarily small. First, Assumption \ref{a:marginalProb} and Lemma \ref{l:defineDotNew} yield the following:
	\begin{align*}
		\E(\bar{\lambda}_{j}^{y}(u_{1}, a_{1})^{2}) & = \E\big((\indicator{\Delta_{1}(\underline{y}) > 0} - \indicator{\Delta_{1}(\bar{y}) > 0}) (e_{j}'a_{1})^{2}\big)\\
		& \le \vertii{\alpha}e_{j}'\E\big((\indicator{\Delta_{1}(\underline{y}) > 0} - \indicator{\Delta_{1}(\bar{y}) > 0}) a_{1}\big)\\
		& = \vertii{\alpha}e_{j}' \int_{\gamma = 0}^{1} \tfrac{\partial}{\partial \gamma}\E\big(\indicator{\Delta_{1}(\bar{y} + \gamma (\underline{y} - \bar{y})) > 0} a_{1}\big) d\gamma \\
		& = -\vertii{\alpha}e_{j}' \int_{\gamma = 0}^{1} \ddot{\Lambda}_{\infty}(\bar{y} + \gamma (\underline{y} - \bar{y})) (\underline{y} - \bar{y})d\gamma\\
		& = -\vertii{\alpha}e_{j}' \ddot{\Lambda}_{\infty}(\bar{y} + \bar{\gamma} (\underline{y} - \bar{y})) (\underline{y} - \bar{y}),
	\end{align*}
	for some $\bar{\gamma} \in [0, 1]$. And since we constrain $y \in B_{\epsilon}(y_{\infty}^{\beta})$, it follows that $\ddot{\Lambda}_{\infty}(\bar{y} + \bar{\gamma} (\underline{y} - \bar{y})) \rightarrow \ddot{\Lambda}_{\infty}(y_{\infty}^{\beta})$ and $(\underline{y} - \bar{y}) \rightarrow 0$ as $\epsilon \rightarrow 0$. Accordingly, $\E(\bar{\lambda}_{j}^{y}(u_{1}, a_{1})^{2}) \rightarrow 0$ as $\epsilon \rightarrow 0$, which with theorems 2.14.2 and 2.14.5 of \cite{VanderVaart1996} establishes the result. 
\end{proof}

\begin{lemma}\label{l:GaussianProcessLambda}
	For any compact $\Omega \subset \mathbbm{R}_{+}^{m}$, there exists $C > 0$ such that
	
	\noindent
	$\E\big(\sup_{y, \bar{y} \in \Omega} \verti{\Lambda_{t}^{b}(y) - \Lambda_{\infty}^{b}(y) - \Lambda_{t}^{\bar{b}}(\bar{y}) + \Lambda_{\infty}^{\bar{b}}(\bar{y})}^{2}\big) \le C/t$ for all $t \in \mathbbm{N}$, $b, \bar{b} \in \mathbbm{R}_{+}^{m}$.
\end{lemma}
\begin{proof}
	Like in the proofs of Lemmas \ref{l:boundOnSquare} and \ref{l:boundOnSquare2}, I will use theorems 2.14.2 and 2.14.5 of \cites{VanderVaart1996}. As before, I will cast the problem as an empirical process with a new set of functions:
	\begin{align*}
		\Lambda_{t}^{b}(y) - \Lambda_{\infty}^{b}(y) - \Lambda_{t}^{\bar{b}}(\bar{y}) + \Lambda_{\infty}^{\bar{b}}(\bar{y}) & = \sum_{s = 1}^{t}\lambda_{y}^{\bar{y}}(u_{s}, a_{s}) /t - \E(\lambda_{y}^{\bar{y}}(u_{1}, a_{1})),\\
		\wq \lambda_{y}^{\bar{y}}(u_{1}, a_{1}) & \equiv  (u_{1} - a_{1}'y)^{+} - (u_{1} - a_{1}'\bar{y})^{+}.
	\end{align*}
	Note that $\verti{\lambda_{y}^{\bar{y}}(u_{1}, a_{1})} \le \vertii{\alpha}\text{diam}(\Omega) < \infty$. Accordingly, it will suffice to show that the bracketing integral of $\{\lambda_{j}^{y}\}_{y \in \Omega, \bar{y} \in \Omega}$ is finite.
	
	To bound this bracketing integral, define $m$-dimensional grid $ G \equiv \gamma\mathbbm{Z}^{m}$, where $\gamma \equiv \nu/(4 \vertii{\alpha} \vertii{\iota})$. Next, let $ \ell(y) \equiv \max \{g \in G \colonBreak g \le y\} $ represent the largest gridpoint that's weakly less than $ y \in \mathbbm{R}^{m}_{+} $ and let $ h(y) \equiv \min \{g \in G \colonBreak g > \ell(y)\} $ represent the smallest gridpoint that's strictly larger than $ \ell(y) $. By design, we have
	\begin{align*}
		\lambda_{\ell(y)}^{h(\bar{y})} - \lambda_{h(y)}^{\ell(\bar{y})} & \le (u_{1} - a_{1}'(y - \gamma\iota))^{+} - (u_{1} - a_{1}'(\bar{y} + \gamma\iota))^{+} \\
		& \quad -  (u_{1} - a_{1}'(y + \gamma\iota))^{+} + (u_{1} - a_{1}'(\bar{y} - \gamma\iota))^{+}\\
		& \le 4 \gamma \vertii{\alpha}\vertii{\iota}\\
		& = \nu.
	\end{align*}
	Accordingly, the pair $(\lambda_{h(y)}^{\ell(\bar{y})}, \lambda_{\ell(y)}^{h(\bar{y})})$ is a $\nu$-bracket that contains $\lambda_{y}^{\bar{y}}$. Finally, there are only $O(\nu^{2m})$ such brackets; hence, the bracketing integral is finite.
\end{proof}

\begin{lemma}\label{l:GaussianProcess}
	For all sufficiently small $\epsilon > 0$, the random mapping $(j, y) \mapsto \sqrt{t} e_{j}'(\dot{\Lambda}_{t}^{b}(y) - \dot{\Lambda}_{\infty}^{b}(y))$, with $j \in [m]$ and $y \in R_{+}^{m}$, weakly converges, as $t \rightarrow \infty$, to a mean-zero Gaussian process with domain $[m] \times R_{+}^{m}$ and covariance function $\Sigma_{j\bar{j}}(y, \bar{y})\equiv e_{j}'\E(\indicator{\Delta_{1}(y) > 0}\indicator{\Delta_{1}(\bar{y}) > 0}a_{1}a_{1}') - \E(\indicator{\Delta_{1}(y) > 0}a_{1})\E(\indicator{\Delta_{1}(\bar{y}) > 0}a_{1}') e_{\bar{j}}$.
\end{lemma}
\begin{proof}
	This is a direct application of theorem 2.3 of \citeauthor{Kosorok2008}, a classical empirical processes result. The proof of Lemma \ref{l:boundOnSquare} establishes that the corresponding bracketing integral is finite.
\end{proof}

\end{document}